\documentclass[10pt]{article} % For LaTeX2e
\usepackage[preprint]{tmlr}
\usepackage{tikz}
\usepackage{graphicx}
\usepackage{subcaption}
\usepackage{float}
\usepackage{xcolor}
\usepackage{amsmath}
\usepackage{booktabs}
\usepackage{standalone}
\usepackage{tikz}
\usepackage{longtable}
\usepackage{placeins}
\usepackage{algorithm}
\usepackage{algpseudocode}
\usetikzlibrary{arrows.meta, positioning, calc, shapes.arrows}
\usepackage{amsmath,amsfonts,bm}

\def\eqref#1{equation~\ref{#1}}
\def\1{\bm{1}}

\DeclareMathAlphabet{\mathsfit}{\encodingdefault}{\sfdefault}{m}{sl}
\SetMathAlphabet{\mathsfit}{bold}{\encodingdefault}{\sfdefault}{bx}{n}

\usepackage{hyperref}
\usepackage{url}
\hypersetup{
    colorlinks,
    linkcolor={red!50!black},
    citecolor={blue!50!black},
    urlcolor={blue!80!black}
}
\title{Automated Discovery Has No Universally Superior Harness}
\author{\name Akshat Gupta \email akshat.gupta@berkeley.edu \\
    \addr UC Berkeley
    \AND
    \name Jermaine Lei \email jermainelei@berkeley.edu \\
    \addr UC Berkeley
    \AND
    \name Alexander Lu \email alexander.lu@berkeley.edu\\
    \addr UC Berkeley
    \AND
    \name Gopala Anumanchipalli \email gopala@berkeley.edu \\
    \addr UC Berkeley
    \AND
    \name Leshem Choshen \email leshem.choshen@weizmann.ac.il\\
    \addr MIT, MIT-IBM Watson AI Lab
}

\definecolor{findingbg}{RGB}{243,248,255}
\definecolor{findingborder}{RGB}{58,102,171}
\newcommand{\findingbox}[1]{%
  \par\medskip
  \noindent\begingroup
  \setlength{\fboxsep}{8pt}%
  \setlength{\fboxrule}{0.6pt}%
  \fcolorbox{findingborder}{findingbg}{%
    \parbox{\dimexpr\linewidth-2\fboxsep-2\fboxrule\relax}{#1}%
  }%
  \endgroup
  \par\medskip
}

\newcommand{\treesweepsubfiguregraphics}[1]{\makebox[\linewidth][c]{\includegraphics[height=0.235\textheight]{#1}}}

\newcommand{\appendixfullgridgraphics}[1]{\makebox[\linewidth][c]{\includegraphics[height=0.235\textheight]{#1}}}
\newcommand{\appendixprogressiongraphics}[1]{\makebox[\linewidth][c]{\includegraphics[height=0.185\textheight]{#1}}}
\newcommand{\appendixsweepgraphics}[1]{\makebox[\linewidth][c]{\includegraphics[height=0.235\textheight]{#1}}}

\def\month{MM}  % Insert correct month for camera-ready version
\def\year{YYYY} % Insert correct year for camera-ready version
\def\openreview{\url{https://openreview.net/forum?id=XXXX}} % Insert correct link to OpenReview for camera-ready version

\begin{document}

\raggedbottom

\maketitle

\begin{abstract}
Autonomous discovery systems such as OpenEvolve and TTT-Discover are often used as general-purpose harnesses. However, in practice these are composite systems combining several design choices about archives, parent selection, exploration, and budget allocation into a single recipe. Because discovery runs are expensive and inherently stochastic, existing harnesses are often compared using too few independent trials to distinguish key methodological improvements from run-to-run variance. We systematically decompose OpenEvolve-style evolutionary search and the TTT-Discover search harness into its constituent components and systematically evaluate 30 budget-matched harnesses across 12 model–problem pairs using more than 3.1 million LLM rollouts and repeated-trial statistical analysis. Our results show that \textbf{discovery harnesses have a generalization problem}: No fixed harness is reliably superior across the evaluated model–problem pairs, and variants of \textbf{OpenEvolve generally underperform} simpler alternatives. Thus, \textbf{harness choice is better viewed as a hyperparameter} rather than as a universal recipe, and should be tailored to the specific problem and underlying model. We also find that early discovery progress predicts final performance, and use this property to present a budget-matched adaptive-allocation experiment that starts multiple harnesses, prunes weak partial runs, and reallocates compute to stronger survivors, outperforming both commitment to a randomly sampled fixed harness and a non-adaptive harness ensemble. \textbf{Together, these results motivate shifting from fixed harness selection to online adaptation guided by early performance.} We release all run pools including baseline null distributions for every model–problem pair as reusable statistical infrastructure against for future harness proposals.  Code, data, and run pools are available at \url{https://github.com/akshat57/harness-generalization}.
\end{abstract}

\section{Introduction}
LLM-guided autonomous discovery has increasingly moved from one-shot generation to iterative generate–evaluate–improve loops. In systems such as FunSearch \citep{romera2024mathematical}, AlphaEvolve \citep{novikov2025alphaevolve}, OpenEvolve \citep{openevolve}, and TTT-Discover \citep{yuksekgonul2026learning}, an LLM proposes candidate programs or solutions, an external evaluator scores them, and a harness decides which previous artifacts should be expanded next. The harness therefore plays a central role in directing the discovery process. Existing discovery harnesses typically combine several design choices—including archive construction, parent selection, exploration, search structure, and budget allocation—into a single composite recipe, and the discovery process itself is inherently stochastic, partly by design. Due to compute and cost constraints, these composite systems are typically evaluated with only a handful of independent trials---far too few to separate genuine methodological ingredients from run-to-run variance.  As a result, it is often unclear whether and how individual components contribute to final performance, and whether their benefits transfer across models and problems. Our central question in this paper is not whether LLM-guided discovery systems can solve challenging tasks. Recent work has already shown that they can. Instead, we ask a more diagnostic question: \emph{when a discovery system succeeds, do its harness choices transfer as general methodological principles, or are they model- and problem-dependent hyperparameters?}

\newcommand{\mainfigpanelheight}{0.30\textheight}

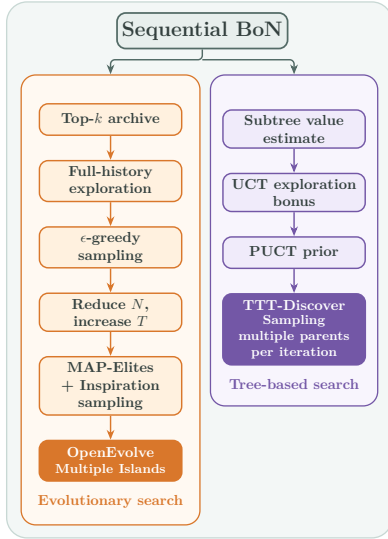
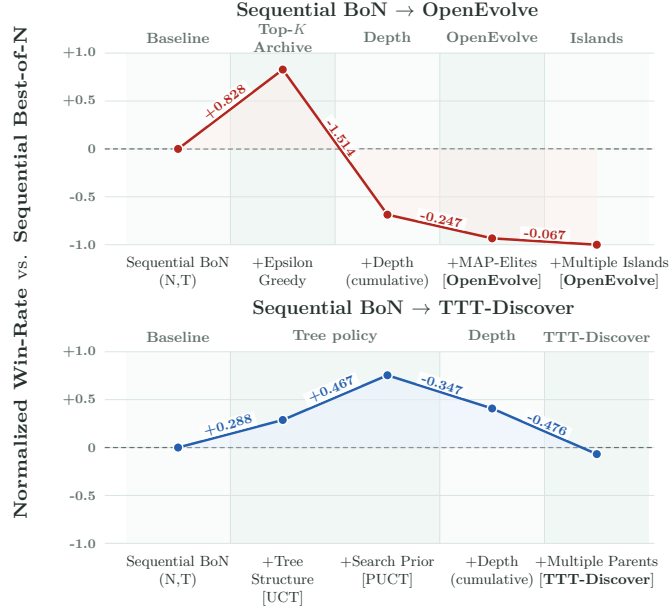
\begin{figure*}[t]
\centering

\begin{subfigure}[t]{0.31\textwidth}
  \vspace{0pt}
  \centering
  % Simple progression from Sequential BoN to OpenEvolve and TTT-Discover.
% Preamble requirements:
%   \usepackage{tikz}
%   \usetikzlibrary{arrows.meta,positioning,calc,fit,backgrounds}
\usetikzlibrary{fit,backgrounds}
\centering

\definecolor{flowSlate}{HTML}{33403E}
\definecolor{flowGray}{HTML}{63736F}
\definecolor{flowCanvas}{HTML}{F3F7F6}
\definecolor{flowCanvasEdge}{HTML}{C8D8D4}
\definecolor{flowRoot}{HTML}{E8EFED}
\definecolor{flowRootEdge}{HTML}{536B66}
\definecolor{flowOrange}{HTML}{FFF1DF}
\definecolor{flowOrangeEdge}{HTML}{D9772B}
\definecolor{flowOrangePanel}{HTML}{FFF9F2}
\definecolor{flowPurple}{HTML}{F0EBFA}
\definecolor{flowPurpleEdge}{HTML}{7357A6}
\definecolor{flowPurplePanel}{HTML}{F8F5FD}
\providecommand{\mainfigpanelheight}{0.30\textheight}

\resizebox{\linewidth}{!}{%
\begin{tikzpicture}[
  font=\sffamily,
  node distance=5.8mm and 6mm,
  root/.style={
    rounded corners=6pt, draw=flowRootEdge, fill=flowRoot,
    line width=1.2pt, minimum width=3.55cm, minimum height=0.86cm,
    inner ysep=4pt, align=center, font=\Large\bfseries, text=flowSlate
  },
  openstep/.style={
    rounded corners=6pt, draw=flowOrangeEdge, fill=flowOrange,
    line width=1pt, text width=3.02cm, minimum height=0.76cm,
    inner xsep=6pt, inner ysep=3.6pt,
    align=center, font=\normalsize\bfseries, text=flowSlate
  },
  tttstep/.style={
    rounded corners=6pt, draw=flowPurpleEdge, fill=flowPurple,
    line width=1pt, text width=3.02cm, minimum height=0.76cm,
    inner xsep=6pt, inner ysep=3.6pt,
    align=center, font=\normalsize\bfseries, text=flowSlate
  },
  openterminal/.style={openstep, fill=flowOrangeEdge, text=white,
    line width=1.2pt, minimum height=0.96cm, inner ysep=4pt},
  tttterminal/.style={tttstep, fill=flowPurpleEdge, text=white,
    line width=1.2pt, minimum height=0.96cm, inner ysep=4pt},
  branchlabel/.style={font=\normalsize\bfseries, align=center},
  branchpanel/.style={
    rounded corners=10pt, line width=0.9pt, inner sep=8pt
  },
  canvas/.style={
    rounded corners=14pt, draw=flowCanvasEdge, fill=flowCanvas,
    line width=0.9pt, inner xsep=18pt, inner ysep=7pt
  },
  arrow/.style={-{Stealth[length=2.4mm,width=1.9mm]},
    line width=1pt, draw=flowRootEdge, rounded corners=3pt},
  openarrow/.style={-{Stealth[length=2.4mm,width=1.9mm]},
    line width=1pt, draw=flowOrangeEdge},
  tttarrow/.style={-{Stealth[length=2.4mm,width=1.9mm]},
    line width=1pt, draw=flowPurpleEdge}
]

% Shared starting point
\node[root] (bon) {Sequential BoN};

% First stages
\node[branchlabel, text=flowOrangeEdge, below=9mm of bon, xshift=-2.22cm]
  (openlabel) {};
\node[branchlabel, text=flowPurpleEdge, below=9mm of bon, xshift=2.22cm]
  (tttlabel) {};

\node[openstep, below=1.5mm of openlabel] (topk) {Top-$k$ archive};
\node[tttstep, below=3mm of tttlabel] (subtree)
  {Subtree value\\estimate};

% OpenEvolve branch
\node[openstep, below=of topk] (history) {Full-history exploration};
\node[openstep, below=of history] (epsilon) {$\epsilon$-greedy sampling};
\node[openstep, below=of epsilon] (budgetopen)
  {Reduce $N$, increase $T$};
\node[openstep, below=of budgetopen] (mapelites)
  {MAP-Elites + Inspiration\\sampling};
\node[openterminal, below=of mapelites] (islands)
  {OpenEvolve\\[-1pt]\small Multiple Islands};

% TTT-Discover branch.
\node[tttstep, below=of subtree] (uct)
  {UCT exploration\\bonus};
\node[tttstep, below=of uct] (puct)
  {PUCT prior};
\node[tttterminal, below=of puct] (multiparent)
  {TTT-Discover\\[-1pt]\small Sampling multiple parents\\[-1pt]
   \small per iteration};

% Branch family labels sit at the bottom of each colored panel.
\node[branchlabel, text=flowOrangeEdge, below=2mm of islands]
  (openfamily) {Evolutionary search};
\node[branchlabel, text=flowPurpleEdge, below=2mm of multiparent]
  (tttfamily) {Tree-based search};
\coordinate (canvasbottom) at ($(openfamily.south)+(0,-3.5mm)$);

% Soft branch backgrounds, matching the card treatment in the other figures.
\begin{scope}[on background layer]
  \node[canvas, fit=(bon)(openlabel)(islands)(openfamily)(canvasbottom)(tttlabel)(multiparent)(tttfamily)]
        (figurecanvas) {};
  \node[branchpanel, fit=(openlabel)(topk)(history)(epsilon)(budgetopen)
        (mapelites)(islands)(openfamily), fill=flowOrangePanel, draw=flowOrangeEdge]
        (openpanel) {};
  \node[branchpanel, fit=(tttlabel)(subtree)(uct)(puct)(multiparent)(tttfamily),
        fill=flowPurplePanel, draw=flowPurpleEdge]
        (tttpanel) {};
\end{scope}

% Split from the root into the two vertical lanes.
\coordinate (split) at ($(bon.south)+(0,-2.5mm)$);
\draw[arrow] (bon.south) -- (split) -| (openpanel.north);
\draw[arrow] (split) -| (tttpanel.north);

\draw[openarrow] (topk) -- (history);
\draw[openarrow] (history) -- (epsilon);
\draw[openarrow] (epsilon) -- (budgetopen);
\draw[openarrow] (budgetopen) -- (mapelites);
\draw[openarrow] (mapelites) -- (islands);

\draw[tttarrow] (subtree) -- (uct);
\draw[tttarrow] (uct) -- (puct);
\draw[tttarrow] (puct) -- (multiparent);

\end{tikzpicture}%
}

\label{fig:search_progression_flow}
  \vspace{1.7em}
  \caption{\textbf{Harness decomposition}}
  \label{fig:search_progression}
\end{subfigure}
\hfill
\begin{subfigure}[t]{0.58\textwidth}
  \centering
  \vspace{0pt}
  \centering
\definecolor{deltaInk}{HTML}{26312F}
\definecolor{deltaMuted}{HTML}{65736F}
\definecolor{deltaGrid}{HTML}{D8E5E1}
\definecolor{deltaBandA}{HTML}{FAFCFB}
\definecolor{deltaBandB}{HTML}{F1F7F5}
\definecolor{deltaRed}{HTML}{B3261E}
\definecolor{deltaRedFill}{HTML}{FDECEA}
\definecolor{deltaBlue}{HTML}{255FAE}
\definecolor{deltaBlueFill}{HTML}{E8F1FF}

\newcommand{\plotbox}[1]{\noindent\resizebox{\linewidth}{!}{#1}}
\newlength{\yAxisTitleShift}
\setlength{\yAxisTitleShift}{-2.85cm}

\noindent
\begin{tabular}{@{}c@{\hspace{0.55em}}c@{}}
\makebox[0pt][c]{\raisebox{\yAxisTitleShift}{\rotatebox{90}{\scriptsize\bfseries\color{deltaInk}Normalized Win-Rate \textnormal{vs.} Sequential Best-of-N}}}
&
\begin{minipage}{0.94\textwidth}
\plotbox{%
\begin{tikzpicture}[
  x=2.24cm,
  y=2.05cm,
  font=\rmfamily,
  axis/.style={draw=deltaGrid, line width=0.45pt},
  zero/.style={draw=deltaMuted, line width=0.75pt, dashed, dash pattern=on 3pt off 2pt},
  line/.style={draw=deltaRed, line width=1.55pt, line cap=round, line join=round},
  point/.style={circle, draw=white, fill=deltaRed, line width=0.9pt, inner sep=2.05pt},
  xtick/.style={font=\small, text=deltaInk, align=center},
  ytick/.style={font=\footnotesize\bfseries, text=deltaMuted, anchor=east},
  group/.style={font=\small\bfseries, text=deltaMuted, align=center},
  title/.style={font=\large\bfseries, text=deltaInk, align=center},
  value/.style={
    font=\footnotesize\bfseries,
    text=deltaRed,
    align=center,
    sloped,
    above,
    fill=white,
    fill opacity=0.92,
    text opacity=1,
    inner xsep=2.4pt,
    inner ysep=1.2pt
  }
]
  \def\ymin{-1.00}
  \def\ymax{1.00}
  \def\xmin{-0.70}
  \def\xmax{4.55}
  \path[use as bounding box] (-1.35,-1.50) rectangle (5.15,1.52);

  \fill[deltaBandA] (-0.5,\ymin) rectangle (0.5,\ymax);
  \fill[deltaBandB] (0.5,\ymin) rectangle (1.5,\ymax);
  \fill[deltaBandA] (1.5,\ymin) rectangle (2.5,\ymax);
  \fill[deltaBandB] (2.5,\ymin) rectangle (3.5,\ymax);
  \fill[deltaBandA] (3.5,\ymin) rectangle (4.5,\ymax);

  \foreach \x in {0.5,1.5,2.5,3.5} {
    \draw[axis] (\x,\ymin) -- (\x,\ymax);
  }
  \foreach \y/\lab in {-1/-1.0,-0.5/-0.5,0/0,+0.5/+0.5,+1/+1.0} {
    \draw[axis] (\xmin,\y) -- (\xmax,\y);
    \node[ytick] at (\xmin-0.03,\y) {\lab};
  }
  \draw[zero] (\xmin,0) -- (\xmax,0);

  \node[group] at (0,1.16) {Baseline};
  \node[group] at (1,1.16) {Top-$K$\\Archive};
  \node[group] at (2,1.16) {Depth};
  \node[group] at (3,1.16) {OpenEvolve};
  \node[group] at (4,1.16) {Islands};

  \node[title] at (2,1.44) {Sequential BoN $\to$ OpenEvolve};

  \fill[deltaRedFill, opacity=0.42]
    (0,0) -- (1,0.8281) -- (2,-0.6863) -- (3,-0.9334) --
    (4,-1.0000) -- (4,0) -- cycle;

  \draw[line]
    (0,0) -- node[value, midway] {+0.828}
    (1,0.8281) -- node[value, midway] {-1.514}
    (2,-0.6863) -- node[value, midway] {-0.247}
    (3,-0.9334) -- node[value, midway] {-0.067}
    (4,-1.0000);

  \foreach \x/\y in {0/0.0000,1/0.8281,2/-0.6863,3/-0.9334,4/-1.0000} {
    \node[point] at (\x,\y) {};
  }

  \node[xtick, anchor=north] at (0,\ymin-0.070) {Sequential BoN\\(N,T)};
  \node[xtick, anchor=north] at (1,\ymin-0.070) {+Epsilon\\Greedy};
  \node[xtick, anchor=north] at (2,\ymin-0.070) {+Depth\\(cumulative)};
  \node[xtick, anchor=north] at (3,\ymin-0.070) {+MAP-Elites\\{[\textbf{OpenEvolve}]}};
  \node[xtick, anchor=north] at (4.12,\ymin-0.070) {+Multiple Islands\\{[\textbf{OpenEvolve}]}};
\end{tikzpicture}%
}

\vspace{0.25em}

\plotbox{%
\begin{tikzpicture}[
  x=2.24cm,
  y=2.05cm,
  font=\rmfamily,
  axis/.style={draw=deltaGrid, line width=0.45pt},
  zero/.style={draw=deltaMuted, line width=0.75pt, dashed, dash pattern=on 3pt off 2pt},
  line/.style={draw=deltaBlue, line width=1.55pt, line cap=round, line join=round},
  point/.style={circle, draw=white, fill=deltaBlue, line width=0.9pt, inner sep=2.05pt},
  xtick/.style={font=\small, text=deltaInk, align=center},
  ytick/.style={font=\footnotesize\bfseries, text=deltaMuted, anchor=east},
  group/.style={font=\small\bfseries, text=deltaMuted, align=center},
  title/.style={font=\large\bfseries, text=deltaInk, align=center},
  value/.style={
    font=\footnotesize\bfseries,
    text=deltaBlue,
    align=center,
    sloped,
    above,
    fill=white,
    fill opacity=0.92,
    text opacity=1,
    inner xsep=2.4pt,
    inner ysep=1.2pt
  }
]
  \def\ymin{-1.00}
  \def\ymax{1.00}
  \def\xmin{-0.70}
  \def\xmax{4.55}
  \path[use as bounding box] (-1.35,-1.50) rectangle (5.15,1.52);

  \fill[deltaBandA] (-0.5,\ymin) rectangle (0.5,\ymax);
  \fill[deltaBandB] (0.5,\ymin) rectangle (2.5,\ymax);
  \fill[deltaBandA] (2.5,\ymin) rectangle (3.5,\ymax);
  \fill[deltaBandB] (3.5,\ymin) rectangle (4.5,\ymax);

  \foreach \x in {0.5,2.5,3.5} {
    \draw[axis] (\x,\ymin) -- (\x,\ymax);
  }
  \foreach \y/\lab in {-1/-1.0,-0.5/-0.5,0/0,+0.5/+0.5,+1/+1.0} {
    \draw[axis] (\xmin,\y) -- (\xmax,\y);
    \node[ytick] at (\xmin-0.03,\y) {\lab};
  }
  \draw[zero] (\xmin,0) -- (\xmax,0);

  \node[group] at (0,1.16) {Baseline};
  \node[group] at (1.5,1.16) {Tree policy};
  \node[group] at (3,1.16) {Depth};
  \node[group] at (4,1.16) {TTT-Discover};

  \node[title] at (2.25,1.44) {Sequential BoN $\to$ TTT-Discover};

  \fill[deltaBlueFill, opacity=0.48]
    (0,0) -- (1,0.2876) -- (2,0.7546) -- (3,0.4077) --
    (4,-0.0686) -- (4,0) -- cycle;

  \draw[line]
    (0,0) -- node[value, midway] {+0.288}
    (1,0.2876) -- node[value, midway] {+0.467}
    (2,0.7546) -- node[value, midway] {-0.347}
    (3,0.4077) -- node[value, midway] {-0.476}
    (4,-0.0686);

  \foreach \x/\y in {0/0.0000,1/0.2876,2/0.7546,3/0.4077,4/-0.0686} {
    \node[point] at (\x,\y) {};
  }

  \node[xtick, anchor=north] at (0,\ymin-0.070) {Sequential BoN\\(N,T)};
  \node[xtick, anchor=north] at (1,\ymin-0.070) {+Tree\\Structure\\{[UCT]}};
  \node[xtick, anchor=north] at (2,\ymin-0.070) {+Search Prior\\{[PUCT]}};
  \node[xtick, anchor=north] at (3,\ymin-0.070) {+Depth\\(cumulative)};
  \node[xtick, anchor=north] at (4,\ymin-0.070) {+Multiple Parents\\{[\textbf{TTT-Discover}]}};
\end{tikzpicture}%
}
\end{minipage}
\end{tabular}
  \caption{\textbf{Cross-pair generalization}}
  \label{fig:progression_delta}
\end{subfigure}

\caption{\textbf{Harnesses in discovery systems are composite recipes whose ingredients do not transfer uniformly.}
Panel~(a) shows how we decompose two representative discovery harness families into building blocks, diverging from the same root of greedy Sequential Best-of-N. Panel~(b) summarizes the cross-pair majority-win analysis across all 12 model--problem pairs used in this paper, using Sequential BoN as the baseline. Moving from Sequential BoN to lightweight archive exploration or UCT/PUCT-style search can improve aggregate performance across our suite, but adding the full OpenEvolve-style or TTT-Discover style machinery does not consistently help and can reverse gains obtained by simpler intermediate variants.}
\label{fig:search_and_config}
\end{figure*}

Answering this question requires repeated, budget-matched evaluation that can separate genuine harness effects from run-to-run variance. We first decompose two representative harness families: OpenEvolve-style evolutionary search \citep{openevolve} and TTT-Discover-style PUCT search\footnote{For TTT-Discover, we ablate the search harness, not the test-time-training component.} \citep{yuksekgonul2026learning}. Starting from a Sequential Best-of-$N$ (BoN) baseline, we rebuild these harnesses one component at a time (Figure \ref{fig:search_progression}). We evaluate more than 3.1 million LLM rollouts across 30 harnesses, covering 12 model--problem pairs across models ranging from 3B to 120B parameters. All comparisons are budget matched by rollout counts, and we use bootstrap statistical tests against a repeated Sequential BoN baseline\footnote{We create a 100-run baseline pool for Qwen models, and a 30-run baseline pool for GPT-OSS models.} together with a cross-pair analysis to distinguish robust improvements from isolated lucky runs. To make this evaluation protocol reusable beyond this paper, we will release the complete run pools behind our study — over 3.1 million evaluated rollouts with per-step evaluator scores, so that our analyses can be reproduced exactly and future harness proposals can be compared against these empirical reference distributions. 

We find that \textbf{harnesses for discovery systems have a generalization problem -- no single fixed harness is best across all model--problem pairs}, and surprisingly, more complex OpenEvolve-style harnesses generally rank below simpler lightweight-exploration and PUCT-style variants in our cross-pair analysis (Figure \ref{fig:progression_delta}, Section \ref{sec:config_battle}). As each harness excels in different settings, we propose to \textbf{treat harness choice as a model- and problem-dependent hyperparameter rather than a universally transferable recipe in automated discovery}. This however creates a practical dilemma when given a new problem -- how much budget should be spent repeating a chosen harness to obtain a strong result, and how much budget should be spent on exploring alternative harnesses? This motivates a budget-matched adaptive-allocation experiment (Section \ref{sec:ensembleevolve}). A key finding we use is that early discovery progress is strongly correlated with final performance (Section \ref{sec:predictive_property}), using which we show that instead of choosing one harness upfront, we can run several harnesses partially, use partial-run evaluator scores to prune less promising runs and reallocate their remaining compute to stronger survivors. Our experiments therefore identify two empirical conditions that motivate adaptive allocation: fixed-harness performance varies across model--problem pairs, and partial-run evaluator feedback is informative enough to guide allocation. Together, \textbf{these findings motivate treating harness choice as an online, model- and problem-dependent decision guided by partial-run performance, rather than as a fixed choice made before search begins.}

\textbf{Contributions.} Our main contributions are:

\begin{enumerate}
  \item To our knowledge, we provide the most extensive statistically controlled evaluation of discovery harnesses to date, using more than 3.1 million rollouts to study 30 budget-matched harnesses across 12 model–problem pairs, with repeated stochastic trials and pair-level and cross-pair statistical tests.

  \item We show that no fixed discovery harness is universally superior across the evaluated model--problem pairs, revealing a generalization problem in discovery harnesses.

  \item We show that choosing among harnesses online as rollouts progress alleviates this generalization problem, motivating discovery systems that adapt their harnesses during problem solving.

  \item We release over 3.1 million rollout records, including per-step evaluator scores and baseline null distributions, as reusable statistical infrastructure for evaluating future discovery harnesses.
\end{enumerate}

\section{A Unified View of Harnesses for Discovery Systems}\label{sec:unified-view-of-search-harnesses}
Harnesses in discovery systems can be viewed as a prescribed set of rules for choosing which previously evaluated solutions or programs to expand upon. In this section, we present a unified view of discovery harnesses, focusing on two popular systems -- OpenEvolve \citep{openevolve}, based on evolutionary search, and TTT-Discover \citep{yuksekgonul2026learning}, which uses PUCT-based parent sampling \citep{silver2017mastering}. To study these, we start from a greedy sequential best-of-N baseline, and systematically add one design decision at a time to move it towards either the OpenEvolve or TTT-Discover-style search harness. For TTT-Discover, we focus only on the search harness and not the test-time-training component.

Let $x$ denote a candidate program and $S(x)$ the score returned by the evaluator. At iteration $t$, let $\mathcal{H}_t$ denote the full history of evaluated programs. We define the top-$K$ elite view of this history as $\mathcal{E}^{K}_{t} = \operatorname{TopK}(\mathcal{H}_t, K; S)$ where the $\operatorname{TopK}$ operator returns $K$ programs in $\mathcal{H}_t$ with highest evaluator scores. For a finite archive $\mathcal{A}$, we write $x \sim \mathcal{A}$ to denote uniform sampling of a program from that archive. At each iteration, the harness samples a program from an archive to serve as the parent and generates $N$ candidate children $x_{t,1},\ldots,x_{t,N}$ from this parent. The evaluator scores each child and the full history is updated by appending the evaluated children to $\mathcal{H}_t$. For a fixed number of iterations $T$, the total rollout budget is $B=NT$. In our experiments, we keep the total rollout budget constant while evaluating different harnesses.

\textbf{The Baseline: Greedy Sequential Best-of-N.} The greedy sequential BoN search establishes the baseline for our study. Starting from an initial program, the harness repeatedly mutates only the best program discovered so far. Equivalently, the parent $p_t$ at iteration $t$ can be written as

\begin{equation}
\label{eq:sequential-bon-parent}
p_t = \arg\max_{x \in \mathcal{H}_t} S(x).
\end{equation}

% \begin{equation}
% \label{eq:sequential-bon-top-one}
% \mathcal{E}^{1}_{t} = \operatorname{TopK}(\mathcal{H}_t, 1; S),
% \qquad
% p_t \sim \mathcal{E}^{1}_{t}.
% \end{equation}

Using the TopK-view notation above, the same process can be written as sampling from a top-1 archive, or $p_t \sim \mathcal{E}^{1}_{t}$, equivalent to deterministically selecting the current best solution. The model then generates $N$ candidate modifications from this parent, evaluates them, and appends them to $\mathcal{H}_{t+1}$. In subsequent sections, we show how OpenEvolve and TTT-Discover harnesses can be derived from greedy sequential BoN with modified scoring functions or sampling archives. This, along with the simplicity of Sequential BoN, makes it a useful baseline for isolating contributions of more complicated harnesses.

\textbf{From Sequential BoN to OpenEvolve.} The OpenEvolve harness relaxes the greedy parent selection in sequential BoN in four steps. First, the top-1 archive is replaced with a top-$K$ archive. Second, epsilon-greedy exploration is added by occasionally sampling from the full history $\mathcal{H}_t$ instead of only from the top-K archive. Third, the fixed generation budget is shifted from breadth to depth by decreasing $N$ and increasing $T$. Finally, OpenEvolve includes \emph{inspiration sampling}, placing additional prior programs in-context rather than just the parent program, MAP-Elites to maintain diversity in the inspirations, and multi-island evolution where multiple search processes happen in parallel along with crossovers. This progression is summarized in Figure \ref{fig:search_progression}. We present experiments for each of the above design decisions that get added on top of Sequential BoN to create OpenEvolve. For a more detailed formulation, we refer the reader to Appendix \ref{app:bon-to-openevolve}.

\textbf{From Sequential BoN to TTT-Discover.} Sequential BoN can be seen as the greediest tree policy where parent selection happens without taking into account the sub-tree of a node. Three specific changes are made to Sequential BoN. First, the scoring function for a program $x$ in equation \ref{eq:sequential-bon-parent} is made a value estimate of the entire subtree rooted at $x$. Second, a UCT (Upper Confidence Bound) based exploration bonus is added for each node based on visitation count. Finally, this exploration bonus is given a prior estimate term, modifying the bonus to a PUCT based rule. The TTT-Discover harness also samples multiple parents from the full program history at the same time step instead of sampling just one parent. This progression is also shown in Figure \ref{fig:search_progression}. For more detailed formulation, we refer the reader to Appendix \ref{app:bon-to-tttdiscovery}.

\section{Experimental Setup and Statistical Evaluation}
We evaluate harnesses at two complementary levels. First, for each model--problem pair, we test whether a candidate harness produces an unusually strong best-of-five outcome relative to a budget-matched Sequential BoN reference distribution. Second, we aggregate these outcomes across all 12 pairs to measure whether the same fixed harness transfers broadly. Section~\ref{sec:experiments} uses the pair-level analysis to study individual harness components and the cross-pair analysis to compare fixed harnesses across the full evaluation suite.

\subsection{Models, Problems and Budgets}
%We systematically build towards the OpenEvolve and TTT-Discover harnesses from a greedy sequential BoN baseline. All harnesses are compared under a fixed-budget.

\textbf{Models.} We perform discovery experiments using the following models: Qwen2.5-3B-Instruct~\citep{qwen2025qwen25technicalreport}, Qwen3-4B-Instruct-2507~\citep{yang2025qwen3}, GPT-OSS-20B and GPT-OSS-120B~\citep{agarwal2025gpt}. The models span sizes from 3B to 120B parameters, allowing us to study whether harness behavior is consistent across model scales and capabilities.

\textbf{Problems.} We evaluate each search configuration on three mathematical discovery tasks: circle packing, Heilbronn triangle, and second autocorrelation inequality \citep{openevolve}. These tasks span a range of problem structures, from visually interpretable geometric optimization to a more abstract inequality-based discovery problem. More details on task descriptions are provided in Appendix~\ref{app:task-details}.%\lc{is it based on someone's code or something like that that we should cite in text?}

\textbf{Rollout Budgets.} For each model--problem pair, all harnesses are evaluated under the same per-run rollout budget, although they may allocate that budget differently between breadth and depth. For Qwen2.5-3B-Instruct and Qwen3-4B-Instruct-2507, each independent discovery run uses 1600 LLM rollouts. The Sequential BoN baseline realizes this budget as 100 iterations with 16 children generated per iteration. For GPT-OSS-20B and GPT-OSS-120B, each run uses 320 and 160 rollouts, respectively, realized in Sequential BoN as 40 iterations with 8 and 4 children per iteration. We use the smaller rollout budget for GPT-OSS-120B to keep the inference cost of the larger model tractable. These budgets are matched within each model--problem pair, not across different models. Each harness is evaluated with five independent runs. To construct the Sequential BoN reference pool, we use 100 runs for each Qwen model--problem pairs and 30 runs for each GPT-OSS model--problem pairs.

\subsection{Best-of-Five Significance Test Against Sequential BoN}\label{sec:null-hypothesis-testing-under-a-fixed-search-budget}

We first ask whether a candidate harness produces an observed best-of-five outcome that is unusually strong for a particular model--problem pair. We treat the repeated Sequential BoN runs as an empirical reference distribution and compare the observed candidate maximum against an empirical distribution of Sequential BoN best-of-five maxima constructed by repeatedly resampling five baseline runs and taking their maximum\footnote{Our five-run choice is larger than the common three-run protocol used in prior work \citep{cemri2026adaevolve, wang2025thetaevolve, liu2026evox, assumpccao2025codeevolve}}. For a given model--problem pair $g$ and one candidate harness $h$, let $\mathbf{s}^{0} = (s_1^{0},\ldots,s_{n_0}^{0})$ denote the final scores from a pool of $n_0$ sequential BoN baseline runs, where $n_0$ is either $100$ or $30$ depending on the model. $\mathbf{s}^{h} = (s_1^{h},\ldots,s_5^{h})$ denotes five independent runs of the candidate harness $h$. The baseline scores define an empirical null distribution $F_{0}$, to which we compare the samples of other harnesses.

For any five-score tuple $\mathbf{s}=(s_1,\ldots,s_5)$, $M(\mathbf{s})$ represents the max operator on the tuple. We use the best-of-five maximum $T_{h} =M(\mathbf{s}^{h})$ as the test statistic. For each candidate harness, the null hypothesis is that its observed maximum $T_{h}$ could have arisen from a five-run sample of the Sequential BoN baseline. We therefore test the candidate maximum against the null distribution of five-run baseline maxima:

\begin{equation}
\begin{aligned}
  H_0 &: T_{h} \text{ is drawn from the Sequential BoN distribution of best-of-five maxima,} \\
  H_1 &: T_{h} \text{ is larger than expected under that distribution.}
\end{aligned}
\end{equation}

Since the final-score distribution is often skewed, heavy-tailed, non-continuous and task-dependent, we avoid normality or any parametric assumptions. Instead, we approximate the null distribution of the five-run maximum using bootstrap resampling from the $n_0$ baseline runs. For $R=100{,}000$ bootstrap trials, we sample a five-score tuple $\mathbf{s}_r^{0}$ with replacement from $\mathbf{s}^{0}$, compute its maximum $T_r^{0}=M(\mathbf{s}_r^{0})$, and estimate the one-sided empirical bootstrap p-value conditional on the observed baseline pool as
\begin{equation}
  \widehat{p}_h
  =
  \frac{1 + \sum_{r=1}^{R} \mathbf{1}\!\left\{T_{r}^{0} \geq T_{h}\right\}}{R+1}.
\end{equation}

Small values of $\widehat{p}_h$ indicate that the observed best-of-five outcome for harness $h$ is unusually large relative to the budget-matched Sequential BoN reference distribution for a model--problem pair. Rejecting the null therefore provides evidence that the candidate harness produced an unusually strong observed discovery outcome under the same rollout budget. Failure to reject the null indicates insufficient evidence of improvement in the given five-sample experiment. However, it does not imply that the candidate harness is incapable of producing stronger outcomes in future runs. We use a prespecified significance threshold of 0.05, and the plots in section \ref{sec:experiments} mark cases when $\hat{p}_h < 0.05$. These pair-level p-values are unadjusted and are interpreted as nominal, exploratory evidence. Analyses using the mean and median as test statistic are provided in Appendix~\ref{app:complete_per_configuration_results} along with calculated p-values for all experiments.

\subsection{Evaluating Harness Transfer Across Model--Problem Pairs}\label{sec:cross-pair-setup}

Pair-level tests in section \ref{sec:null-hypothesis-testing-under-a-fixed-search-budget} identify settings in which a harness produces an unusually strong outcome, but they do not tell us whether the same harness transfers \emph{universally} across models and problems. We therefore complement them with a cross-pair majority-win score that summarizes how often a fixed harness’s observed best-of-five vector beats matched Sequential BoN reference outcomes on a majority of evaluated pairs. 

Let $\mathcal{G}=\{g_1,\ldots,g_{12}\}$ denote the 12 model--problem pairs (we use 4 models and 3 tasks per model). For a candidate harness $h$, let $\mathbf{s}^{h}_{g_j}$ denote the score vector for the model--problem pair $g_j$. Using the same maximum statistic as before, for a candidate harness $h$, we define its cross-pair maximum-score vector as $\mathbf{M}^{h} = \left( M(\mathbf{s}^{h}_{g_1}),\ldots,M(\mathbf{s}^{h}_{g_{12}}) \right)$. In bootstrap trial $r$, we construct a matched Sequential BoN baseline vector $\mathbf{M}^{0,r}$. To do this, we (i) sample five runs from the $n_0$-run baseline pool with replacement for each model--problem pair, thus giving us a score vector $\mathbf{s}^{r}_{g_j}$ for each model--problem pair $g_j$, and (ii) take the max of each score vector to create the baseline vector $\mathbf{M}^{0,r}$. Let $\mathbf{1}\{\cdot\}$ define an indicator function. Rather than reporting a single bootsrap trial, we define the cross-pair majority-win statistic $\hat{P}_{maj}(h)$ over $R$ bootstrap trials, %\ps{In the equation, use variables to denote 9 (n) and 5 (floor(n/2)+1) which makes clear that you pick 5 because its more then half of 5. }
\begin{equation}\label{eq:pmaj}
  \widehat{P}_{\mathrm{maj}}(h)
  =
  \frac{1}{R}
  \sum_{r=1}^{R}
  \mathbf{1}\!\left\{
  \sum_{j=1}^{12}
  \mathbf{1}\!\left\{M^{h}_{j} > M^{0,r}_{j}\right\}
  \geq 7
  \right\},
  \qquad R=100{,}000.
\end{equation}
Thus, $P_{maj}(h)$ estimates the probability that the observed best-of-five vector for harness $h$ beats a resampled Sequential BoN best-of-five vector on at least 7 of the 12 evaluated model–problem pairs. %Larger values indicate stronger aggregate performance relative to Sequential BoN across these 12 pairs. %However, $P_{maj}(h)$ is an aggregate comparison statistic, not itself a p-value or an estimate of performance on an unseen model–problem pair.

\paragraph{Cross-pair aggregate significance test.}
Let $F_{h,g}$ and $F_{0,g}$ denote the final-score distributions of harness $h$ and Sequential BoN, respectively, for pair $g$. We use \(T_h=\hat{P}_{\mathrm{maj}}(h)\) as the test statistic and test
\[
H_0:\ F_{h,g}=F_{0,g}\ \forall g\in G,
\qquad
H_1:\ T_h\ \text{is larger than expected under }H_0.
\]
Under \(H_0\), candidate-harness and Sequential BoN scores are exchangeable within each model--problem pair. In each permutation trial \(r\), we pool the five candidate scores with the corresponding Sequential BoN scores for every pair, randomly assign five pooled scores to a pseudo-candidate, treat the remainder as the pseudo-baseline, and recompute \(T_h^{*,r}\) using Equation~\ref{eq:pmaj}. The one-sided cross-pair permutation p-value is

\vspace{-2mm}
\begin{equation}
\hat{p}_{\mathrm{cross}}(h)
=
\frac{
1+\sum_{r=1}^{R}
\mathbf{1}\!\left\{T_h^{*,r}\ge T_h\right\}
}{
R+1
},
\label{eq:global-permutation-pvalue}
\end{equation}

We use $R=100,000$ Monte Carlo trials for both the bootstrap and permutation procedures. We apply Holm’s step-down correction to the raw cross-pair p-values across all evaluated harness configurations. Small values indicate that a baseline-equivalent pseudo-candidate rarely attains a majority-win statistic at least as large as the observed harness. We use a significance threshold of \(0.05\). We refer to $\hat{p}_{\mathrm{cross}}(h)(h)<0.05$ before correction as nominal significance and reserve claims of cross-harness significance for Holm-adjusted $\hat{p}_{\mathrm{cross}}(h)(h)<0.05$. From this point when we discuss significance we mean a cross-pair significance, unless the context is on a specific model-task pair in which we refer to the aforementioned significance.

\begin{figure}[t]
  \centering
  \includegraphics[width=0.92\linewidth]{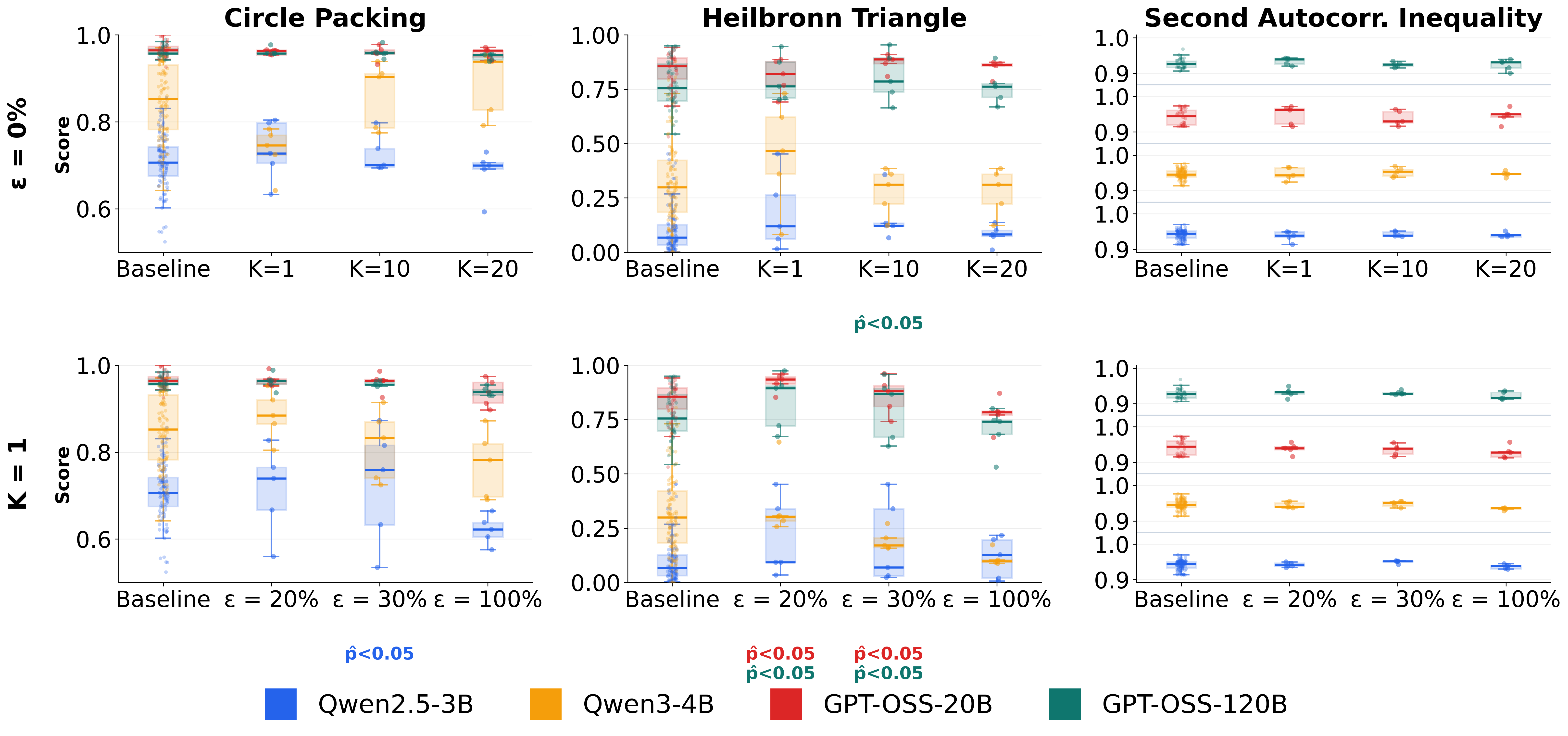}
  \caption{\textbf{Archive and Exploration Ablations.} The top row varies only Top-$K$ archive size with $\epsilon = 0\%$; the bottom row fixes $K=1$ and varies exploration probability $\epsilon$. Nominal pair-level results with $\hat{p}_h<0.05$ are indicated at the bottom of the plot. Larger archives alone yield no significant gains, while exploration helps some model--problem pairs but hurts others.} %.\lc{why is the rightmost graphs different? what is the *.049? the <0.01 is that the only place where something is significant? Need to explain this better (state something in the caption or somewhere else)}}
  \label{fig:main_archive_exploration_ablation_summary}
\end{figure}

\begin{figure}[t]
  \centering
  \includegraphics[width=0.91\linewidth]{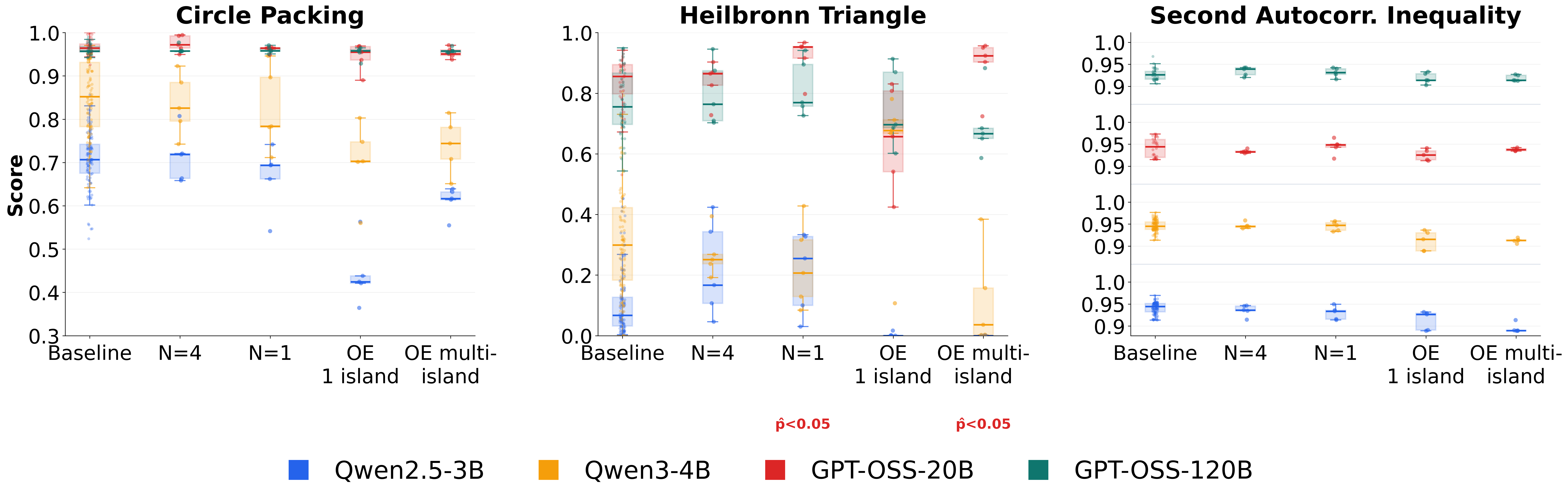}
  \caption{\textbf{Progression toward OpenEvolve-style search.} Configurations progressively shift budget from breadth to depth and add MAP-Elites, inspiration sampling, and multiple islands. Added complexity does not produce monotonic gains, and full OpenEvolve-style variants often underperform simpler intermediates.}
  \label{fig:main_openevolve_progression_summary}
\end{figure}

\begin{figure}[t]
  \centering
  \includegraphics[width=0.91\linewidth]{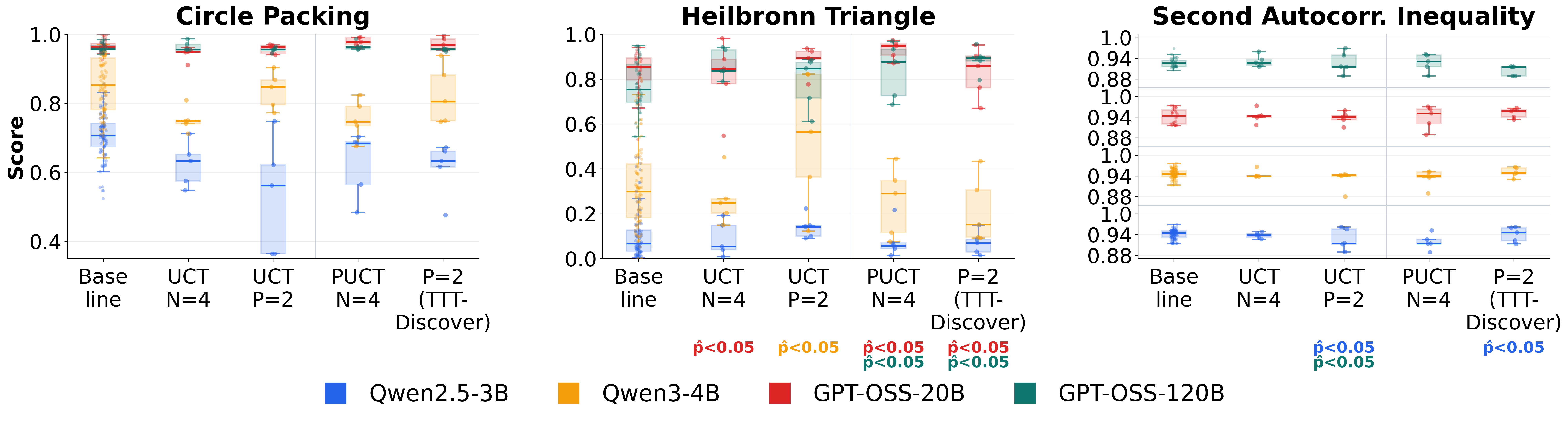}
  \caption{\textbf{TTT-Discover-style ablations.} Configurations vary tree-search policy, branching width, and the number of parents expanded per iteration. Tree search improves selected pairs, especially Heilbronn Triangle, but later TTT-Discover-style additions do not consistently help. }
  \label{fig:main_uct_puct_summary}
\end{figure}

\section{Fixed Harnesses Do Not Generalize Uniformly Across Model--Problem Pairs}\label{sec:experiments}
The analyses described in the previous section answer complementary questions: the pair-level test (Section \ref{sec:ablating-harness-components}), identifies model–problem pairs where a harness significantly outperforms Sequential BoN, whereas the cross-pair test (Section \ref{sec:config_battle}), asks whether a fixed harness is significantly better overall across all 12 pairs.

\subsection{Ablating Harness Components}\label{sec:ablating-harness-components}
\paragraph{Archive and Exploration Choices.} Figure~\ref{fig:main_archive_exploration_ablation_summary} examines two lightweight ways of adding exploration to Sequential BoN. $\epsilon$ denotes the probability of sampling a parent from the full search history $\mathcal{H}_t$. The top row fixes $\epsilon=0\%$ and varies the Top-$K$ archive size, allowing the search to revisit multiple high-scoring solutions rather than always expanding the current best. None of the larger Top-$K$ configurations produces a best-of-five outcome that has a pair-level $\hat{p}_h < 0.05$ against the Sequential BoN reference. The bottom row fixes $K=1$ and varies $\epsilon$. Full-history exploration produces nominally significant gains for select model--problem pairs, including Qwen2.5-3B on Circle Packing and both GPT-OSS models on Heilbronn Triangle, but scores lower for other pairs and is not universally significant as we will see in the cross-pair analysis in section \ref{sec:config_battle}. Overall, no harness improves performance for all problems given a model or all models on a specific problem. We additionally vary $K$ and $\epsilon$ jointly in Appendix~\ref{sec:detailed_search_harness_ablations}, where we similarly see that lightweight exploration can help in selected settings, but its benefits do not transfer uniformly across model--problem pairs.

\paragraph{From Sequential BoN to OpenEvolve.} We next ask whether progressively adding the components of the OpenEvolve recipe improves discovery under a fixed rollout budget. Starting from the best lightweight exploration configuration, we first shift compute from breadth to depth, then add inspiration sampling with MAP-Elites, followed by multi-island evolution. Figure~\ref{fig:progression_delta} summarizes the cumulative cross-pair trend, while Figure~\ref{fig:main_openevolve_progression_summary} shows the corresponding score distributions for each model--problem pair. Additional components do not produce monotonic gains: the full OpenEvolve-style configurations produce nominally significant gains only for GPT-OSS-20B on Heilbronn Triangle, while often reducing performance on Circle Packing and the Second Autocorrelation Inequality. Moreover, the final OpenEvolve configurations are frequently worse than intermediate, simpler variants. \textbf{Overall, adding more OpenEvolve machinery does not consistently improve transfer, and the complete recipe underperforms the baselines that strip its components away.} We describe the cross-pair effect of the OpenEvolve harness in section \ref{sec:config_battle}.

\paragraph{From Sequential BoN to TTT-Discover Search.}
We see that adding evolutionary machinery in OpenEvolve does not produce monotonic gains. We next follow the tree-search branch of Figure~\ref{fig:search_progression}, progressively replacing greedy parent selection with UCT, adding the prior-weighted exploration term (PUCT), shifting the fixed rollout budget toward deeper search, and finally expanding multiple parents per iteration as in the search component of TTT-Discover. Figure~\ref{fig:progression_delta} summarizes the cumulative cross-pair effects, while Figure~\ref{fig:main_uct_puct_summary} shows the corresponding pair-level score distributions. At the pair level, the largest observed gains occur after introducing UCT/PUCT-based parent selection. Subsequent changes to search depth and multi-parent expansion do not improve performance consistently and sometimes reverse those gains.

\subsection{Choosing One Fixed Harness}\label{sec:config_battle}

The component ablations show that harness choices can produce significant gains for individual model--problem pairs while hurting others.  We now ask a deployment-oriented question: \emph{if one fixed harness must be selected without knowing in advance the configuration that is best for a given model--problem pair, which one should it be?} Or in other words - which harness choice performs most consistently across the 12 evaluated pairs? We use the cross-pair permutation p-value ($\hat{p}_{\mathrm{cross}}$) to do so, as defined in Section~\ref{sec:cross-pair-setup}.

Figure \ref{fig:config_battle} ranks representative fixed harnesses by \(\hat{P}_{\mathrm{maj}}\), with cross-pair p-values before and after Holm's correction. The strongest observed configuration is an epsilon-greedy-style harness with $K=1$ and $\epsilon=20\%$, where it samples a parent from $\mathcal{H}_t$ $20\%$ of the time, and otherwise uses the best solution so far as parent, which has \(\hat{P}_{\mathrm{maj}}=0.914\) and \(\hat{p}_{\mathrm{cross}}=0.023\). However after Holm's correction the p-value rises to \(\hat{p}^{holm}_{\mathrm{cross}}=0.678\) and is no longer significant. In contrast, the complete OpenEvolve-style configurations have the lowest observed majority-win score in the evaluated suite with \(\hat{P}_{\mathrm{maj}}=0.033\) and \(\hat{p}_{\mathrm{cross}}=0.978\). . The full results for all harnesses can be found in Appendix \ref{app:full_battle}.

Together with the pair-level results in Section~\ref{sec:ablating-harness-components}, these findings show that the strongest observed fixed choice changes across models and problems. Lightweight exploration methods and simpler tree-search algorithms have stronger observed cross-pair performance than the complete OpenEvolve-style recipes, but the evidence does not support replacing one universal harness recipe with another.

\findingbox{\textbf{Finding 1.} No tested fixed harness is significantly better than a simple baseline in our evaluation suite, and the strongest observed harness varies across settings.}

\section{Early Feedback Enables Online Harness Allocation}
\label{sec:predictive_analysis}

Section \ref{sec:experiments} shows that the strongest fixed harness varies across the evaluated model--problem pairs, making it risky to commit the entire discovery budget to one configuration in advance. We therefore ask whether harness choice can instead be made during the discovery process. Such an approach requires early evaluator feedback to be informative about final performance. We first test whether partial-run performance can provide this signal and then use it in a budget-matched online allocation policy.

\subsection{Early Discovery Progress Predicts Final Performance}\label{sec:predictive_property}
For each discovery run, we record the best score reached after 10\%, 25\%, and 50\% of its allocated steps and compute the Spearman rank correlation between partial-run and final performance. This measures whether configurations that lead early in the search also tend to finish among the strongest configurations.

\begin{figure*}[t]
\centering
\newcommand{\figurefivepanelheight}{0.225\textheight}
\begin{subfigure}[t]{0.59\textwidth}
  \centering
  \def\configbattlepanelheight{\figurefivepanelheight}%
  \definecolor{battleblue}{HTML}{2F6FD6}
\definecolor{battlebluefill}{HTML}{DCEAFF}
\definecolor{battlegreen}{HTML}{3D7F2A}
\definecolor{battlegreenfill}{HTML}{E7F4DE}
\definecolor{battleorange}{HTML}{F08A00}
\definecolor{battleorangefill}{HTML}{FFF0D6}
\definecolor{battlepurple}{HTML}{6B4AA0}
\definecolor{battlepurplefill}{HTML}{F0EAF8}
\definecolor{battlered}{HTML}{B3261E}
\definecolor{battleredfill}{HTML}{FCE8E6}
\definecolor{battleslate}{HTML}{33403E}
\definecolor{battlegray}{HTML}{63736F}
\definecolor{battleborder}{HTML}{CFE2DE}
\definecolor{battlepval}{HTML}{5B7896}
\definecolor{battleholm}{HTML}{C46A1A}

\providecommand{\configbattlepanelheight}{!}
\resizebox{!}{\configbattlepanelheight}{%
\begin{tikzpicture}[
  panel/.style={rounded corners=7pt, draw=battleborder, fill=white,
                line width=0.9pt, minimum width=15.4cm, minimum height=9.95cm,
                anchor=north west},
  ptitle/.style={font=\LARGE\bfseries, text=battleslate, align=center},
  axis/.style={draw=battlegray, line width=0.5pt},
  label/.style={font=\small, text=battleslate, anchor=east, align=right},
  value/.style={font=\normalsize\bfseries, text=battleslate, anchor=east, fill=white, inner sep=1.5pt},
  pvalue/.style={font=\normalsize\bfseries, anchor=east, fill=white, inner sep=1.5pt},
  pheader/.style={font=\normalsize\bfseries, anchor=east, align=center},
  tick/.style={font=\footnotesize, text=battlegray, align=center},
  note/.style={font=\small, text=battlegray, align=center},
  elitebar/.style={fill=battlegreenfill, fill opacity=0.7, draw=battlegreen, line width=0.75pt},
  allocbar/.style={fill=battlepurplefill, fill opacity=0.7, draw=battlepurple, line width=0.75pt},
  puctbar/.style={fill=battlebluefill, fill opacity=0.7, draw=battleblue, line width=0.75pt},
  uctbar/.style={fill=battleorangefill, fill opacity=0.7, draw=battleorange, line width=0.75pt},
  openbar/.style={fill=battleredfill, fill opacity=0.7, draw=battlered, line width=0.75pt},
]

\node[panel] (p) at (0,0) {};
\begin{scope}[shift={(p.north west)}]
  \def\plotxshift{0.95}

  \node[ptitle] at (7.0,-0.45) {Max Statistic};

  % Axis at the bottom
  \draw[axis] ({3.7+\plotxshift},-8.15) -- ({9.85+\plotxshift},-8.15);
  \foreach \x/\t in {3.7/0,6.775/.5,9.85/1.0} {
    \draw[axis] ({\x+\plotxshift},-8.08) -- ({\x+\plotxshift},-8.22);
    \node[tick] at ({\x+\plotxshift},-8.42) {\t};
  }
  \node[note] at ({6.775+\plotxshift},-8.78) {Battle Win Rate};
  \node[pheader, text=battleslate] at ({10.65+\plotxshift},-0.62) {$\hat{P}_{\mathrm{maj}}$};
  \node[pheader, text=battlepval] at ({11.85+\plotxshift},-0.62) {$\hat{p}_{\text{cross}}$};
  \node[pheader, text=battleholm] at ({13.40+\plotxshift},-0.62) {$\hat{p}^{\text{holm}}_{\text{cross}}$};

  % --- Selected configs sorted by observed p_maj (max statistic) ---
  % Bar right edge = 3.7 + observed_pmaj * 6.15

  % source: elite:E1:X80
  \node[label] at ({3.55+\plotxshift},-1.10) {$K{=}1, \epsilon{=}20\%$};
  \filldraw[elitebar] ({3.7+\plotxshift},-1.30) rectangle ({9.321+\plotxshift},-0.90);
  \node[value] at ({10.65+\plotxshift},-1.10) {.914};
  \node[pvalue, text=battlepval] at ({11.85+\plotxshift},-1.10) {.023};
  \node[pvalue, text=battleholm] at ({13.40+\plotxshift},-1.10) {.678};

  % source: puct_uct:c_ablation:PUCT:C1
  \node[label] at ({3.55+\plotxshift},-1.78) {PUCT, $C{=}1, N{=}16, P{=}1$};
  \filldraw[puctbar] ({3.7+\plotxshift},-1.98) rectangle ({9.095+\plotxshift},-1.58);
  \node[value] at ({10.65+\plotxshift},-1.78) {.877};
  \node[pvalue, text=battlepval] at ({11.85+\plotxshift},-1.78) {.085};
  \node[pvalue, text=battleholm] at ({13.40+\plotxshift},-1.78) {1.000};

  % source: puct_uct:c_ablation:PUCT:C10
  \node[label] at ({3.55+\plotxshift},-2.46) {PUCT, $C{=}10, N{=}16, P{=}1$};
  \filldraw[puctbar] ({3.7+\plotxshift},-2.66) rectangle ({8.156+\plotxshift},-2.26);
  \node[value] at ({10.65+\plotxshift},-2.46) {.725};
  \node[pvalue, text=battlepval] at ({11.85+\plotxshift},-2.46) {.252};
  \node[pvalue, text=battleholm] at ({13.40+\plotxshift},-2.46) {1.000};

  % source: puct_uct:b_ablation:PUCT:half_b
  \node[label] at ({3.55+\plotxshift},-3.14) {PUCT, Halved $N$};
  \filldraw[puctbar] ({3.7+\plotxshift},-3.34) rectangle ({8.029+\plotxshift},-2.94);
  \node[value] at ({10.65+\plotxshift},-3.14) {.704};
  \node[pvalue, text=battlepval] at ({11.85+\plotxshift},-3.14) {.162};
  \node[pvalue, text=battleholm] at ({13.40+\plotxshift},-3.14) {1.000};

  % source: puct_uct:p_ablation:UCT:P4
  \node[label] at ({3.55+\plotxshift},-3.82) {UCT, $P{=}4, N{=}2$};
  \filldraw[uctbar] ({3.7+\plotxshift},-4.02) rectangle ({7.715+\plotxshift},-3.62);
  \node[value] at ({10.65+\plotxshift},-3.82) {.653};
  \node[pvalue, text=battlepval] at ({11.85+\plotxshift},-3.82) {.210};
  \node[pvalue, text=battleholm] at ({13.40+\plotxshift},-3.82) {1.000};

  % source: puct_uct:c_ablation:UCT:C0.1
  \node[label] at ({3.55+\plotxshift},-4.50) {UCT, $C{=}0.1, N{=}16, P{=}1$};
  \filldraw[uctbar] ({3.7+\plotxshift},-4.70) rectangle ({7.659+\plotxshift},-4.30);
  \node[value] at ({10.65+\plotxshift},-4.50) {.644};
  \node[pvalue, text=battlepval] at ({11.85+\plotxshift},-4.50) {.352};
  \node[pvalue, text=battleholm] at ({13.40+\plotxshift},-4.50) {1.000};

  % source: puct_uct:b_ablation:UCT:half_b
  \node[label] at ({3.55+\plotxshift},-5.18) {UCT, Halved $N$};
  \filldraw[uctbar] ({3.7+\plotxshift},-5.38) rectangle ({7.093+\plotxshift},-4.98);
  \node[value] at ({10.65+\plotxshift},-5.18) {.552};
  \node[pvalue, text=battlepval] at ({11.85+\plotxshift},-5.18) {.313};
  \node[pvalue, text=battleholm] at ({13.40+\plotxshift},-5.18) {1.000};

  % source: puct_uct:c_ablation:PUCT:C0.1
  \node[label] at ({3.55+\plotxshift},-5.86) {PUCT, $C{=}0.1, N{=}16, P{=}1$};
  \filldraw[puctbar] ({3.7+\plotxshift},-6.06) rectangle ({6.643+\plotxshift},-5.66);
  \node[value] at ({10.65+\plotxshift},-5.86) {.479};
  \node[pvalue, text=battlepval] at ({11.85+\plotxshift},-5.86) {.558};
  \node[pvalue, text=battleholm] at ({13.40+\plotxshift},-5.86) {1.000};

  % source: puct_uct:p_ablation:PUCT:P4
  \node[label] at ({3.55+\plotxshift},-6.54) {PUCT, $P{=}4, N{=}2$};
  \filldraw[puctbar] ({3.7+\plotxshift},-6.74) rectangle ({6.564+\plotxshift},-6.34);
  \node[value] at ({10.65+\plotxshift},-6.54) {.466};
  \node[pvalue, text=battlepval] at ({11.85+\plotxshift},-6.54) {.411};
  \node[pvalue, text=battleholm] at ({13.40+\plotxshift},-6.54) {1.000};

  % source: openevolve:generic_1_island
  \node[label, text=battlered] at ({3.55+\plotxshift},-7.22) {OpenEvolve};
  \filldraw[openbar] ({3.7+\plotxshift},-7.42) rectangle ({3.905+\plotxshift},-7.02);
  \node[value] at ({10.65+\plotxshift},-7.22) {.033};
  \node[pvalue, text=battlepval] at ({11.85+\plotxshift},-7.22) {.978};
  \node[pvalue, text=battleholm] at ({13.40+\plotxshift},-7.22) {1.000};

  % Legend -- compact footer row
  \node[font=\small, text=battlegreen, anchor=west] at ({2.40+\plotxshift},-9.40) {\rule{0.65em}{0.65em}\ TopK/$\epsilon$};
  \node[font=\small, text=battlepurple, anchor=west] at ({4.45+\plotxshift},-9.40) {\rule{0.65em}{0.65em}\ Breadth--Depth};
  \node[font=\small, text=battleblue, anchor=west] at ({7.20+\plotxshift},-9.40) {\rule{0.65em}{0.65em}\ PUCT};
  \node[font=\small, text=battleorange, anchor=west] at ({8.75+\plotxshift},-9.40) {\rule{0.65em}{0.65em}\ UCT};
  \node[font=\small, text=battlered, anchor=west] at ({10.15+\plotxshift},-9.40) {\rule{0.65em}{0.65em}\ OpenEvolve};
\end{scope}

\end{tikzpicture}%
}
  \caption{\textbf{Cross-pair generalization.}}
  \label{fig:config_battle}
\end{subfigure}
\hfill
\begin{subfigure}[t]{0.40\textwidth}
  \centering
  \includegraphics[height=\figurefivepanelheight]{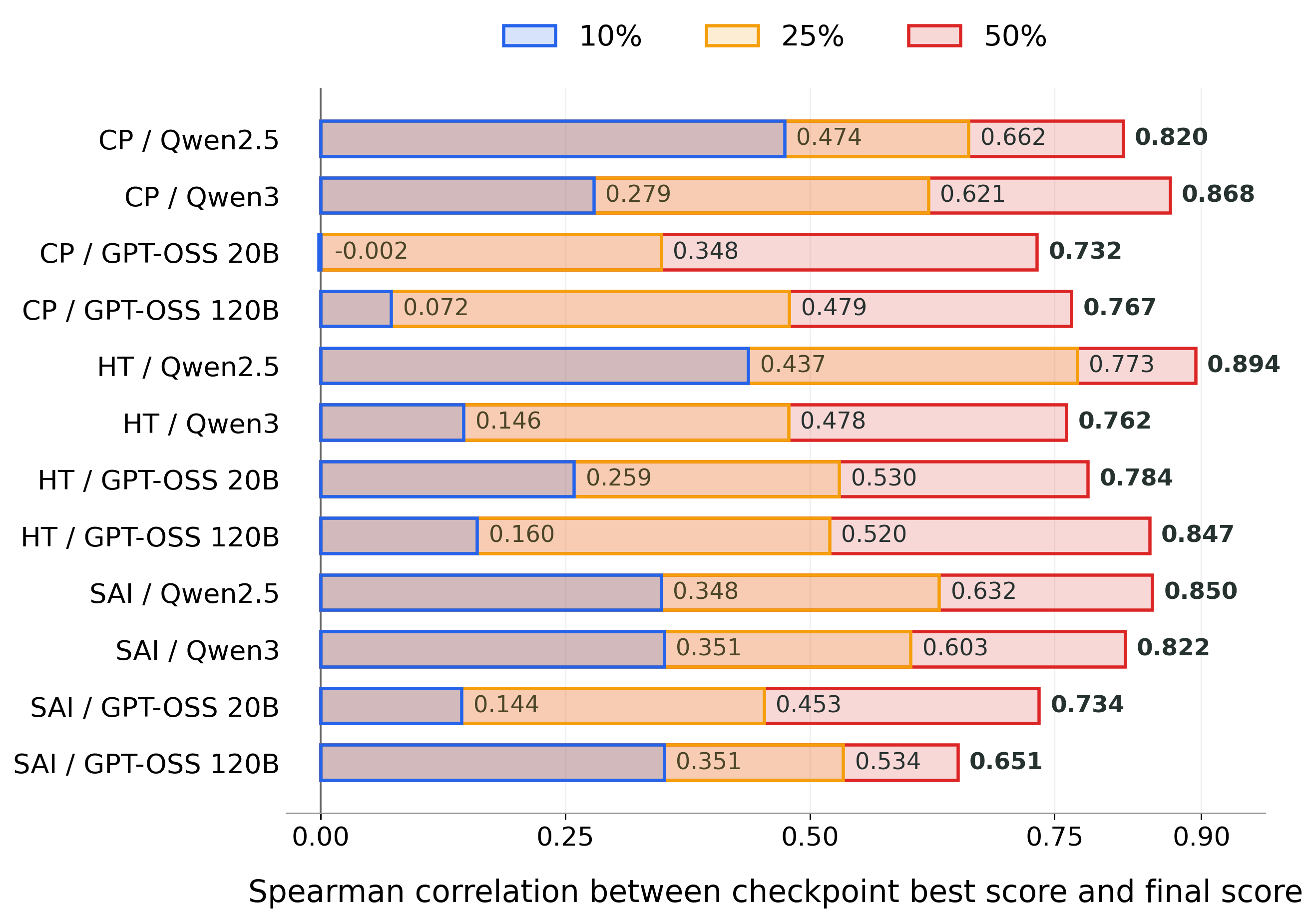}%
  \caption{\textbf{Early predictiveness.}}
  \label{fig:predictive_core_stats_by_checkpoint}
\end{subfigure}

\caption{\textbf{Fixed harnesses do not reliably generalize, but early progress is informative.}
Panel (a) shows the cross-pair majority-win statistic \(\hat{P}_{\mathrm{maj}}\) for representative fixed harnesses across all 12 model--problem pairs along with the cross-pair p-values before and after Holm's correction. Panel (b) shows checkpoint-to-final rank correlation, motivating early-feedback-based allocation.}
\label{fig:generalization_and_predictiveness}
\end{figure*}
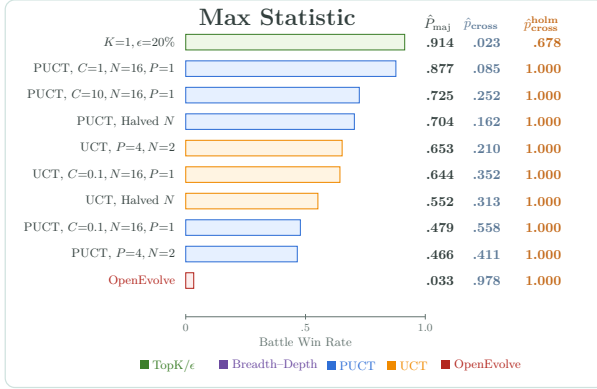
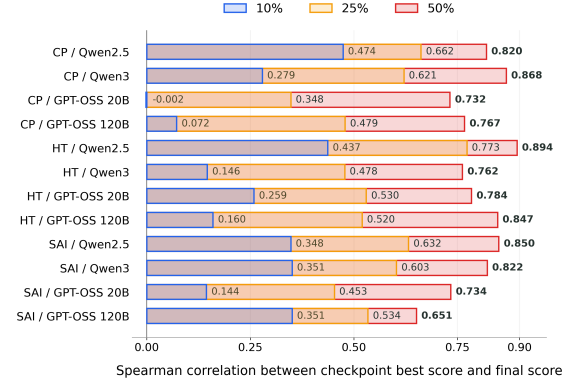

\textbf{Results.} The results can be seen in Figure \ref{fig:predictive_core_stats_by_checkpoint}. At the 10\% checkpoint, partial-run and final performance have weak-to-moderate Spearman correlations, ranging from 0.000 to 0.474. At the 50\% checkpoint, the correlation with final performance exceeds 0.70 for 11 of the 12 model–problem pairs, and remains positive at 0.651 for GPT-OSS-120B on the Second Autocorrelation Inequality. Thus, by the halfway point, partial-run performance provides a strong signal about final outcomes. This shows that early evolution progress is strongly correlated with final performance. For adaptive allocation, the relevant question is not whether early scores perfectly predict final scores, but whether they are informative enough to rank partial runs better than chance. A moderate to strong positive Spearman correlation is already useful for pruning, because the allocation policy only needs to discard clearly weak partial trajectories and preserve promising ones. This motivates allocating the remaining budget according to partial-run performance rather than completing every sampled harness.

\findingbox{\textbf{Finding 2.} Partial-run performance becomes increasingly correlated with final performance as search progresses, providing a useful signal at early checkpoints and a strong signal by the halfway point.}

\newcommand{\ensembleevolvestrategydeviationgraphic}[2]{%
\definecolor{eeblue}{HTML}{2F6FD6}
\definecolor{eebluefill}{HTML}{DCEAFF}
\definecolor{eegreen}{HTML}{3D7F2A}
\definecolor{eegreenfill}{HTML}{E7F4DE}
\definecolor{eeorange}{HTML}{F08A00}
\definecolor{eeorangefill}{HTML}{FFF0D6}
\definecolor{eepurple}{HTML}{6B4AA0}
\definecolor{eepurplefill}{HTML}{F0EAF8}
\definecolor{eeteal}{HTML}{1B8A8F}
\definecolor{eetealfill}{HTML}{DDF4F2}
\definecolor{eerose}{HTML}{B8466B}
\definecolor{eerosefill}{HTML}{F8E4EC}
\definecolor{eeslate}{HTML}{33403E}
\definecolor{eegray}{HTML}{63736F}
\definecolor{eeborder}{HTML}{CFE2DE}

\resizebox{#1}{#2}{%
\begin{tikzpicture}[
  rowlabel/.style={font=\footnotesize\bfseries, text=eeslate, anchor=east, align=right},
  sublabel/.style={font=\scriptsize, text=eegray, anchor=east, align=right},
  tick/.style={font=\scriptsize, text=eegray, align=center},
  run/.style={rounded corners=2pt, draw=eeblue, fill=eebluefill, line width=0.7pt, minimum height=0.23cm},
  configa/.style={rounded corners=2pt, draw=eegreen, fill=eegreenfill, line width=0.7pt, minimum height=0.23cm},
  configb/.style={rounded corners=2pt, draw=eeorange, fill=eeorangefill, line width=0.7pt, minimum height=0.23cm},
  configc/.style={rounded corners=2pt, draw=eepurple, fill=eepurplefill, line width=0.7pt, minimum height=0.23cm},
  configd/.style={rounded corners=2pt, draw=eeteal, fill=eetealfill, line width=0.7pt, minimum height=0.23cm},
  confige/.style={rounded corners=2pt, draw=eerose, fill=eerosefill, line width=0.7pt, minimum height=0.23cm},
  checkpoint/.style={draw=eeslate, line width=0.7pt},
  arrow/.style={-{Stealth[length=2.0mm,width=1.6mm]}, draw=eeslate, line width=0.65pt},
  note/.style={font=\scriptsize, text=eegray, align=center},
  title/.style={font=\large\bfseries, text=eeslate, align=center}
]

\node[title] at (8.4,0.55) {Equal-Budget Evaluation Against Adaptive Harness Ensemble};

% axis
\draw[checkpoint] (4.1,-0.15) -- (15.6,-0.15);
\foreach \x/\t in {4.1/0\%,6.975/25\%,9.85/50\%,12.725/75\%,15.6/100\%} {
  \draw[checkpoint] (\x,-0.08) -- (\x,-0.25);
  \node[tick] at (\x,-0.47) {\t};
}

% fixed single config
\node[rowlabel] at (3.72,-1.18) {Single config};
\node[sublabel] at (3.72,-1.52) {5 full runs};
\foreach \y in {-0.86,-1.02,-1.18,-1.34,-1.50} {
  \node[run, minimum width=11.5cm, anchor=west] at (4.1,\y) {};
}

% fixed full portfolio
\node[rowlabel] at (3.72,-2.38) {Full portfolio};
\node[sublabel] at (3.72,-2.72) {5 configs, no pruning};
\foreach \y/\cfg in {-2.06/configa,-2.22/configb,-2.38/configc,-2.54/configd,-2.70/confige} {
  \node[\cfg, minimum width=11.5cm, anchor=west] at (4.1,\y) {};
}

% single stage
\node[rowlabel] at (3.72,-3.78) {Single-stage\\Adaptive Harness Ensemble};
\node[sublabel] at (3.72,-4.32) {many partial runs, keep 1};
\foreach \y/\cfg in {-3.22/configa,-3.36/configb,-3.50/configc,-3.64/configd,-3.78/confige,
                     -3.92/configa,-4.06/configb,-4.20/configc,-4.34/configd,-4.48/confige} {
  \node[\cfg, minimum width=2.875cm, anchor=west] at (4.1,\y) {};
}
\node[confige, minimum width=8.625cm, anchor=west] at (6.975,-3.78) {};

% multi stage
\node[rowlabel] at (3.72,-5.68) {Multi-stage\\Adaptive Harness Ensemble};
\node[sublabel] at (3.72,-6.22) {10 $\rightarrow$ 5 $\rightarrow$ 3 $\rightarrow$ 2};
\foreach \y/\cfg in {-5.06/configa,-5.18/configb,-5.30/configc,-5.42/configd,-5.54/confige,
                     -5.66/configa,-5.78/configb,-5.90/configc,-6.02/configd,-6.14/confige} {
  \node[\cfg, minimum width=2.875cm, anchor=west] at (4.1,\y) {};
}
\foreach \y/\cfg in {-5.24/configb,-5.42/configd,-5.60/confige,-5.78/configa,-5.96/configc} {
  \node[\cfg, minimum width=2.875cm, anchor=west] at (6.975,\y) {};
}
\foreach \y/\cfg in {-5.42/configd,-5.60/confige,-5.78/configc} {
  \node[\cfg, minimum width=2.875cm, anchor=west] at (9.85,\y) {};
}
\foreach \y/\cfg in {-5.51/configd,-5.69/confige} {
  \node[\cfg, minimum width=2.875cm, anchor=west] at (12.725,\y) {};
}
\draw[arrow] (6.975,-6.38) -- (6.975,-6.16);
\draw[arrow] (9.85,-6.38) -- (9.85,-6.02);
\draw[arrow] (12.725,-6.38) -- (12.725,-5.84);
\node[note] at (9.85,-6.68) {sequential pruning as checkpoint evidence improves};

\end{tikzpicture}%
}
}

\iffalse
\begin{figure*}[t]
\centering
\ensembleevolvestrategydeviationgraphic{\textwidth}{!}

\caption{\textbf{Strategy-deviation evaluation.} The comparison is budget matched across fixed repeated-run baselines, a non-adaptive full portfolio, single-stage Adaptive Harness Ensemble, and multi-stage Adaptive Harness Ensemble. Starting with the full-portfolio baseline, different hues denote distinct search configurations; in the Adaptive Harness Ensemble rows, retained configurations keep their hue after pruning. The deviation metric measures how far each strategy falls short of the best observed score for the same model--task pair.}
\label{fig:ensemble_evolve_strategy_deviation}
\end{figure*}
\fi

\vspace{-2mm}
\subsection{Budget-Matched Online Harness Allocation}\label{sec:ensembleevolve}
Section \ref{sec:predictive_property} shows that partial-run scores become informative about final performance before a discovery run is complete. This suggests an alternative to selecting one fixed harness upfront: start several plausible harnesses, observe their early progress, and spend the remaining budget on the strongest partial runs. This approach is related to algorithm portfolios \citep{gomes2001algorithm}, Successive Halving \citep{jamieson2016non}, Hyperband \citep{li2018hyperband}, and asynchronous early-stopping methods such as ASHA \citep{li2020system}, which use intermediate feedback to allocate resources across competing configurations.

\textbf{Budgeted policy design.} We instantiate this idea as an online allocation policy over the fixed discovery harnesses studied in section \ref{sec:experiments}. The policy starts multiple harness configurations, advances them to one or more checkpoints, ranks the partial runs by the best evaluator score observed so far, prunes weaker runs, and allocates the remaining compute to the survivors. We call this the \emph{adaptive harness ensemble}. The exact algorithms can be found in Appendix \ref{app:ensemble}. We measure compute in full-run equivalents and fix the total budget to $B_e=5$, matching the paper’s standard best-of-five evaluation. Every policy must therefore return one final score using no more compute than five complete discovery runs. Whereas the Sequential BoN reference spends this budget on five full runs, the adaptive harness ensemble spends part of it evaluating more harnesses partially and uses the remainder to complete fewer survivors. For example, evaluating 17 runs to the 25\% checkpoint and completing one survivor uses 17(0.25)+1(0.75)=5 full-run equivalents. Figure~\ref{fig:ensemble_evolve_strategy_deviation} (appendix) illustrates this breadth--depth tradeoff, where adaptive allocation can inspect many harnesses briefly, but doing so leaves enough budget to complete only a small number of survivors. 

\textbf{Pruning Policies.} A \emph{single-stage policy} starts $m$ partial runs, advances each to a normalized checkpoint $q \in (0,1)$, retains the $s$ survivor runs with the highest score, and completes those survivors, satisfying

\vspace{-3mm}
\begin{equation}\label{eq:single-prune}
  m q + s(1-q) \leq B_e.
\end{equation}

As the number of survivors $s$ increases, the number of starting configurations $m$ must decrease, since the left hand side of equation \ref{eq:single-prune} must sum to $B_e$. A \emph{multi-stage pruning policy} uses several pruning checkpoints. Let
$0=q_0 <q_1<\cdots<q_L=1$ denote $L-1$ normalized checkpoints at which we prune the less promising configurations and let
$m_1>m_2>\cdots>m_L$ denote the number of active runs advanced over successive checkpoint intervals, where $m_L=s$ is the number of final survivors. At stage $\ell$, $m_\ell$ represents the number of active configurations for the duration of $q_{\ell-1}$ to $q_\ell$. The budget constraint can therefore be written as

\vspace{-4mm}
\begin{equation}
  \sum_{\ell=1}^{L}
  m_\ell (q_\ell-q_{\ell-1})
  \leq B_e.
\end{equation}

In our experiments, we use the checkpoint family motivated by Section~\ref{sec:predictive_analysis}, with pruning opportunities at $25\%$, $50\%$, and $75\%$ of a full run, and compare final survivor counts $s\in\{1,2,3,4\}$.

\textbf{Baselines.} We compare the adaptive harness ensemble with three budget-matched strategies available without prior knowledge of the best harness for the current model--problem pair. \emph{Sequential BoN} performs five independent full runs and provides the common reference used throughout the paper. It is also a strong practical baseline, since the fixed harnesses evaluated in Section \ref{sec:ablating-harness-components} rarely outperform it consistently across model–problem pairs. The \emph{single-harness baseline} samples one configuration from the harness pool and spends all five runs repeating it, representing commitment to an unknown fixed harness. The \emph{unpruned harness ensemble} samples five configurations and runs each to completion, providing harness diversity without using intermediate feedback. The unpruned harness ensemble is the key control: because it draws from the same configuration pool as the adaptive ensemble, comparing the two isolates the value of using partial-run feedback to reallocate compute rather than merely trying multiple harnesses. A visual comparison between adaptive harness ensemble and the baselines is shown in Figure \ref{fig:ensemble_evolve_strategy_deviation}. We do not treat the highest-ranked configuration from Section \ref{sec:config_battle} as a policy baseline because it is a retrospective ranking result, not a budget-allocation strategy that generates a new outcome under the resampling protocol used here.

\textbf{Policy-Level Evaluation.} We evaluate each strategy by repeatedly simulating how it would spend a fresh budget of five full-run equivalents. In each trial and for each model--problem pair, we sample the required runs with replacement from the available empirical run pools, apply any prescribed pruning decisions, and record the best final score returned by the strategy. The single-harness strategy samples one configuration and repeats it for five full runs; the unpruned samples five configurations and runs each once; and adaptive harness ensemble samples partial trajectories and completes only those retained by its pruning schedule. Sequential BoN is evaluated under the same procedure using five full baseline runs. We repeat this process for 100,000 resampling trials and report the mean returned score for each model--problem pair, together with the unweighted average across all 12 pairs. These results estimate the mean outcome of each complete budget-allocation strategy under empirical resampling, rather than reporting the outcome of a single stochastic run.

\textbf{Results.} Table~\ref{tab:expanded_strategy_deviation} shows a clear ordering under the same five-full-run-equivalent budget; the vector-battle resampling test used to mark statistically significant averages is described in Appendix~\ref{app:full_budget_matched_allocation_results}. The single-harness baseline achieves an average score of 82.49\%, while the unpruned ensemble improves slightly to 84.54\% by introducing harness diversity. Every evaluated adaptive pruning schedule exceeds both baselines, with average scores ranging from 84.87\% to 85.75\%. Sequential BoN is also a strong reference because it spends the full budget on five independent runs of a competitive simple harness and returns the best outcome. Nevertheless, every adaptive schedule exceeds it on average, with the strongest policy improving the average score from 84.35\% to 85.75\%. Because the unpruned ensemble draws from the same harness pool but does not use intermediate feedback, this consistent improvement supports the conclusion that partial-run scores are useful for deciding where to spend the remaining compute, rather than the gains arising only from trying multiple harnesses. We only present the results with a maximum of 2 survivors in the main paper. The full list of ablations is shown in Appendix \ref{app:full_budget_matched_allocation_results}. 

\begin{table*}[!t]
\centering
\caption{
\textbf{Budget-matched online harness allocation.} We report mean final scores over 100,000 empirical resampling trials per model–problem pair under a budget of five full-run equivalents. Avg. is the unweighted average across the 12 model–problem pairs, and SE is its Monte Carlo standard error. Scores are reported as percentages. Higher is better; bold marks the best value in each score column, and underlined Avg. values indicate a statistically significant vector-battle advantage over the Sequential BoN (Appendix \ref{app:full_budget_matched_allocation_results}).}
\label{tab:expanded_strategy_deviation}
\normalsize
\setlength{\tabcolsep}{2.2pt}
\renewcommand{\arraystretch}{1.08}
\resizebox{\textwidth}{!}{%
\begin{tabular}{@{}l*{14}{r}@{}}
\toprule
Strategy / schedule
& \shortstack{CP\\Q3B} & \shortstack{CP\\Q4B} & \shortstack{CP\\GPT\\20B} & \shortstack{CP\\GPT\\120B} & \shortstack{HT\\Q3B} & \shortstack{HT\\Q4B} & \shortstack{HT\\GPT\\20B} & \shortstack{HT\\GPT\\120B} & \shortstack{SA\\Q3B} & \shortstack{SA\\Q4B} & \shortstack{SA\\GPT\\20B} & \shortstack{SA\\GPT\\120B} & Avg. & SE \\
\midrule
Sequential BoN reference, 5 full & 78.42 & 93.60 & 98.63 & 97.56 & 24.50 & \textbf{56.05} & 91.29 & 89.80 & 95.38 & 96.09 & \textbf{96.59} & 94.27 & 84.35 & 0.02 \\
Single-harness baseline, 5 full & 74.46 & 89.61 & 97.77 & 96.90 & 25.39 & 43.90 & 91.69 & 89.81 & 95.06 & 95.47 & 95.65 & 94.20 & 82.49 & 0.02 \\
Unpruned harness portfolio, 5 full & 77.33 & 93.13 & 98.37 & 97.37 & 29.11 & 51.08 & 94.01 & 91.89 & 95.36 & 95.97 & 96.15 & \textbf{94.73} & \underline{84.54} & 0.02 \\

\addlinespace[2pt]
\multicolumn{15}{@{}l}{\textit{Single-stage pruning at 25\%}} \\
25\%: 17$\to$1 & 77.74 & 94.78 & 98.68 & 97.39 & 35.82 & 45.48 & 93.92 & 92.57 & 95.44 & 96.30 & 96.03 & 94.27 & \underline{84.87} & 0.02 \\
25\%: 14$\to$2 & 78.68 & \textbf{94.79} & 98.72 & 97.60 & \textbf{35.93} & 48.59 & 94.60 & 93.55 & 95.49 & \textbf{96.32} & 96.26 & 94.29 & \underline{85.40} & 0.02 \\

\addlinespace[2pt]
\multicolumn{15}{@{}l}{\textit{Single-stage pruning at 50\%}} \\
50\%: 9$\to$1 & 78.13 & 94.07 & 98.50 & 97.58 & 32.97 & 50.58 & 94.25 & 92.73 & 95.45 & 96.17 & 96.18 & 94.53 & \underline{85.09} & 0.02 \\
50\%: 8$\to$2 & 78.55 & 94.06 & 98.59 & 97.59 & 32.69 & 52.29 & 94.65 & 93.07 & 95.48 & 96.18 & 96.25 & 94.50 & \underline{85.32} & 0.02 \\

\addlinespace[2pt]
\multicolumn{15}{@{}l}{\textit{Two-stage pruning at 25\%/50\%}} \\
10$\to$6$\to$2 & \textbf{79.24} & 94.39 & 98.67 & 97.69 & 34.03 & 52.87 & 94.90 & 93.40 & 95.50 & 96.26 & 96.32 & 94.62 & \underline{85.66} & 0.02 \\
10$\to$8$\to$1 & 78.52 & 94.23 & 98.55 & 97.63 & 33.48 & 51.35 & 94.44 & 92.96 & 95.47 & 96.21 & 96.23 & 94.63 & \underline{85.31} & 0.02 \\
12$\to$6$\to$1 & 78.79 & 94.48 & 98.66 & 97.69 & 34.73 & 51.29 & 94.82 & 93.44 & 95.49 & 96.25 & 96.23 & 94.62 & \underline{85.54} & 0.02 \\
14$\to$4$\to$1 & 78.56 & 94.67 & 98.72 & 97.69 & 35.58 & 50.17 & 95.05 & \textbf{93.78} & \textbf{95.51} & 96.28 & 96.22 & 94.38 & \underline{85.55} & 0.02 \\

\addlinespace[2pt]
\multicolumn{15}{@{}l}{\textit{Three-stage pruning at 25\%/50\%/75\%}} \\
8$\to$6$\to$5$\to$1 & 78.43 & 94.04 & 98.50 & 97.57 & 32.64 & 54.21 & 94.63 & 92.70 & 95.47 & 96.18 & 96.21 & 94.71 & \underline{85.44} & 0.02 \\
10$\to$5$\to$3$\to$2 & 79.23 & 94.42 & 98.68 & 97.69 & 33.99 & 53.09 & 95.07 & 93.50 & 95.51 & 96.26 & 96.31 & 94.58 & \underline{85.69} & 0.02 \\
10$\to$6$\to$3$\to$1 & 79.08 & 94.39 & 98.63 & 97.69 & 34.02 & 53.62 & 94.96 & 93.25 & 95.49 & 96.25 & 96.28 & 94.66 & \underline{85.69} & 0.02 \\
12$\to$5$\to$2$\to$1 & 79.17 & 94.58 & \textbf{98.73} & \textbf{97.72} & 35.07 & 52.42 & \textbf{95.11} & 93.62 & 95.51 & 96.28 & 96.26 & 94.51 & \textbf{\underline{85.75}} & 0.02 \\
\bottomrule
\end{tabular}%
}
\end{table*}

The strongest policies are those starting broader and cutting more. Specifically, the strongest policy is the three-stage 12→5→2→1 schedule, which prunes at 25\%, 50\%, and 75\% of a full run. It achieves the highest average score, 85.75\%, compared with 84.54\% for the unpruned harness portfolio, 82.49\% for single-harness commitment, and 84.35\% for five Sequential BoN runs. It also outperforms the unpruned harness portfolio on 11 of the 12 model–problem pairs. Overall, the strongest policies begin with broader portfolios and progressively concentrate the budget as additional feedback becomes available. \textbf{This result illustrates the benefit of beginning with a diverse set of partial runs and progressively concentrating the budget as more informative feedback becomes available.}

\findingbox{\textbf{Finding 3.} Adapting harness choice online improves average final performance over both single-harness commitment and an unpruned harness portfolio under the same budget.}

\section{Related Work}

\textbf{LLM-guided discovery systems and evolutionary coding agents.} Recent work uses LLMs as generators inside iterative discovery loops, where candidate programs are proposed, evaluated, and improved over time. FunSearch \citep{romera2024mathematical}, LLMs as Optimizers \citep{yang2024large}, AlphaEvolve \citep{novikov2025alphaevolve}, and TTT-Discover \citep{yuksekgonul2026learning} show that combining LLM generation with evaluator feedback and search can produce useful mathematical, algorithmic, and scientific discoveries. A growing set of follow-up systems---including OpenEvolve \citep{openevolve}, ShinkaEvolve \citep{lange2025shinkaevolve}, DeltaEvolve \citep{jiang2026deltaevolve}, AdaEvolve \citep{cemri2026adaevolve}, CodeEvolve \citep{assumpccao2025codeevolve}, DeepEvolve \citep{liu2025scientific}, and GigaEvo \citep{khrulkov2025gigaevo}---extend these loops with additional harness ingredients such as novelty filtering, adaptive optimization, island models, and inspiration sampling, while a parallel line evolves optimization heuristics directly \citep{ye2024reevo, liu2024evolution, liu2026eoh}. These works evaluate new composite systems or ingredients end to end. We instead treat the harness itself as the object of study, decomposing representative recipes into simpler search choices and asking which ingredients transfer across model--problem pairs under matched budgets and repeated trials.

\textbf{Evaluation rigor and simple baselines.} Concerns about attributing gains to methods rather than run-to-run variance are long-standing: deep RL results are highly seed-sensitive \citep{henderson2018deep}, random search is a strong baseline for neural architecture search \citep{li2020random, yang2019evaluation}, repeated sampling alone scales remarkably well for LLMs \citep{brown2024large}, and agent evaluations often lack seed and cost reporting \citep{kapoor2024ai}. Similar concerns are emerging for LLM-guided discovery. \citet{gideoni2026simple} show that simple baselines match sophisticated code-evolution pipelines when compared end to end, attributing performance primarily to search-space and prompt design, while \citet{tanveer2026levi} shows that harness design is high-leverage: a redesigned search architecture substitutes for a larger mutation model, achieving far better sample efficiency than OpenEvolve. Both indicate that default composite recipes are beatable---whether by simpler baselines or by different designs. \citet{pelleriti2026evolutionary} release annotated search traces to analyze which code edits drive gains in evolutionary coding agents. Our study differs in granularity and inferential machinery: we ablate individual harness ingredients under matched budgets against multi-run empirical null distributions, showing that ingredients yielding statistically significant gains on some model--problem pairs can simultaneously degrade others, and our released run pools serve as reusable null distributions for hypothesis testing rather than traces for mechanism analysis.

\textbf{Search algorithms and budgeted allocation.} Our decompositions connect LLM-guided discovery to classical mechanisms: UCT \citep{kocsis2006bandit} and PUCT-style tree search \citep{silver2017mastering}, and archive-based diversity via MAP-Elites \citep{mouret2015illuminating} and Novelty Search \citep{lehman2011abandoning}. Recent discovery harnesses embed these ideas but are typically evaluated end to end; our ablations instead isolate how each choice behaves under a matched rollout budget. The allocation side of our study builds on algorithm selection and portfolios \citep{rice1976algorithm, gomes2001algorithm} and fixed-budget resource allocation such as Successive Halving \citep{jamieson2016non}, Hyperband \citep{li2018hyperband}, and ASHA \citep{li2020system}. In concurrent work, \citet{xing2026compute} characterize depth--breadth allocation in LLM-guided evolutionary search, finding task-specific, capability-gated fitness surfaces---consistent with our finding that breadth--depth preferences do not transfer---and propose a bandit that allocates LLM calls across parallel trajectories, while \citet{ishibashi2026effective} report that fewer, deeper generations outperform many shallow ones on circle packing under a fixed token budget---an allocation preference along a different budget axis (reasoning tokens per candidate rather than rollouts) whose direction is pair-dependent in our rollout-budget sweeps, itself an instance of allocation preferences failing to transfer. These allocators, like AdaEvolve's island-level scheduler \citep{cemri2026adaevolve}, operate within a single run, whereas our Adaptive Harness Ensemble allocates across entire harness configurations. We propose no new bandit or early-stopping algorithm; we use allocation as a causal test of our findings: if fixed harnesses do not transfer and early feedback is informative, online allocation across harnesses should improve over non-adaptive ways of spending the same budget.

\textbf{Harness adaptation and meta-harness search.} A closely related line treats the harness around an LLM as an object to search over, synthesize, or adapt. This idea predates discovery systems: ADAS \citep{hu2025automated} and AgentSquare \citep{shang2025agentsquare} search modular agent design spaces, and Archon \citep{saad2024archon} casts the composition of inference-time techniques as a hyperparameter optimization objective. In the discovery setting, Meta-Harness \citep{lee2026meta} searches over complete harness programs before solving held-out tasks, and EvoX \citep{liu2026evox} meta-evolves discovery strategies themselves. A rapidly growing harness-engineering literature studies automatic synthesis or modification of agent harnesses \citep{zhang2026self, lin2026agentic, lou2026autoharness, chen2026harnessforge, ning2026code}, while HARBOR \citep{sengupta2026harbor} tunes harness feature flags via Bayesian optimization and Continual Harness \citep{karten2026continual} adapts an agent's harness online during deployment. These works presuppose that adapting the harness is worthwhile; our study establishes the empirical conditions under which it is. In our work, we do not propose a learned or meta-evolved harness. Instead, our controlled decompositions show that fixed harnesses---including simple baselines and the ingredients of complex recipes---can yield statistically significant gains on some model--problem pairs while degrading others, so no fixed choice can be safely committed to in advance. Our proposal is therefore a framing rather than a mechanism: when fixed harnesses fail to transfer and early evaluator feedback is informative, harness selection is better treated as an online resource-allocation problem than as a one-time design decision.

\section{Conclusion}
Through large-scale and statistically-controlled evaluation of complex discovery harnesses, we find that fixed harnesses do not reliably generalize across models and problems. Harness components that help in one setting may provide little benefit—or even hurt—in another, and no fixed configuration emerges as a reliably superior choice. These results suggest that harness choice should be treated as a model- and problem-dependent decision rather than as a transferable methodological recipe. At the same time, partial-run performance provides useful evidence about which harnesses usefulness. By using this evidence to prune weaker harnesses and reallocate their budget, online adaptation improves average final performance over both committing to a single harness and maintaining an unpruned harness portfolio. This points toward discovery systems that adapt not only their candidate solutions, but also the harness used to generate them. We release our rollout-level data and statistical artifacts for future research.

\subsubsection*{Acknowledgments}
This research was been funded in part by JP Morgan AI Faculty Research Award and recieved support from Google.org. We thank Pranav Shetty for his valuable feedback, insightful discussions, and detailed comments on earlier versions of this paper, and Zhiqiang Ma for ongoing discussions throughout the project.

\bibliography{main}
\bibliographystyle{tmlr}

\FloatBarrier
\clearpage
\appendix
\section*{Appendix}

\section{Online Budget Allocation Algorithms}\label{app:ensemble}

\begin{figure}[H]
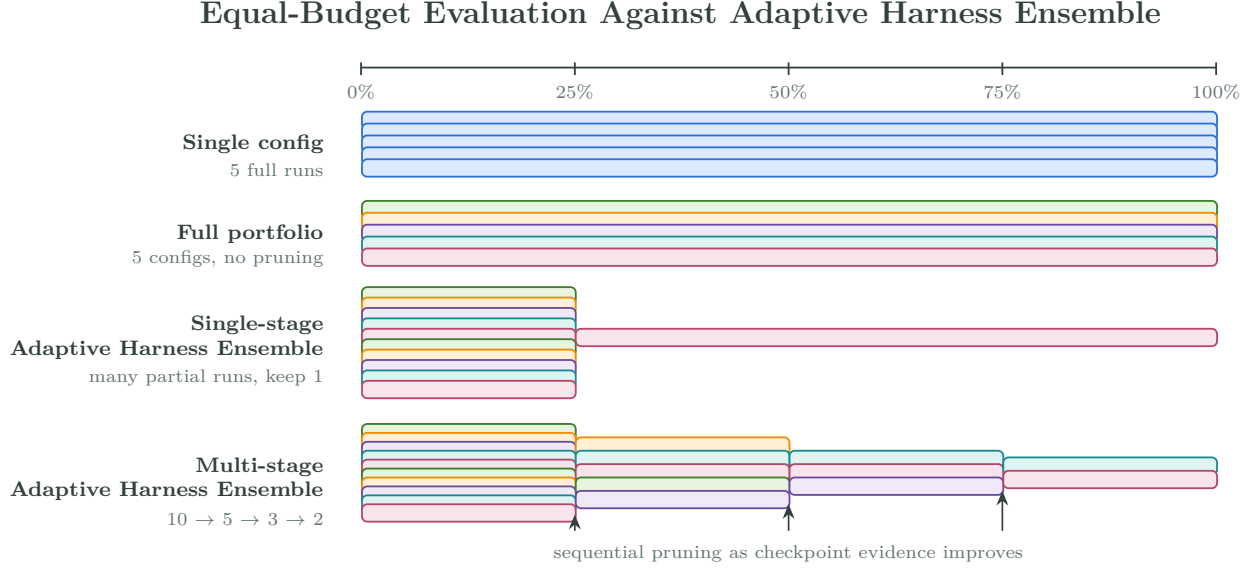

\centering
\ensembleevolvestrategydeviationgraphic{\textwidth}{!}
\caption{\textbf{Adaptive allocation schematic.} The comparison is budget matched across fixed repeated-run baselines, a non-adaptive full ensemble, single-stage Adaptive Harness Ensemble, and multi-stage Adaptive Harness Ensemble. Starting with the full-ensemble baseline, different hues denote distinct search configurations; in the Adaptive Harness Ensemble rows, retained configurations keep their hue after pruning.}
\label{fig:ensemble_evolve_strategy_deviation}
\end{figure}

\begin{algorithm}[H]
\caption{Single-Harness Baseline}
\label{alg:single-harness}
\begin{algorithmic}[1]
\Require Harness pool $\mathcal{H}$; run budget $R=5$
\Ensure Best final score $s^\star$
\State Sample one harness $h \sim \mathrm{Uniform}(\mathcal{H})$
\State $S \gets \emptyset$
\For{$r = 1,\ldots,R$}
    \State $s \gets \textsc{RunToCompletion}(h)$
    \State $S \gets S \cup \{s\}$
\EndFor
\State $s^\star \gets \max S$
\State \Return $s^\star$
\end{algorithmic}
\end{algorithm}

\begin{algorithm}[H]
\caption{Unpruned Harness Ensemble}
\label{alg:unpruned-portfolio}
\begin{algorithmic}[1]
\Require Harness pool $\mathcal{H}$; run budget $R=5$
\Ensure Best final score $s^\star$
\State Sample ensemble $\mathcal{P} \subseteq \mathcal{H}$ with $|\mathcal{P}|=R$
\State $S \gets \emptyset$
\ForAll{$h \in \mathcal{P}$}
    \State $s \gets \textsc{RunToCompletion}(h)$
    \State $S \gets S \cup \{s\}$
\EndFor
\State $s^\star \gets \max S$
\State \Return $s^\star$
\end{algorithmic}
\end{algorithm}

\begin{algorithm}[H]
\caption{Adaptive Harness Ensemble}
\label{alg:ensemble-evolve}
\begin{algorithmic}[1]
\Require Harness pool $\mathcal{H}$; checkpoints $(c_1,\ldots,c_K)$; survivor counts $(m_0,m_1,\ldots,m_K)$
\Ensure Best final score $s^\star$
\State Sample initial active set $\mathcal{A} \subseteq \mathcal{H}$ with $|\mathcal{A}|=m_0$
\For{$k = 1,\ldots,K$}
    \ForAll{$h \in \mathcal{A}$}
        \State Run $h$ until checkpoint $c_k$
        \State $q_h \gets \textsc{CurrentBestScore}(h)$
    \EndFor
    \State $\mathcal{A} \gets \textsc{TopM}(\mathcal{A}, \{q_h : h \in \mathcal{A}\}, m_k)$
\EndFor
\State $S \gets \emptyset$
\ForAll{$h \in \mathcal{A}$}
    \State $s \gets \textsc{RunToCompletion}(h)$
    \State $S \gets S \cup \{s\}$
\EndFor
\State $s^\star \gets \max S$
\State \Return $s^\star$
\end{algorithmic}
\end{algorithm}

\section{Detailed Search-Harness Definitions}

\subsection{Search-Harness Ablation Details}\label{app:search_harness_ablation_details}

Section~\ref{sec:experiments} presents the ablations as two compact progressions: Sequential BoN $\rightarrow$ archive search $\rightarrow$ exploration $\rightarrow$ OpenEvolve-style refinement, and Sequential BoN $\rightarrow$ UCT $\rightarrow$ PUCT. The appendix expands each step with the full task-level sweeps, model-specific results, and implementation details needed to interpret the compressed main-text figures. Table~\ref{tab:appendix_ablation_detail_map} provides a guide to this material.

\begin{table}[ht]
  \centering
  \caption{\textbf{Guide to the appendix ablations.} Each row links a compressed main-text comparison to its complete task-level evidence.}
  \label{tab:appendix_ablation_detail_map}
  \resizebox{\textwidth}{!}{%
  \begin{tabular}{p{0.22\textwidth}p{0.30\textwidth}p{0.36\textwidth}}
    \toprule
    Main-text step & Detailed figures & What the detailed view adds \\
    \midrule
    Sequential BoN baseline & Figure~\ref{fig:sequential_baseline_kdes} & Baseline score distributions by task and model, used to interpret later bootstrap comparisons. \\
    Top-$K$ archive & Figure~\ref{fig:elite_archive_sweep} & Full task-level archive-size sweep with significance markers against Sequential BoN. \\
    General-history exploration & Figure~\ref{fig:exploration_sweep} & Full $\epsilon$-probability sweep with $K=1$, showing where exploration helps or destabilizes search. \\
    Full epsilon-greedy grid & Figures~\ref{fig:circle_packing_full_epsilon_greedy}, \ref{fig:heilbronn_triangle_full_epsilon_greedy}, and~\ref{fig:second_autocorrelation_inequality_full_epsilon_greedy} & Interaction between elite archive size and $\epsilon$ probability across each task. \\
    OpenEvolve progression & Figure~\ref{fig:openevolve_progression_all_models} & Breadth--depth and island-style refinements showing that OpenEvolve-style search is competitive but not uniformly dominant. \\
    UCT/PUCT family & Figures~\ref{fig:appendix_uct_puct_c_ablation}, \ref{fig:puct_b_ablation}, and~\ref{fig:puct_p_ablation} & Detailed tree-search knob sweeps for breadth--depth, number of parent programs, and exploration constant. \\
    \bottomrule
  \end{tabular}%
  }
\end{table}

The detailed sweeps diagnose individual mechanisms, whereas Section~\ref{sec:config_battle} asks whether a complete configuration transfers across model--problem pairs. The two analyses therefore answer complementary questions: the ablations identify where particular design choices help or hurt, and the battle analysis measures whether those gains persist across settings.

\input{figures/search_algorithms_openevolve}
% =====================================================================
% figures/search_algorithms_puct.tex
% PREAMBLE needs:
%   \usepackage{tikz}
%   \usepackage{xcolor}
%   \usepackage{amsmath}
%   \usetikzlibrary{arrows.meta, positioning, calc, shapes.arrows}
% =====================================================================

\begin{figure*}[t]
\centering

\definecolor{cardbg}{HTML}{F7FAF9}
\definecolor{cardborder}{HTML}{CFE2DE}
\definecolor{purplebg}{HTML}{F3F0FA}
\definecolor{bluefill}{HTML}{BFD7FF}
\definecolor{blueedge}{HTML}{2F6FD6}
\definecolor{orangefill}{HTML}{FDBB5B}
\definecolor{orangeedge}{HTML}{F08A00}
\definecolor{purplefill}{HTML}{C9B8E8}
\definecolor{purpleedge}{HTML}{6B4AA0}
\definecolor{greenfill}{HTML}{B7D99C}
\definecolor{greenedge}{HTML}{3D7F2A}
\definecolor{slate}{HTML}{33403E}
\definecolor{graytext}{HTML}{63736F}
\definecolor{arrowgray}{HTML}{4B5452}

\resizebox{\textwidth}{!}{%
\begin{tikzpicture}[
  panel/.style={rounded corners=8pt, fill=cardbg, draw=cardborder,
                line width=0.9pt, minimum width=6.25cm, minimum height=7.30cm,
                anchor=north west},
  treepanel/.style={panel, minimum width=10.95cm, fill=purplebg, draw=purpleedge},
  scorepanel/.style={panel, minimum width=6.25cm},
  node/.style={circle, draw=orangeedge, fill=orangefill, line width=0.9pt,
               minimum size=5.3mm, inner sep=0pt,
               font=\scriptsize\bfseries, text=slate},
  parent/.style={circle, draw=blueedge, fill=bluefill, line width=1.05pt,
                 minimum size=5.7mm, inner sep=0pt,
                 font=\scriptsize\bfseries, text=slate},
  revisit/.style={circle, draw=purpleedge, fill=purplefill, line width=0.95pt,
                  minimum size=5.5mm, inner sep=0pt,
                  font=\scriptsize\bfseries, text=slate},
  best/.style={circle, draw=greenedge, fill=greenfill, line width=1.1pt,
               minimum size=5.5mm, inner sep=0pt},
  edge/.style={-{Stealth[length=1.5mm,width=1.2mm]}, draw=slate,
               line width=0.55pt, shorten >=1.5pt, shorten <=1.5pt},
  choice/.style={-{Stealth[length=1.8mm,width=1.4mm]}, draw=blueedge,
                 line width=1.0pt, shorten >=1.5pt, shorten <=1.5pt},
  revisitedge/.style={-{Stealth[length=1.8mm,width=1.4mm]}, dashed,
                      draw=purpleedge, dash pattern=on 2pt off 1.6pt,
                      line width=0.85pt, shorten >=1.5pt, shorten <=1.5pt},
  flowarrow/.style={single arrow, single arrow head extend=3.4mm,
                    draw=arrowgray, fill=arrowgray, text=white,
                    font=\small\bfseries, align=center,
                    inner sep=3pt, minimum height=1.18cm, minimum width=1.58cm},
  title/.style={font=\large\bfseries, text=slate, align=center},
  ptitle/.style={font=\large\bfseries, text=purpleedge, align=center},
  subtitle/.style={font=\footnotesize, text=graytext, align=center},
  note/.style={font=\footnotesize, text=graytext, align=center},
  eqbox/.style={rounded corners=5pt, draw=blueedge, fill=white,
                line width=0.8pt, align=center, inner xsep=7pt,
                inner ysep=5pt, font=\scriptsize, text=slate},
  callout/.style={rounded corners=5pt, draw=cardborder, fill=white,
                  line width=0.75pt, font=\footnotesize, text=slate,
                  align=center, inner xsep=7pt, inner ysep=5pt}
]

\node[panel] (P1) at (0,0) {};
\node[treepanel] (P2) at (7.65,0) {};
\node[scorepanel] (P3) at (20.20,0) {};

% ---------------------------------------------------------------------
% Panel 1: Sequential BoN baseline
% ---------------------------------------------------------------------
\begin{scope}[shift={(P1.north west)}]
  \node[title, text width=5.6cm] at (3.12,-0.50) {1. Sequential BoN};
  \node[subtitle] at (3.12,-0.88)
    {$p_t=\arg\max_{x \in \mathcal{H}_t}S(x)$};

  \node[parent] (x0) at (3.12,-1.42) {$x_0$};
  \node[node] (x1) at (1.75,-2.35) {$x$};
  \node[node] (x2) at (3.12,-2.35) {$x$};
  \node[node] (x3) at (4.49,-2.35) {$x$};
  \draw[edge] (x0) -- (x1);
  \draw[choice] (x0) -- (x2);
  \draw[edge] (x0) -- (x3);

  \node[parent] (pt) at (3.12,-3.35) {$p_t$};
  \draw[choice] (x2) -- (pt);
  \node[node] (c1) at (1.75,-4.25) {$x$};
  \node[node] (c2) at (3.12,-4.25) {$x$};
  \node[node] (c3) at (4.49,-4.25) {$x$};
  \draw[edge] (pt) -- (c1);
  \draw[choice] (pt) -- (c2);
  \draw[edge] (pt) -- (c3);

  \node[best] (hi) at (3.12,-5.18) {};
  \draw[choice] (c2) -- (hi);
  \node[note, text width=5.25cm] at (3.12,-6.22)
    {parent selection ignores the rest of the mutation tree};
\end{scope}

\node[flowarrow, shape border rotate=0, text width=1.35cm]
  at ($(P1.east)!0.5!(P2.west)+(0,-3.10)$) {tree view};

% ---------------------------------------------------------------------
% Panel 2: UCT/PUCT tree
% ---------------------------------------------------------------------
\begin{scope}[shift={(P2.north west)}]
  \node[ptitle, text width=9.9cm] at (5.47,-0.50) {2. UCT / PUCT Tree};
  \node[subtitle, text=slate] at (5.47,-0.88)
    {all nodes are programs $x \in \mathcal{H}_t$};

  \node[parent] (root) at (5.47,-1.35) {$x_0$};
  \node[node] (left) at (1.65,-2.30) {$x$};
  \node[revisit] (mid) at (5.47,-2.30) {$x$};
  \node[node] (right) at (9.30,-2.30) {$x$};
  \draw[edge] (root) -- (left);
  \draw[edge] (root) -- (mid);
  \draw[edge] (root) -- (right);

  \node[node] (l1) at (0.95,-3.20) {$x$};
  \node[node] (l2) at (1.65,-3.20) {$x$};
  \node[node] (l3) at (2.35,-3.20) {$x$};
  \draw[edge] (left) -- (l1);
  \draw[choice] (left) -- (l2);
  \draw[edge] (left) -- (l3);
  \node[note, text width=2.40cm] at (1.65,-4.08)
    {high $S(x)$, then repeated expansion};

  \node[node] (m1) at (4.55,-3.20) {$x$};
  \node[parent] (m2) at (5.47,-3.20) {$p_t$};
  \node[best] (m3) at (6.40,-3.20) {};
  \draw[edge] (mid) -- (m1);
  \draw[choice] (mid) -- (m2);
  \draw[edge] (mid) -- (m3);
  \node[note, text width=2.58cm] at (5.47,-4.08)
    {$Q_t(x)$ uses $\mathcal{D}_t(x)$};

  \node[node] (r1) at (8.40,-3.20) {$x$};
  \node[revisit] (r2) at (9.30,-3.20) {$x$};
  \node[node] (r3) at (10.20,-3.20) {$x$};
  \draw[edge] (right) -- (r1);
  \draw[choice] (right) -- (r2);
  \draw[edge] (right) -- (r3);
  \node[node] (r4) at (8.75,-4.25) {$x$};
  \node[node] (r5) at (9.85,-4.25) {$x$};
  \draw[edge] (r2) -- (r4);
  \draw[choice] (r2) -- (r5);
  \node[note, text width=2.70cm] at (9.35,-5.08)
    {under-expanded branch can be selected};

  \draw[revisitedge, line width=1.0pt] (l2) to[out=30,in=190] (mid);
  \draw[revisitedge, line width=1.0pt] (m2) to[out=-20,in=205] (r2);
  \draw[revisitedge, line width=1.0pt] (r2) to[out=155,in=25] (mid);

  \node[callout, draw=purpleedge, text width=8.65cm] at (5.47,-6.42)
    {$p_t=\arg\max_{x \in \mathcal{H}_t}\operatorname{UCT}_t(x)$
     or
     $p_t=\arg\max_{x \in \mathcal{H}_t}\operatorname{PUCT}_t(x)$};
\end{scope}

\node[flowarrow, shape border rotate=0, text width=1.45cm]
  at ($(P2.east)!0.5!(P3.west)+(0,-3.10)$) {select $p_t$};

% ---------------------------------------------------------------------
% Panel 3: Parent selection
% ---------------------------------------------------------------------
\begin{scope}[shift={(P3.north west)}]
  \node[title, text width=5.8cm] at (3.12,-0.50) {3. Parent Selection};
  \node[subtitle] at (3.12,-0.88) {same history, different scoring rule};

  \node[eqbox, text width=5.35cm] at (3.12,-1.62)
    {$Q_t(x)=\max_{d \in \mathcal{D}_t(x)\cup\{x\}}S(d)$};

  \node[eqbox, text width=5.45cm] at (3.12,-2.92)
    {$\operatorname{UCT}_t(x)
      =Q_t(x)+c_{\text{uct}}
      \sqrt{\dfrac{\log(N_p(t)+1)}{1+N_x(t)}}$};

  \node[eqbox, text width=5.45cm] at (3.12,-4.78)
    {$\operatorname{PUCT}_t(x)
      =Q_t(x)+c_{\text{puct}}P_t(x)
      \dfrac{\sqrt{1+t}}{1+N_x(t)}$\\[3pt]
     $P_t(x)=
      \dfrac{n_t-r_t(x)}{\sum_{i=1}^{n_t} i}$};

  \node[note, text width=5.45cm] at (3.12,-6.55)
    {PUCT adds the rank-based prior $P_t(x)$ to the UCT-style subtree score.};
\end{scope}

% ---------------------------------------------------------------------
% Legend
% ---------------------------------------------------------------------
\draw[rounded corners=5pt, draw=cardborder, fill=white, line width=0.8pt]
  (2.05,-8.55) rectangle (24.40,-7.70);

\node[node, minimum size=4.3mm] at (2.82,-8.13) {$x$};
\node[note, anchor=west, text=slate] at (3.15,-8.13) {program in $\mathcal{H}_t$};

\node[parent, minimum size=4.3mm] at (8.62,-8.13) {$p_t$};
\node[note, anchor=west, text=slate] at (8.95,-8.13) {selected parent};

\node[revisit, minimum size=4.3mm] at (13.50,-8.13) {$x$};
\node[note, anchor=west, text=slate] at (13.83,-8.13) {later selected program};

\node[best, minimum size=4.3mm] at (19.15,-8.13) {};
\node[note, anchor=west, text=slate] at (19.48,-8.13) {highest $S(x)$ found};

\end{tikzpicture}%
}

\caption{\textbf{UCT/PUCT as tree-based search.}
Sequential BoN selects $p_t$ using only the highest evaluator score in the full history $\mathcal{H}_t$.
UCT and PUCT instead score each program $x \in \mathcal{H}_t$ using the induced mutation tree: $Q_t(x)$ propagates the best score in the descendant set $\mathcal{D}_t(x)$, visitation counts adjust exploration, and PUCT additionally uses the rank-based prior $P_t(x)$.
The LLM generation step and history update remain the same after $p_t$ is selected.}
\label{fig:puct_search}
\end{figure*}
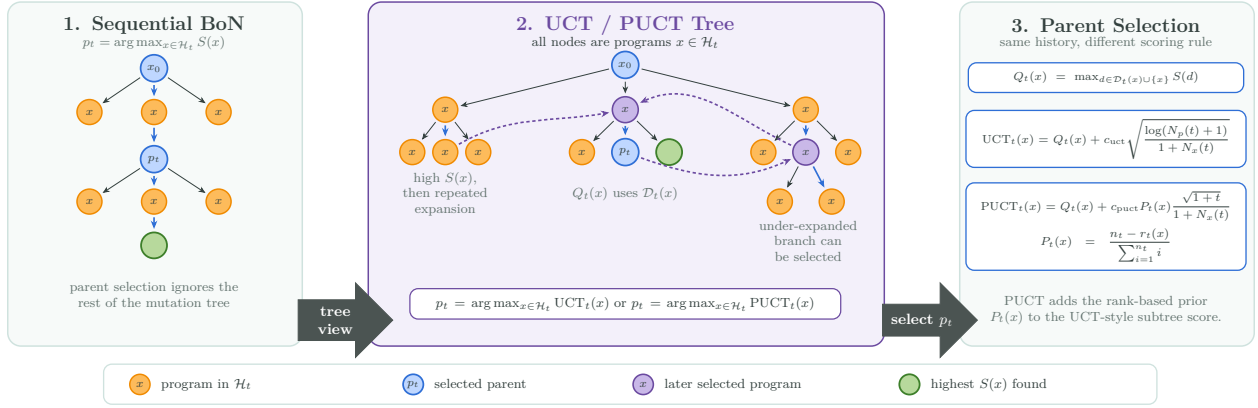

\subsection{From Sequential BoN to OpenEvolve}\label{app:bon-to-openevolve}
This section explains how we can systematically build towards OpenEvolve from a simple Sequential BoN baseline.

\subsubsection{Adding a TopK Elite Archive}
Sequential BoN keeps only the current best program active, which corresponds to sampling from $\mathcal{E}^{1}_{t}$. A direct generalization is to sample parents from the top-$K$ elite archive,

\begin{equation}
\label{eq:top-k-parent-selection}
\mathcal{E}^{K}_{t} = \operatorname{TopK}(\mathcal{H}_t, K, S),
\qquad
p_t \sim \mathcal{E}^{K}_{t}.
\end{equation}

The case $K=1$ recovers Sequential BoN, while $K>1$ allows the search to revisit several high-scoring programs rather than committing all future mutations to a single best-so-far lineage. This preserves a small amount of diversity while still focusing compute on programs that have already achieved high evaluator scores.

\subsubsection{Adding Epsilon-Greedy Exploration}
The elite archive can still become too greedy when its high-scoring programs represent closely related lineages. To permit occasional lineage changes, OpenEvolve mixes elite-archive sampling with sampling from the full history. Let $\epsilon$ denote the exploration probability. The parent-selection rule is

\begin{equation}
\label{eq:epsilon-greedy-parent-selection}
p_t \sim
\begin{cases}
\mathcal{E}^{K}_{t}, & \text{with probability } 1-\epsilon,\\
\mathcal{H}_t, & \text{with probability } \epsilon.
\end{cases}
\end{equation}

When $K=1$ and $\epsilon=0$, this reduces to Sequential BoN. When $K>1$, exploitation samples among the top-$K$ programs, and when $\epsilon>0$, exploration samples uniformly from all previously evaluated programs in $\mathcal{H}_t$. This separates two sources of non-greediness: keeping multiple elite lineages through $K$, and occasionally revisiting non-elite historical programs through $\epsilon$.

%\lc{if you need space, you can remove from each such paragraph the details. So for exxample all the parts of first this and then this and so on. Those are described in the setup and not critical here as you will not single out any of them anyway.}
\subsubsection{Trading Breadth for Depth}
The archive and epsilon-greedy variants generate $N$ children from each selected parent before updating the history. OpenEvolve-style search is more depth-oriented: it generates fewer children per expansion and performs more sequential updates. We compare these regimes at a fixed rollout budget, $B=NT$. Decreasing $N$ therefore requires increasing $T$. Larger $N$ spends more budget exploring alternatives around one parent, whereas larger $T$ permits longer chains of refinement. The setting $N=1$ is the most depth-oriented and most closely matches the OpenEvolve allocation.

\subsubsection{OpenEvolve-Specific Context and Parallelism}
After adding elite-archive sampling, epsilon-greedy exploration, and depth-oriented budget allocation, the remaining OpenEvolve components modify the context and parallelization structure rather than the basic history update rule:

\begin{enumerate}
  \item \textbf{MAP-Elites inspiration sampling.} In addition to the selected parent $p_t$, OpenEvolve provides the LLM with a small set of inspiration programs sampled from a MAP-Elites grid over past solutions. In our notation, this augments the context with additional programs from $\mathcal{H}_t$ while leaving the parent-selection rule unchanged.
  \item \textbf{Multi-island evolution and migration.} OpenEvolve runs several searches in parallel rather than maintaining a single history. Each island maintains its own history and elite archive, and high-scoring programs are periodically migrated between islands. This corresponds to applying the archive update rule to several histories with occasional exchange of elite programs.
\end{enumerate}

\subsection{From Sequential BoN to TTT-Discover}\label{app:bon-to-tttdiscovery}
The first step in this progression is Upper Confidence Bounds applied to Trees (UCT) \citep{kocsis2006bandit}. UCT replaces the node-local score $S(x)$ with a value estimate for the entire subtree rooted at $x$. Let $\mathcal{D}_t(x)$ denote the descendants of program $x$ in the mutation tree induced by $\mathcal{H}_t$. Following the evolutionary-search formulation used in \cite{yuksekgonul2026learning}, we propagate the best descendant score rather than the average rollout return used in conventional Monte Carlo Tree Search:

\begin{equation}
\label{eq:uct-subtree-value}
  Q_t(x) = \max_{d \in \mathcal{D}_t(x) \cup \{x\}} S(d).
\end{equation}

UCT then selects the next parent by adding an exploration bonus based on visitation counts,

\begin{equation}
\label{eq:uct-parent-selection}
  \begin{aligned}
    \operatorname{UCT}_t(x) &= Q_t(x) + c_{\text{uct}} \sqrt{\frac{\log (N_p(t) + 1)}{1 + N_x(t)}}\\
    p_t &= \arg\max_{x \in \mathcal{H}_t} \operatorname{UCT}_t(x),
  \end{aligned}
\end{equation}

where $N_x(t)$ is the number of times program $x$ has been expanded, $N_p(t)$ is the visitation count of its parent node, and $c_{\text{uct}}$ controls the strength of exploration. Relative to Sequential BoN, UCT makes two changes. First, a parent is credited for the best solution discovered anywhere in its descendant subtree. Second, repeatedly expanded nodes receive a smaller exploration bonus, allowing the search to revisit branches that are promising but not yet exhausted. Once UCT selects $p_t$, the LLM generation step and history update are unchanged.

TTT-Discover \citep{yuksekgonul2026learning} adds a rank-based prior over archived programs to estimate which nodes are likely to be useful parents. For $n_t = |\mathcal{H}_t|$, let $r_t(x) \in \{0,\ldots,n_t-1\}$ denote the rank of program $x$ in decreasing order of evaluator score. The PUCT score is

\begin{equation}
\label{eq:puct-parent-selection}
  \begin{aligned}
    \operatorname{PUCT}_t(x) &= Q_t(x) + c_{\text{puct}} P_t(x)\frac{\sqrt{1+t}}{1+N_x(t)}\\
    P_t(x) &= \frac{n_t-r_t(x)}{\sum_{i=1}^{n_t} i}\\
    p_t &= \arg\max_{x \in \mathcal{H}_t} \operatorname{PUCT}_t(x),
  \end{aligned}
\end{equation}

where $P_t(x)$ assigns larger prior mass to higher-ranked programs and $c_{\text{puct}}$ controls the strength of the prior-weighted exploration bonus. Sequential BoN, UCT, and PUCT all select a parent from the evaluated history, but differ in the score used for that selection: node-local evaluator score for Sequential BoN, subtree value plus a visit-count bonus for UCT, and subtree value plus a prior-weighted bonus for PUCT.

\section{Task Details and Qualitative Strategy Analysis}
\label{app:task-details}

\subsection{Task Definitions}

\begin{figure}[t]
\centering
\includegraphics[width=0.95\textwidth]{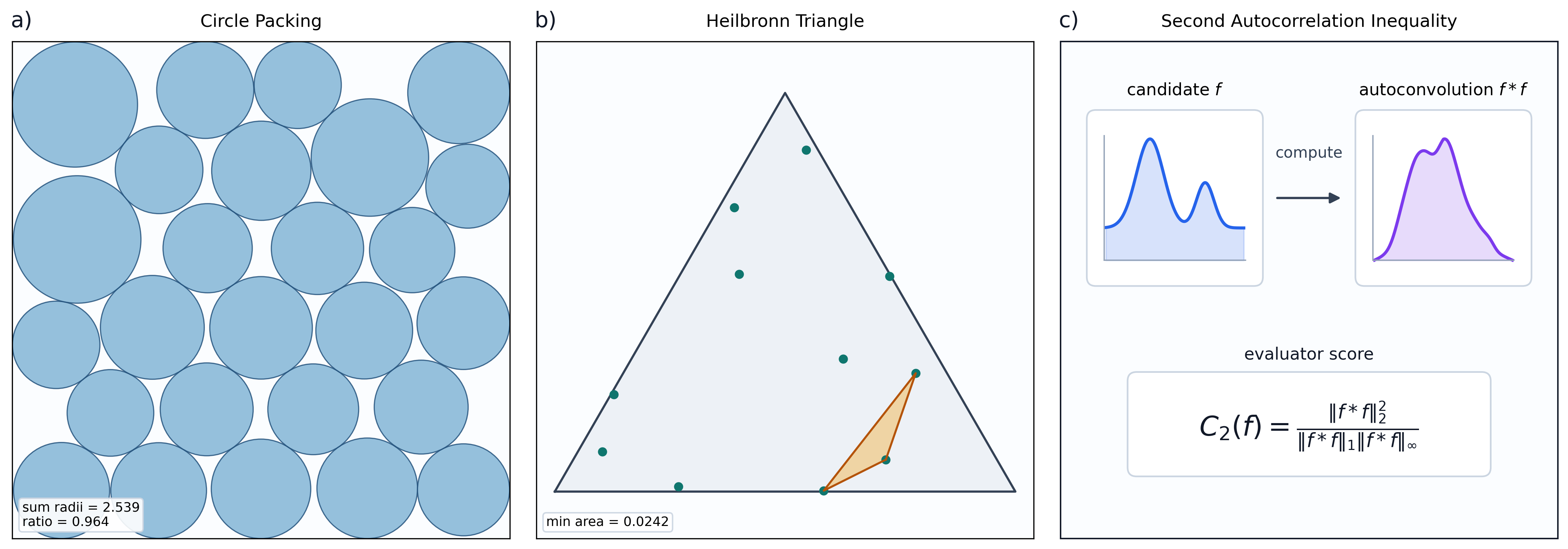}
\caption{\textbf{Representative discovery tasks.} The tasks include two geometric optimization problems and one non-geometric autocorrelation objective.}
\label{fig:task_examples}
\end{figure}

Figure~\ref{fig:task_examples} shows representative examples from each task.

\begin{itemize}
\item \textbf{Circle Packing.} The goal is to fit 26 non-overlapping circles inside a unit square while maximizing the sum of their radii. This task provides an intuitive geometric optimization setting in which candidate improvements can be inspected visually.
\item \textbf{Heilbronn Triangle.} The goal is to place points inside an equilateral triangle while maximizing the minimum area of any triangle formed by three points. The objective is globally coupled because moving one point can affect many of the induced triangles.
\item \textbf{Second Autocorrelation Inequality.} The goal is to construct a non-negative function, represented numerically on a fixed grid, that performs well under an autocorrelation-based objective. This task tests whether search strategies that work on geometric problems transfer to a more abstract, non-geometric setting.
\end{itemize}

For the second autocorrelation inequality task, a candidate program defines a vector of non-negative values representing a discretized function. The evaluator computes the discrete autoconvolution of this vector as a numerical approximation to the continuous autoconvolution
\begin{equation}
(f * f)(x) = \int_{\mathbb{R}} f(t)f(x-t)\,dt.
\end{equation}
The resulting vector is scored using a discretized version of
\begin{equation}
C_2(f) = \frac{\lVert f * f\rVert_2^2}
{\lVert f * f\rVert_1\lVert f * f\rVert_\infty}.
\end{equation}
The search therefore seeks a non-negative vector whose autoconvolution has a favorable balance among its $L_1$, $L_2$, and $L_\infty$ norms. The reported task score is a benchmark-normalized version of this quantity.

\subsection{Qualitative Strategies on Circle Packing}
The circle packing task reveals several qualitative strategy transitions. Because its constraints are local and pairwise, all three models can improve geometric templates, but they differ in whether they treat the task as direct construction or as an explicit optimization problem.

\paragraph{Qwen2.5-3B.}
Qwen2.5-3B mostly remains in a ``better template plus better local heuristic'' regime. It often starts with a ring-style placement of circle centers and attempts to improve it through clipping changes, simple spacing adjustments, and greedy radius-update rules rather than reformulating the task as a solver-based search problem. The first score boundary, around $0.63$, appears to be a template-viability wall: successful crossings usually come from tightening clipping bounds and stabilizing naive overlap handling. Around $0.68$, the model reaches a heuristic-refinement regime in which modest improvements in spread, clipping, and radius updates can help. Roughly $0.73$ is a persistent ceiling. Above this range, the model still tends to edit the packing directly instead of discovering the stronger abstraction that centers should be searched systematically and radii solved globally.

\paragraph{Qwen3-4B.}
Qwen3-4B shows stronger geometric reasoning. Compared with Qwen2.5-3B, it more often abandons weak ring templates in favor of hexagonal, staggered-row, symmetric, or carefully balanced layouts. The broad $0.77$--$0.84$ region is therefore less a single sharp wall than a transition from weaker constructions to denser row and hex templates. Around $0.91$, successful crossings are commonly associated with stronger row or hex geometry, better balance across rows, and more deliberate structured layouts. Near $0.95$, Qwen3-4B reaches its main ceiling: it can design strong dense packings, but usually does not make the next conceptual leap to linear-programming-based radius solving combined with systematic center search. Thus, Qwen3-4B largely replaces random jitter with near-optimal-looking geometric templates, but still tends to engineer the construction itself rather than formalize the underlying optimization problem.

\paragraph{GPT-OSS-20B.}
GPT-OSS-20B is the first model that consistently treats circle packing as an optimization problem rather than only a construction problem. Its strongest solutions decompose the task into choosing candidate centers, solving the best radii globally, and then searching over center positions using hill climbing, annealing, restarts, or related local-search procedures. Around the $0.95$ boundary, explicit optimization begins to matter substantially, with successful crossings adding stronger search behavior on top of already reasonable layouts. The clearest high-score wall appears near $0.9645$, where many runs plateau until they combine LP-based radius solving with better center search, stronger hexagonal or staggered seeds, and larger restart or annealing budgets. Above this wall, near $0.98$, the final improvements often come from moving beyond LP-only radius solving toward joint optimization of centers and radii, for example through SLSQP-style nonlinear optimization. This decomposition of choose centers, solve radii, then search centers, is the key abstraction that allows GPT-OSS-20B to keep improving after the smaller models plateau.

\subsection{Qualitative Strategies on Heilbronn Triangle}
The Heilbronn triangle task also reveals qualitative differences in the constructions discovered by each model. These differences help explain the separated baseline distributions in Figure~\ref{fig:heilbronn_triangle_sequential_baseline}.

\paragraph{Qwen2.5-3B.}
Qwen2.5-3B primarily relies on shallow, local modifications. Its successful solutions are often based on random perturbations, edge or vertex placement heuristics, and simple triangle-aware sampling. The observed score boundaries suggest that moving past roughly $0.096$ often requires placing points on edges or vertices, while crossing roughly $0.115$ typically requires some form of triangle-aware sampling. In the $0.20$--$0.45$ range, successful crossings are dominated by random-perturbation heuristics. This pattern suggests that Qwen2.5-3B improves mostly through micro-optimizations and stochastic nudges rather than through large structural changes in the point configuration.

\paragraph{Qwen3-4B.}
Moving from Qwen2.5-3B to Qwen3-4B produces a visible shift from random jitter toward more structured geometric reasoning. Qwen3-4B more often proposes row, ring, lattice, or symmetric triangular layouts. Around the $0.5$ score boundary, row and ring lattice constructions become common, and near $0.61$ many successful crossings appear to rely on known symmetric lattice templates. At higher thresholds, around $0.72$, the successful configurations increasingly resemble edge/interior templates. One possible interpretation is that the larger Qwen model has stronger access to geometric priors or memorized construction patterns, allowing it to abandon purely random perturbation in favor of near-structured triangular arrangements with small degeneracy-breaking offsets.

\paragraph{GPT-OSS-20B.}
GPT-OSS-20B shows the clearest evidence of phase transitions in strategy. In the lower and middle score ranges, especially around boundaries such as $0.201$, $0.372$, and $0.384$, successful crossings are often associated with introducing an explicit objective optimizer. At the high end, around $0.865$, $0.893$, and $0.925$, the model more frequently relies on annealing-style updates and random restarts. This suggests a two-stage behavior: GPT-OSS-20B first uses direct optimization to reach the mid-range, then switches to temperature-based acceptance and restart budgets to escape local minima. These strategy transitions suggest that performance gains are not merely due to generating more samples. Stronger models appear to discover qualitatively different algorithmic mechanisms for crossing specific score thresholds.

\subsection{Qualitative Strategies on the Second Autocorrelation Inequality}

Unlike the geometry tasks, this task improves an optimization program rather than a visible construction. Successful programs represent the function as a non-negative vector, compute its autoconvolution with an FFT, and optimize the vector with gradients. The main strategic differences concern the initial function shape, the discretization resolution, and the numerical choices that determine whether optimization reaches a useful solution.

\paragraph{Qwen2.5-3B.}
Qwen2.5-3B mostly scales the baseline optimizer rather than redesigning it. It increases the number of vector entries and gradient steps, then gradually replaces random or flat initializations with piecewise profiles containing zero regions, plateaus, and ramps. Its progression is therefore from direct coordinate optimization toward finer discretization with a modest prior over the initial function shape.

\paragraph{Qwen3-4B.}
Qwen3-4B introduces stronger structural hypotheses. Its programs often begin with symmetric piecewise-constant functions and evolve toward segmented or self-similar profiles with non-uniform amplitudes. Some lineages also relax symmetry to test asymmetric solutions. The key transition is from treating the vector as independent coordinates to treating it as a structured function whose segments, amplitudes, and symmetry can be designed before gradient refinement.

\paragraph{GPT-OSS-20B.}
GPT-OSS-20B focuses on conditioning the optimization process. It changes positivity transforms, learning rates, clipping, regularization, and initialization so that the vector can develop sharp step-like regions without unstable gradients. This also explains its broad, approximately bimodal baseline distribution. All 30 plotted runs use the same harness and initial code, but they separate into a smooth low-score basin and a sharper high-score basin. Low-mode programs typically use small learning rates and strong stabilization, while high-mode programs make a larger mutation that relaxes these constraints or introduces a stronger step-like initialization. All nine final programs with learning rate at least $0.1$ enter the high mode, whereas eleven of the sixteen programs with learning rate at most $0.02$ remain in the low mode. This suggests that GPT-OSS-20B alternates between two optimizer regimes rather than gradually moving through one continuous family of solutions.
\FloatBarrier

\section{Full Search-Harness Ablation Results}\label{sec:detailed_search_harness_ablations}

\subsection{Additional OpenEvolve Progression Results}

\begin{figure*}[t]
  \centering
  \begin{subfigure}[t]{0.49\textwidth}
    \centering
    \appendixprogressiongraphics{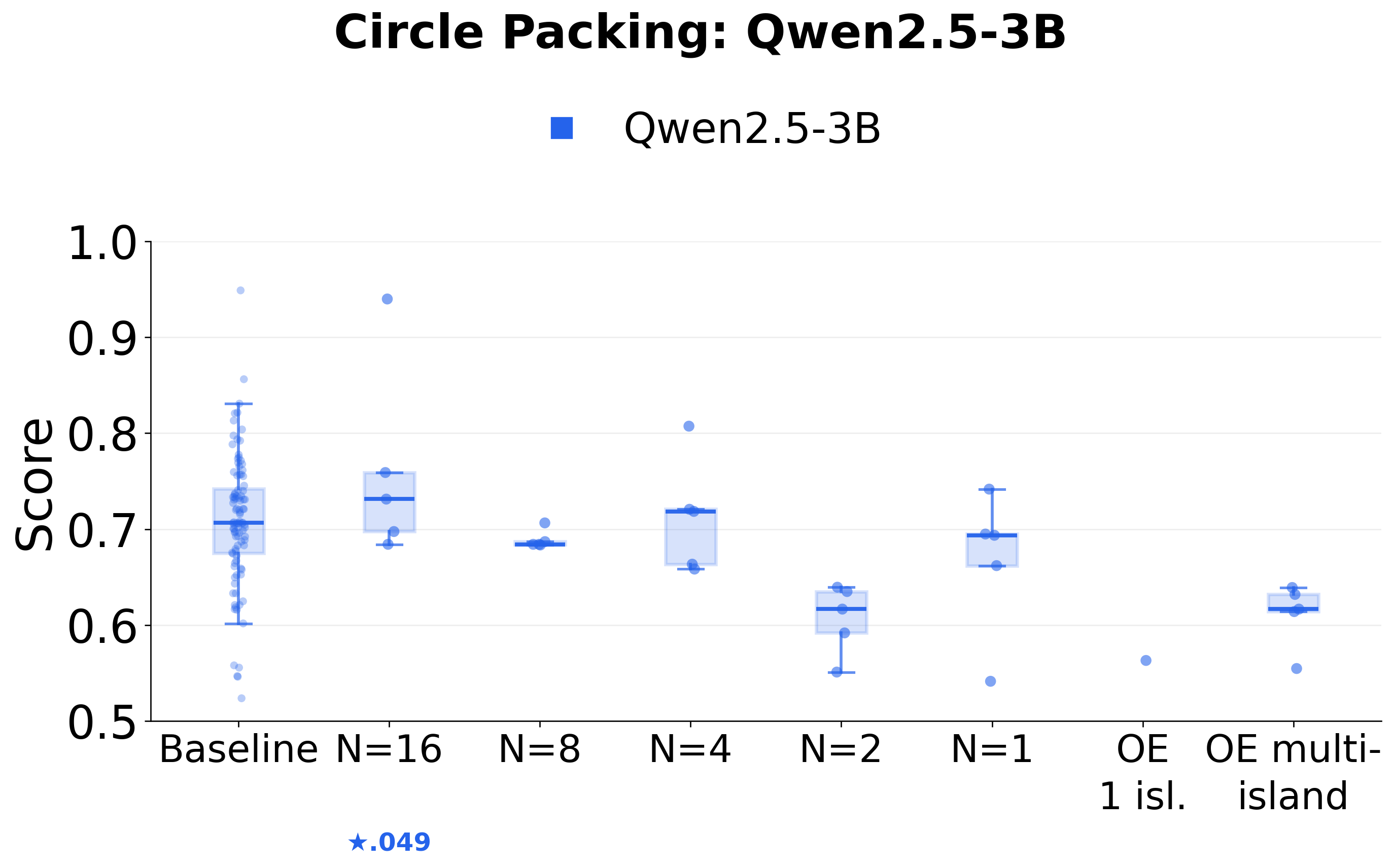}
    \caption{Qwen2.5-3B-Instruct.}
    \label{fig:qwen25_circle_openevolve_progression}
  \end{subfigure}
  \hfill
  \begin{subfigure}[t]{0.49\textwidth}
    \centering
    \appendixprogressiongraphics{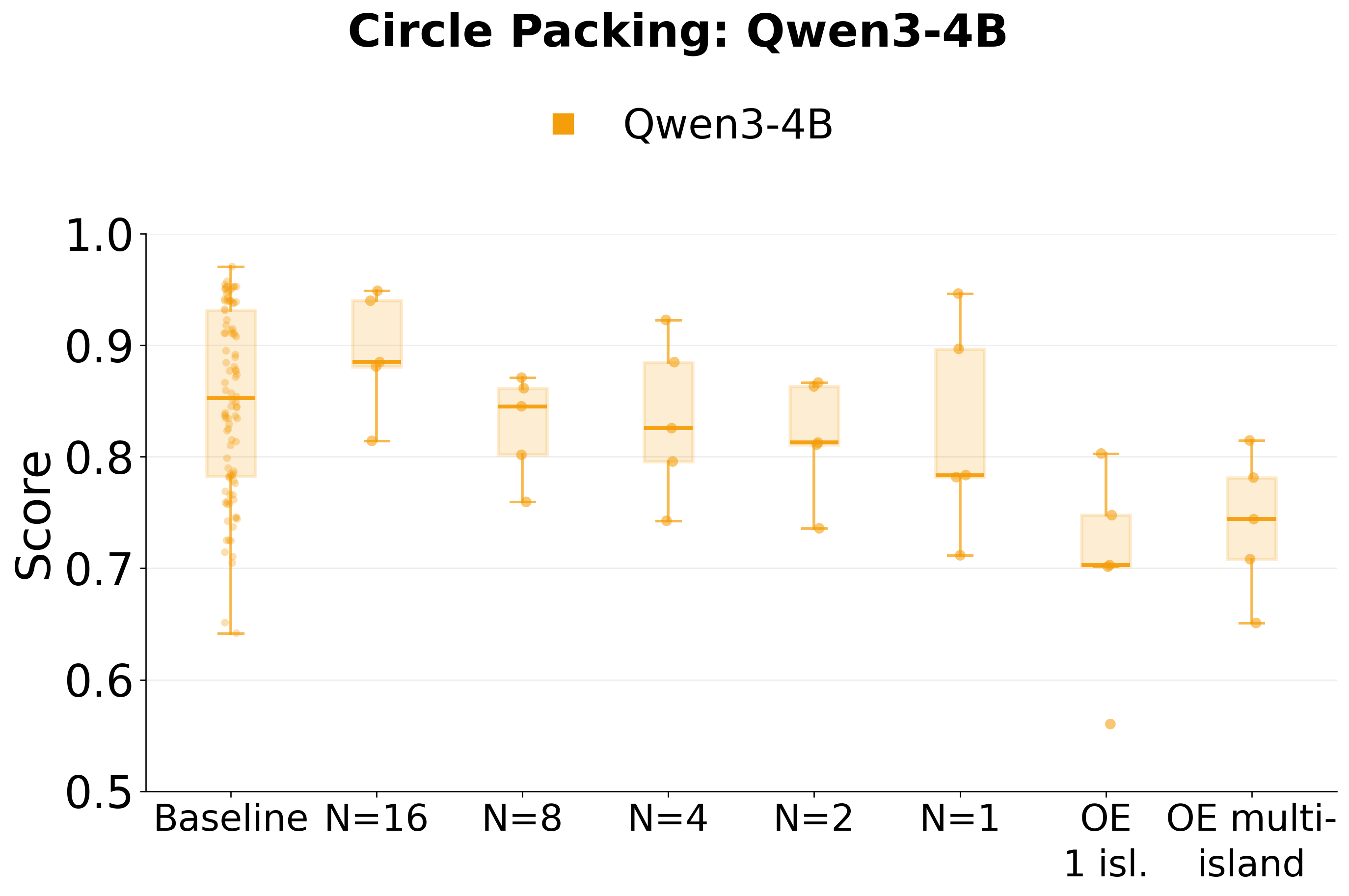}
    \caption{Qwen3-4B-Instruct.}
    \label{fig:qwen3_circle_openevolve_progression}
  \end{subfigure}
  \vspace{0.5em}

  \begin{subfigure}[t]{0.49\textwidth}
    \centering
    \appendixprogressiongraphics{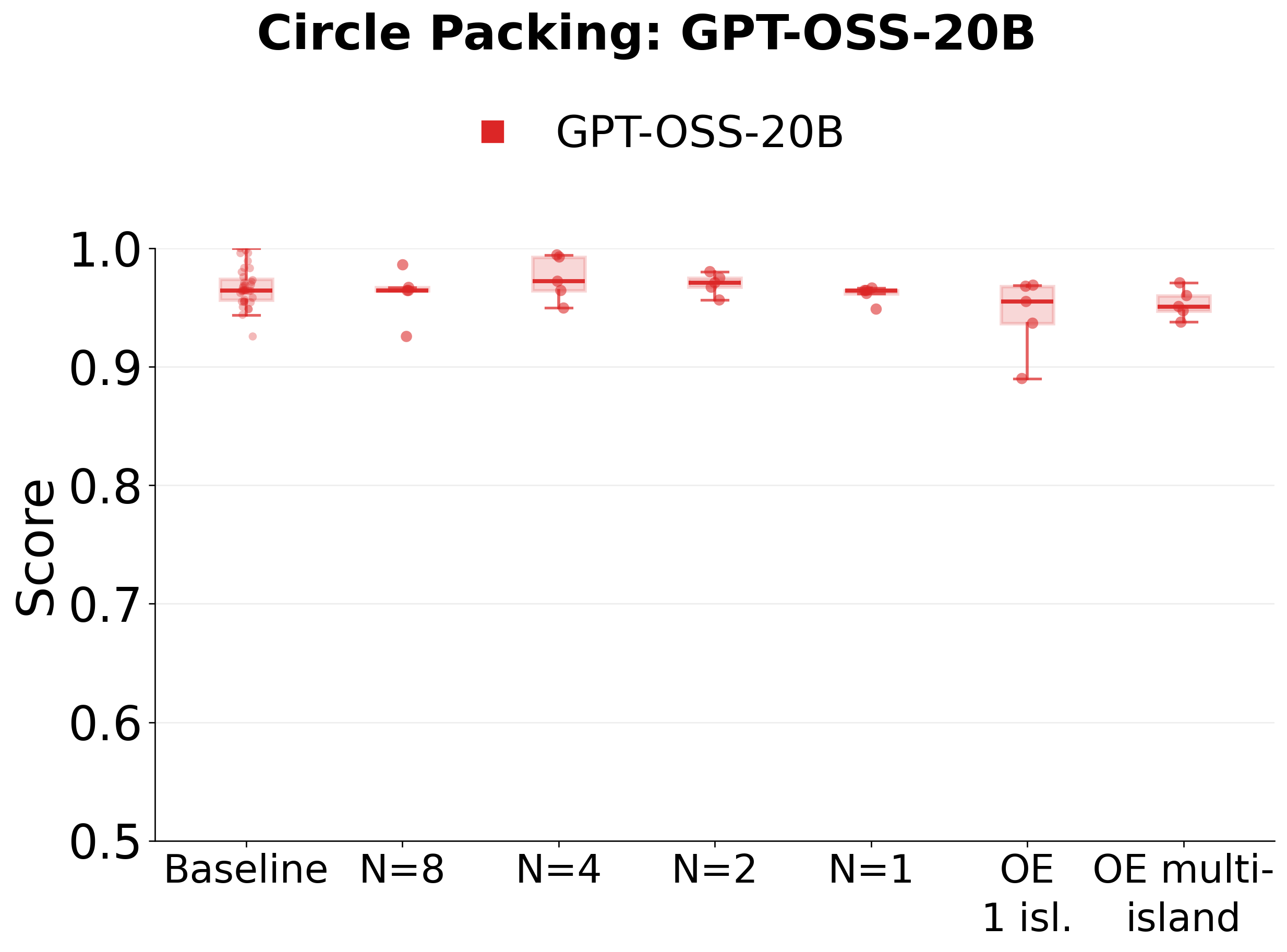}
    \caption{GPT-OSS-20B.}
    \label{fig:gptoss_circle_openevolve_progression}
  \end{subfigure}
  \hfill
  \begin{subfigure}[t]{0.49\textwidth}
    \centering
    \appendixprogressiongraphics{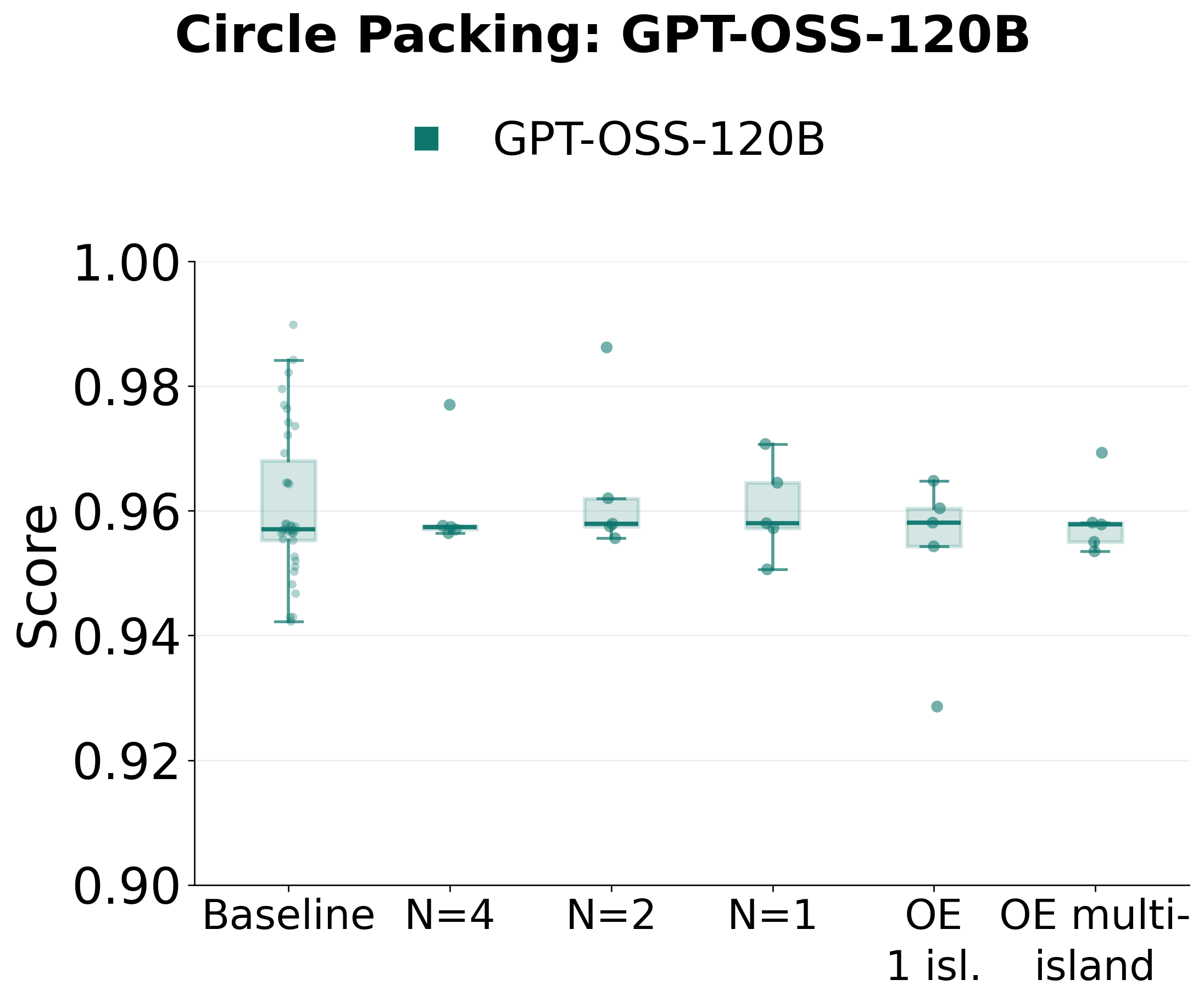}
    \caption{GPT-OSS-120B.}
    \label{fig:gptoss120_circle_openevolve_progression}
  \end{subfigure}
  \caption{\textbf{Progression toward OpenEvolve-style search on circle packing.} The four panels show the model-specific breadth--depth progression and available OpenEvolve variants.}
  \label{fig:openevolve_progression_all_models}
\end{figure*}

\begin{figure*}[t]\ContinuedFloat
  \centering
  \begin{subfigure}[t]{0.49\textwidth}
    \centering
    \appendixprogressiongraphics{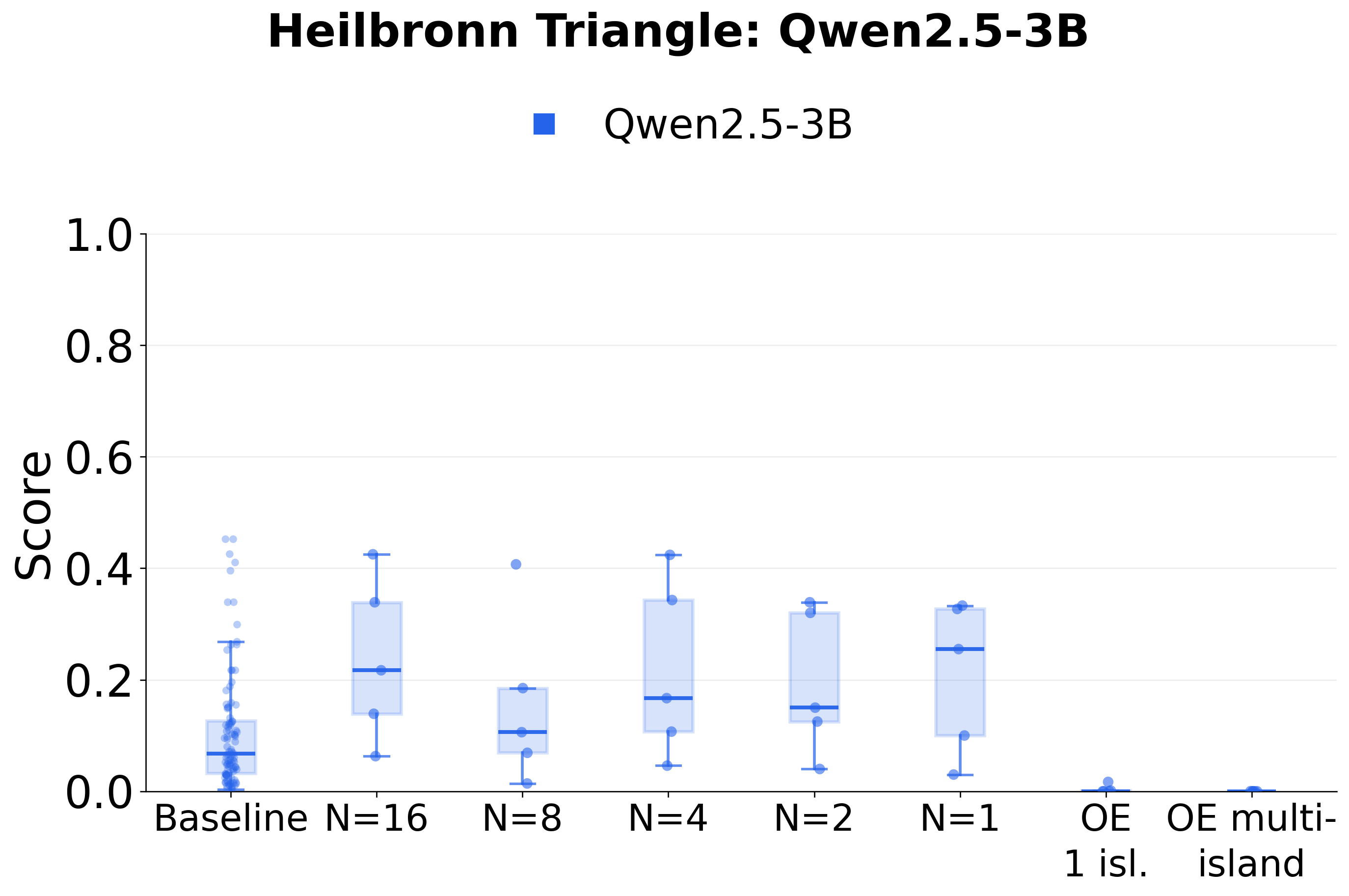}
    \caption{Qwen2.5-3B-Instruct.}
    \label{fig:qwen25_heilbronn_openevolve_progression}
  \end{subfigure}
  \hfill
  \begin{subfigure}[t]{0.49\textwidth}
    \centering
    \appendixprogressiongraphics{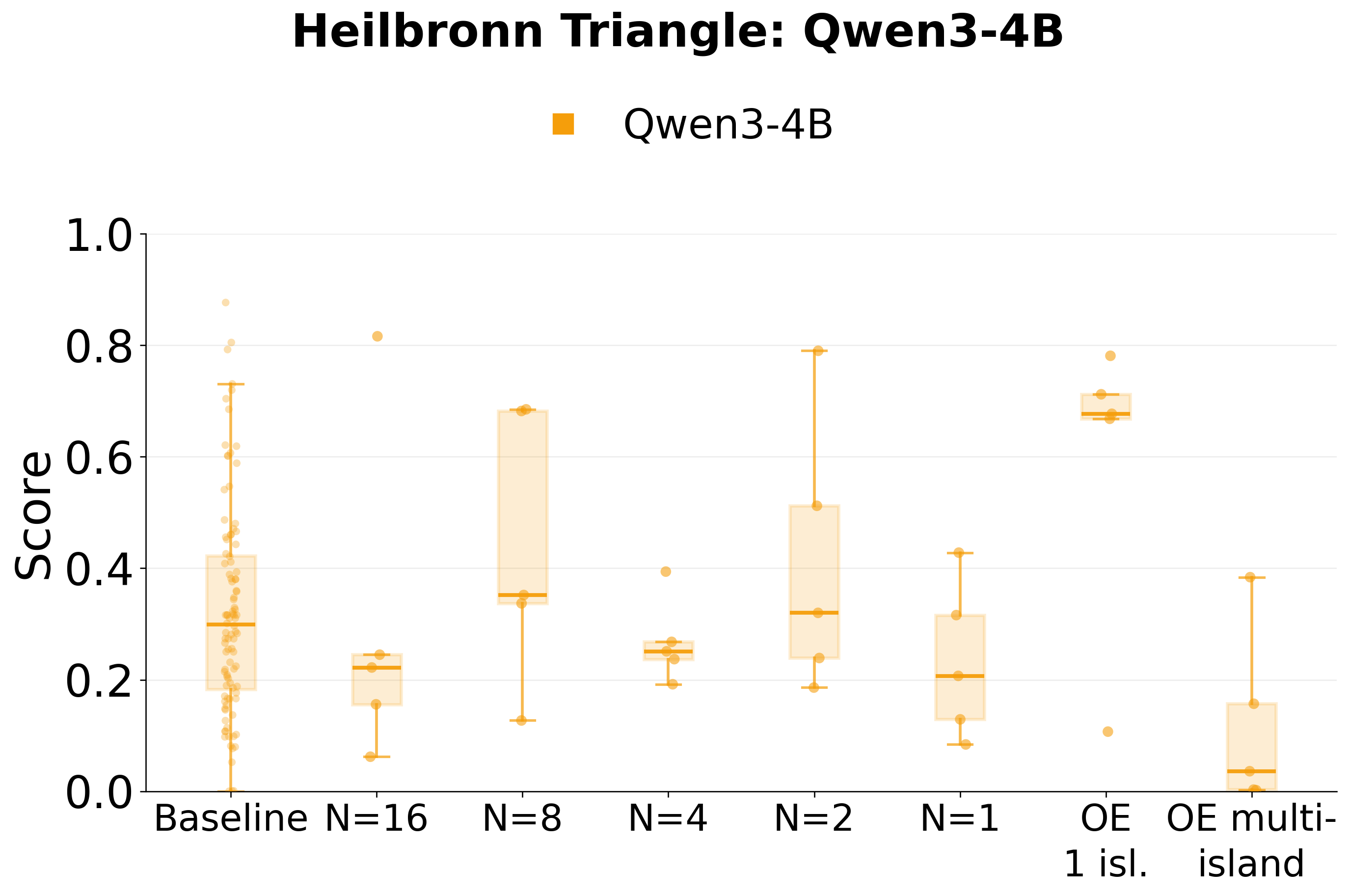}
    \caption{Qwen3-4B-Instruct.}
    \label{fig:qwen3_heilbronn_openevolve_progression}
  \end{subfigure}
  \vspace{0.5em}

  \begin{subfigure}[t]{0.49\textwidth}
    \centering
    \appendixprogressiongraphics{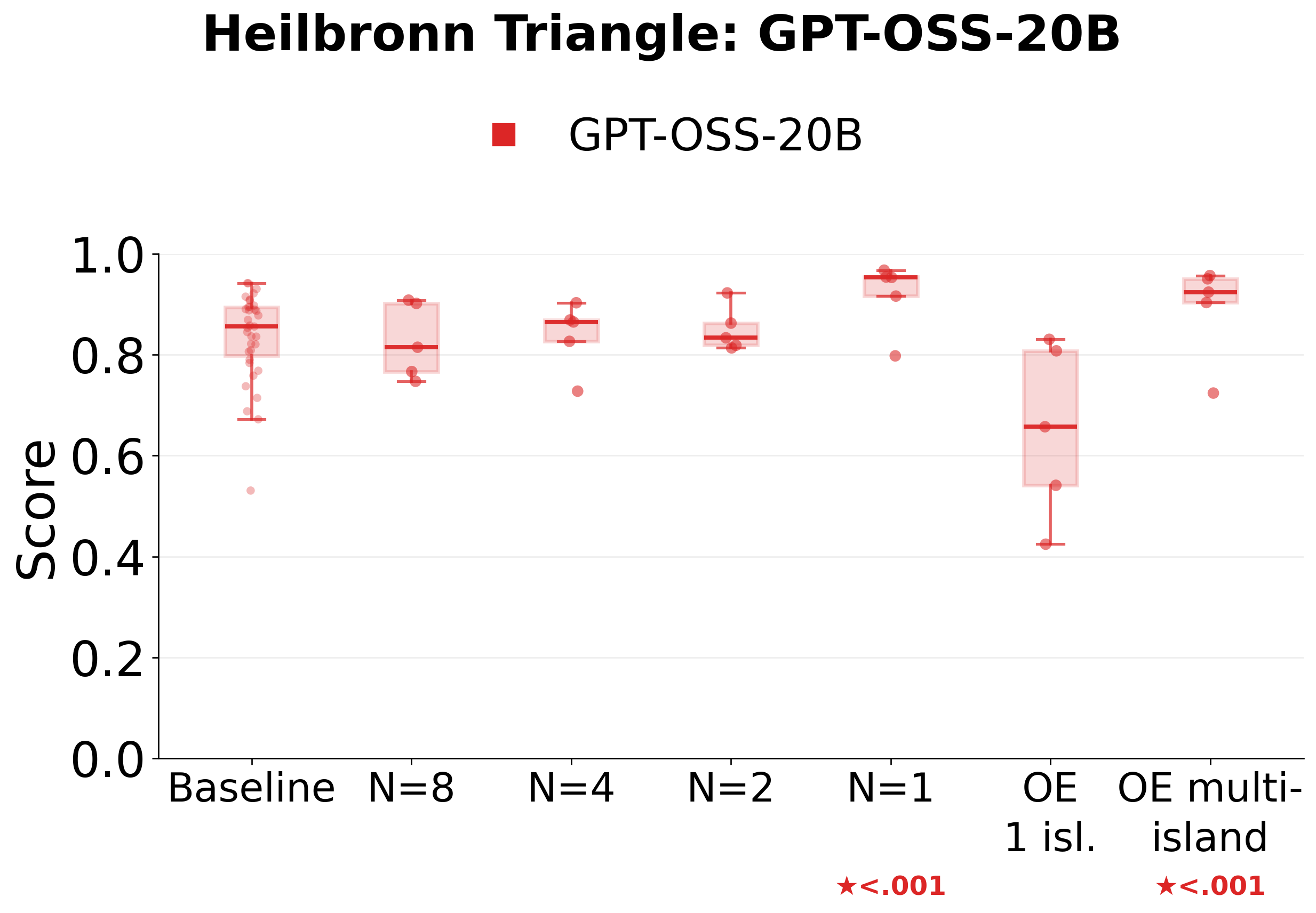}
    \caption{GPT-OSS-20B.}
    \label{fig:gptoss_heilbronn_openevolve_progression}
  \end{subfigure}
  \hfill
  \begin{subfigure}[t]{0.49\textwidth}
    \centering
    \appendixprogressiongraphics{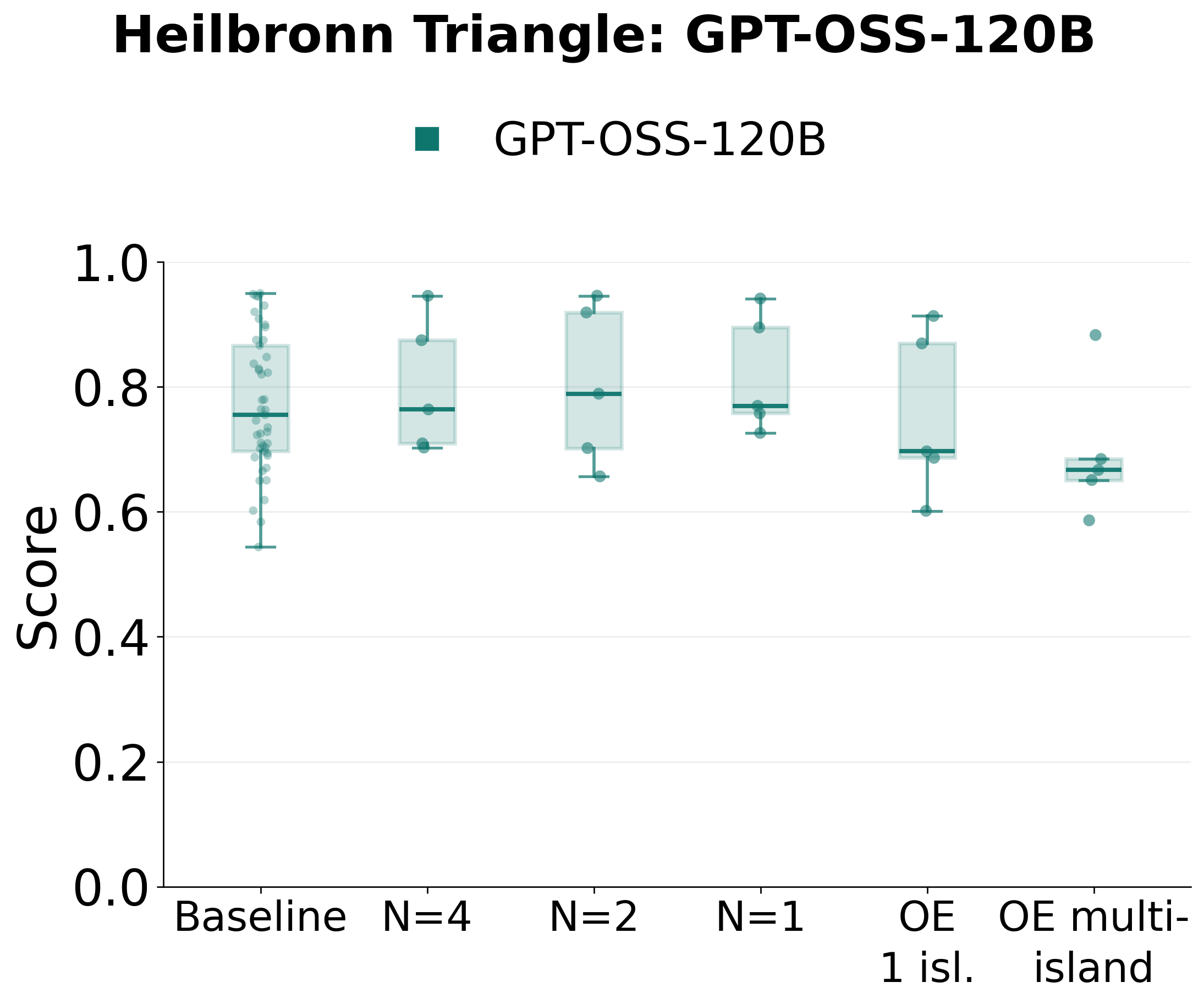}
    \caption{GPT-OSS-120B.}
    \label{fig:gptoss120_heilbronn_openevolve_progression}
  \end{subfigure}
  \caption{\textbf{Progression toward OpenEvolve-style search on the Heilbronn triangle task, continued.}}
\end{figure*}

\begin{figure*}[t]\ContinuedFloat
  \centering
  \begin{subfigure}[t]{0.49\textwidth}
    \centering
    \appendixprogressiongraphics{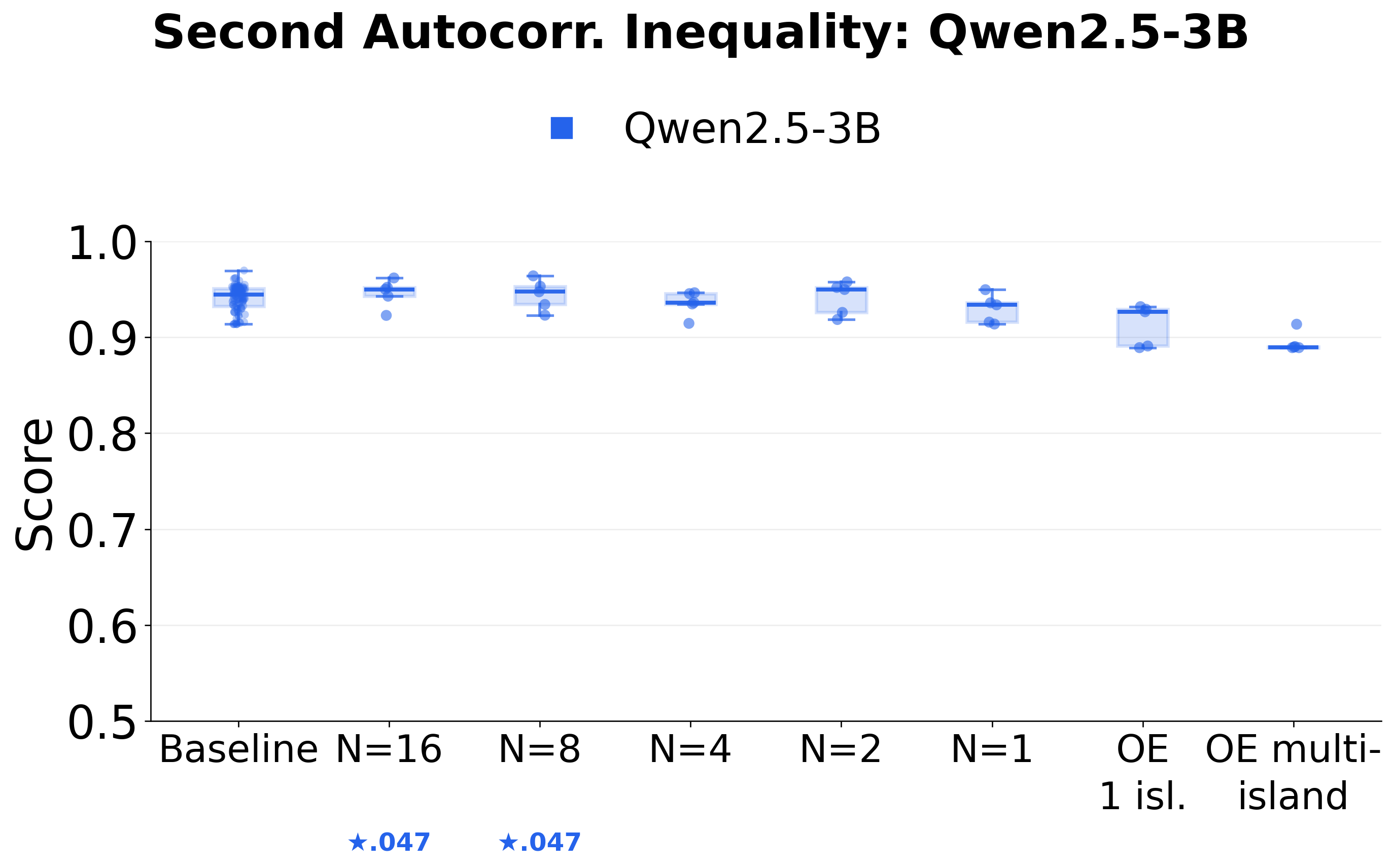}
    \caption{Qwen2.5-3B-Instruct.}
    \label{fig:qwen25_second_autocorr_openevolve_progression}
  \end{subfigure}
  \hfill
  \begin{subfigure}[t]{0.49\textwidth}
    \centering
    \appendixprogressiongraphics{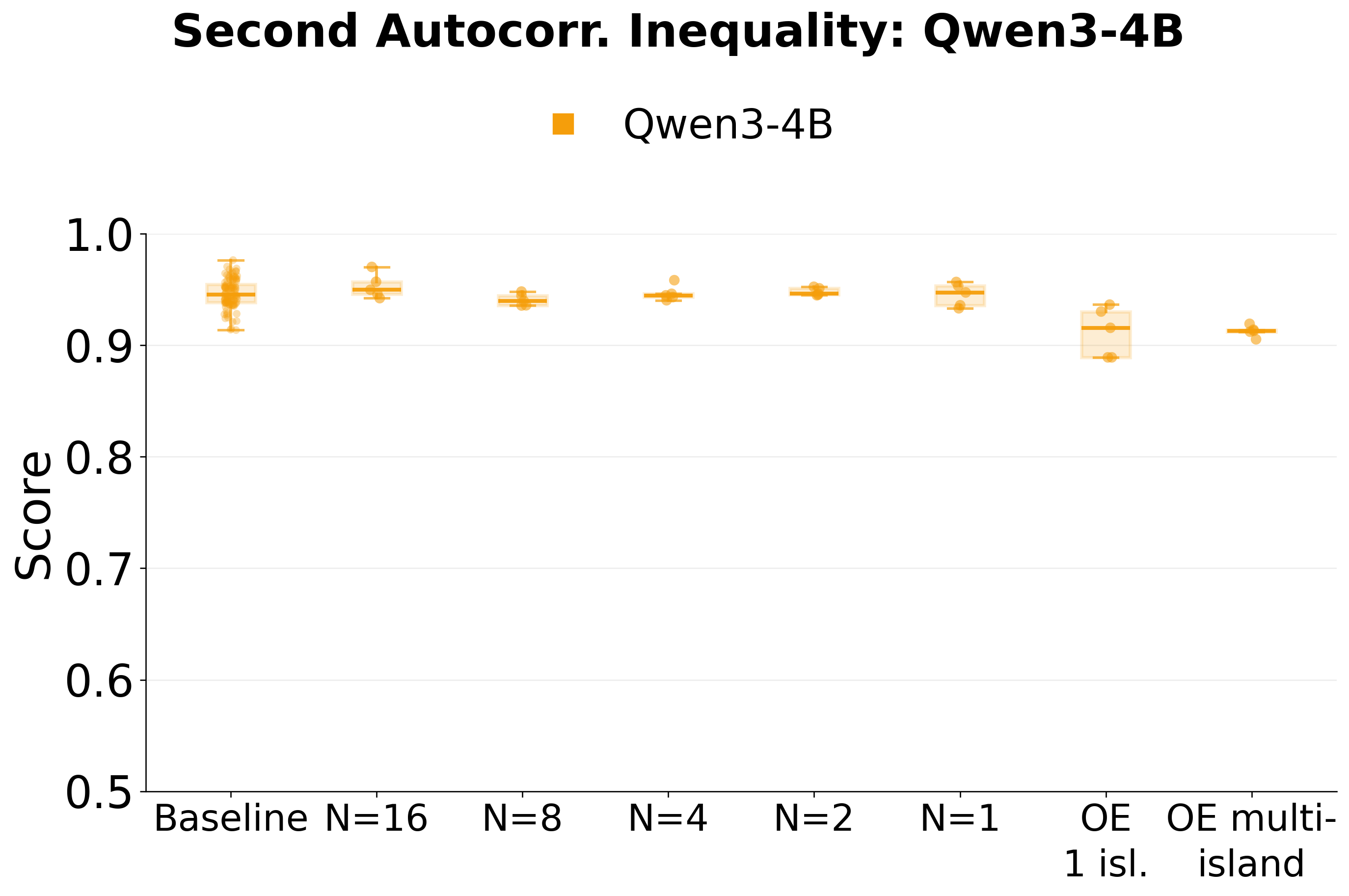}
    \caption{Qwen3-4B-Instruct.}
    \label{fig:qwen3_second_autocorr_openevolve_progression}
  \end{subfigure}
  \vspace{0.5em}

  \begin{subfigure}[t]{0.49\textwidth}
    \centering
    \appendixprogressiongraphics{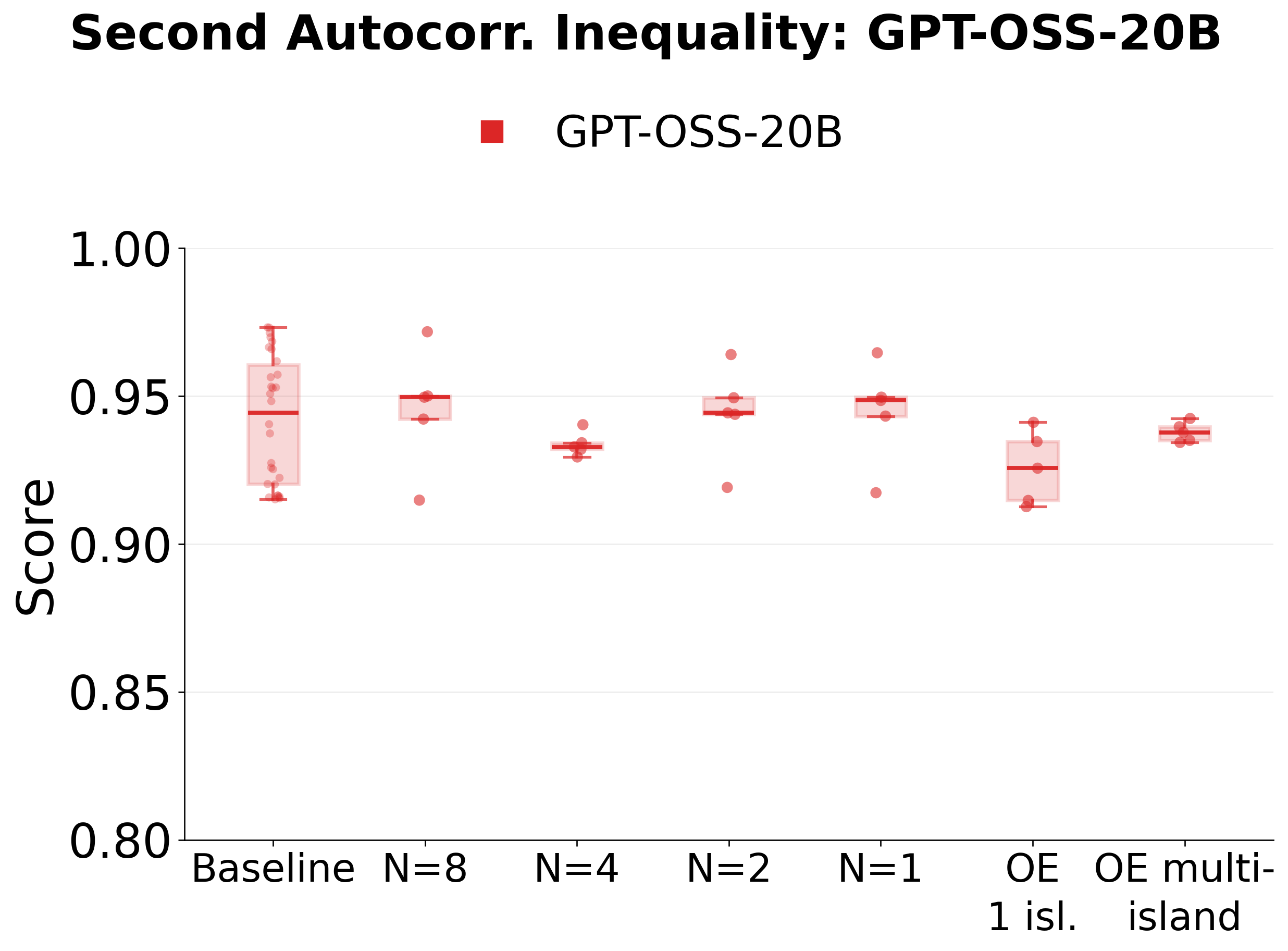}
    \caption{GPT-OSS-20B.}
    \label{fig:gptoss_second_autocorr_openevolve_progression}
  \end{subfigure}
  \hfill
  \begin{subfigure}[t]{0.49\textwidth}
    \centering
    \appendixprogressiongraphics{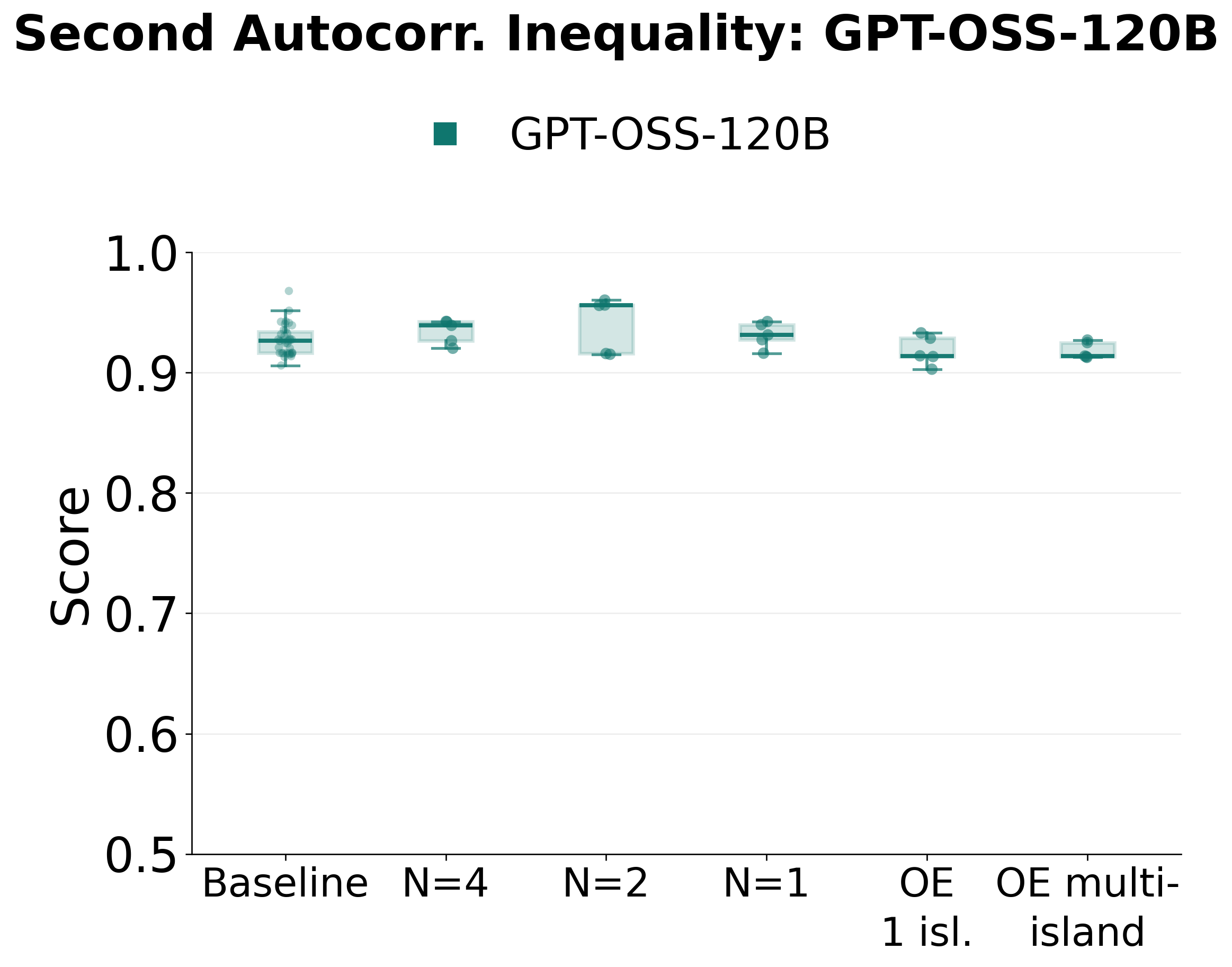}
    \caption{GPT-OSS-120B.}
    \label{fig:gptoss120_second_autocorr_openevolve_progression}
  \end{subfigure}
  \caption{\textbf{Progression toward OpenEvolve-style search on the second autocorrelation inequality task, continued.} Intermediate breadth--depth allocations can be competitive, while the OpenEvolve endpoint does not uniformly dominate across models and tasks.}
\end{figure*}

\FloatBarrier
\clearpage

\subsection{Full Epsilon-Greedy Results}

\begin{figure}[!htbp]
  \centering
  \begin{subfigure}[t]{0.95\textwidth}
    \centering
    \appendixfullgridgraphics{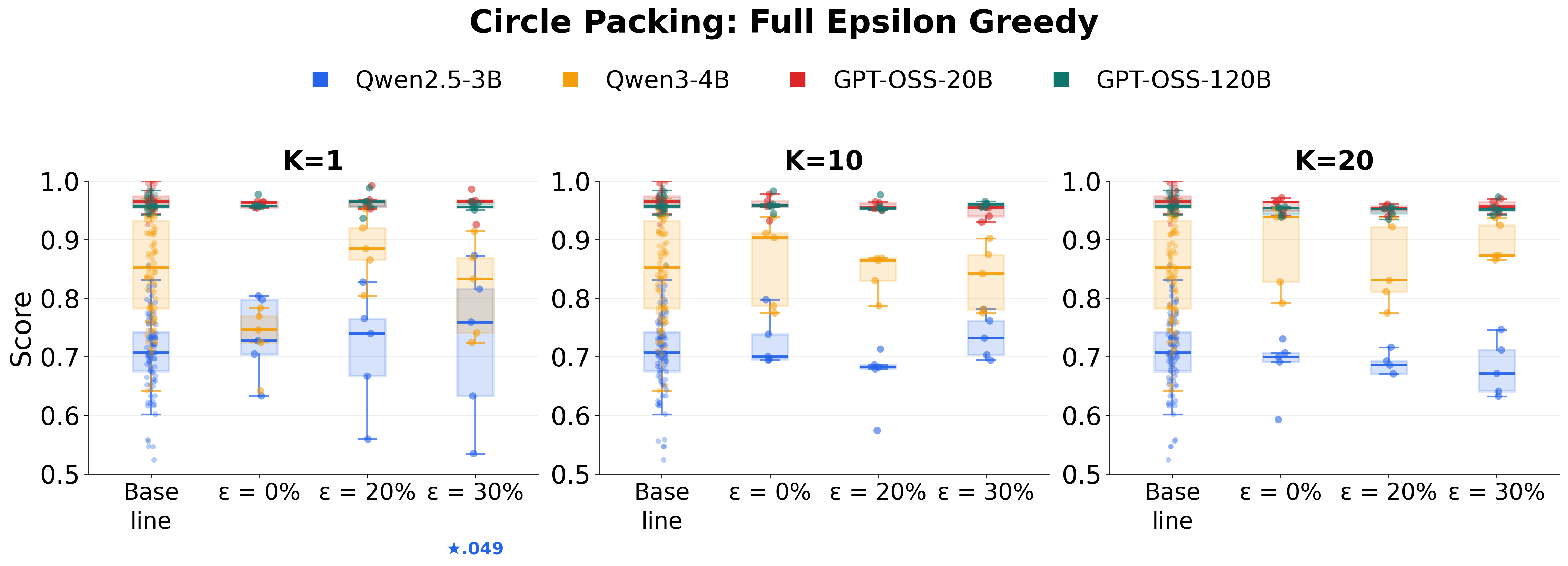}
    \caption{Circle packing.}
    \label{fig:circle_packing_full_epsilon_greedy}
  \end{subfigure}

  \vspace{0.2em}

  \begin{subfigure}[t]{0.95\textwidth}
    \centering
    \appendixfullgridgraphics{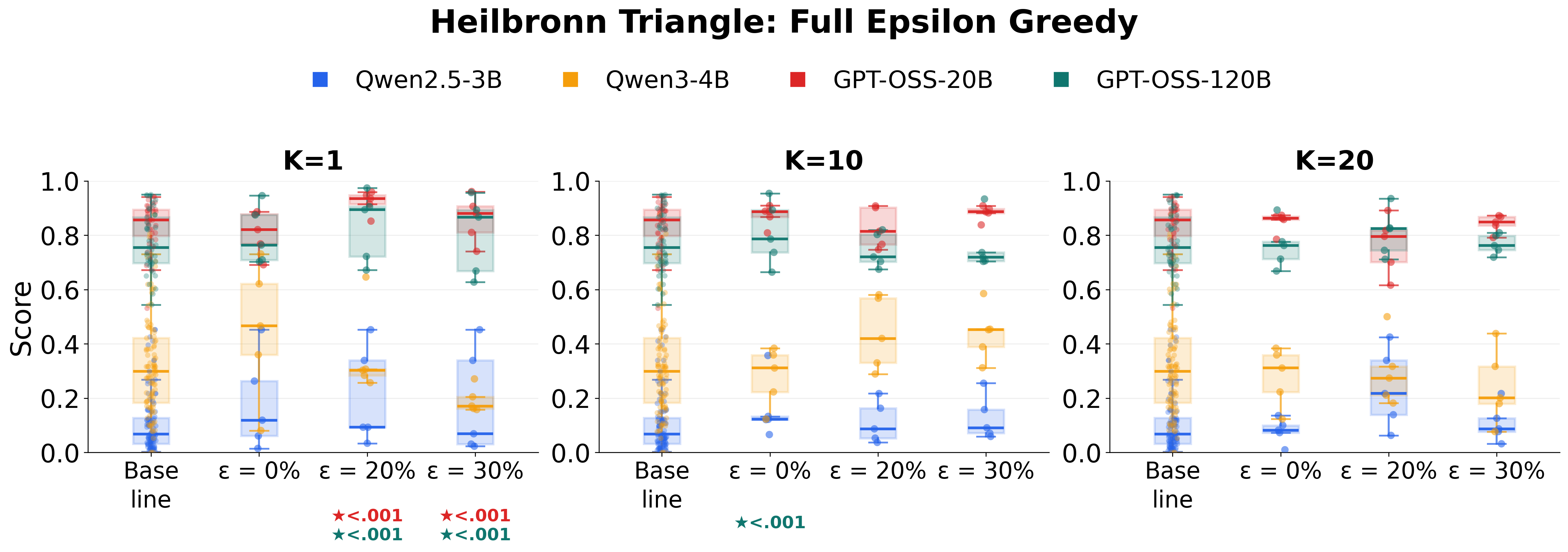}
    \caption{Heilbronn triangle.}
    \label{fig:heilbronn_triangle_full_epsilon_greedy}
  \end{subfigure}

  \vspace{0.2em}

  \begin{subfigure}[t]{0.95\textwidth}
    \centering
    \appendixfullgridgraphics{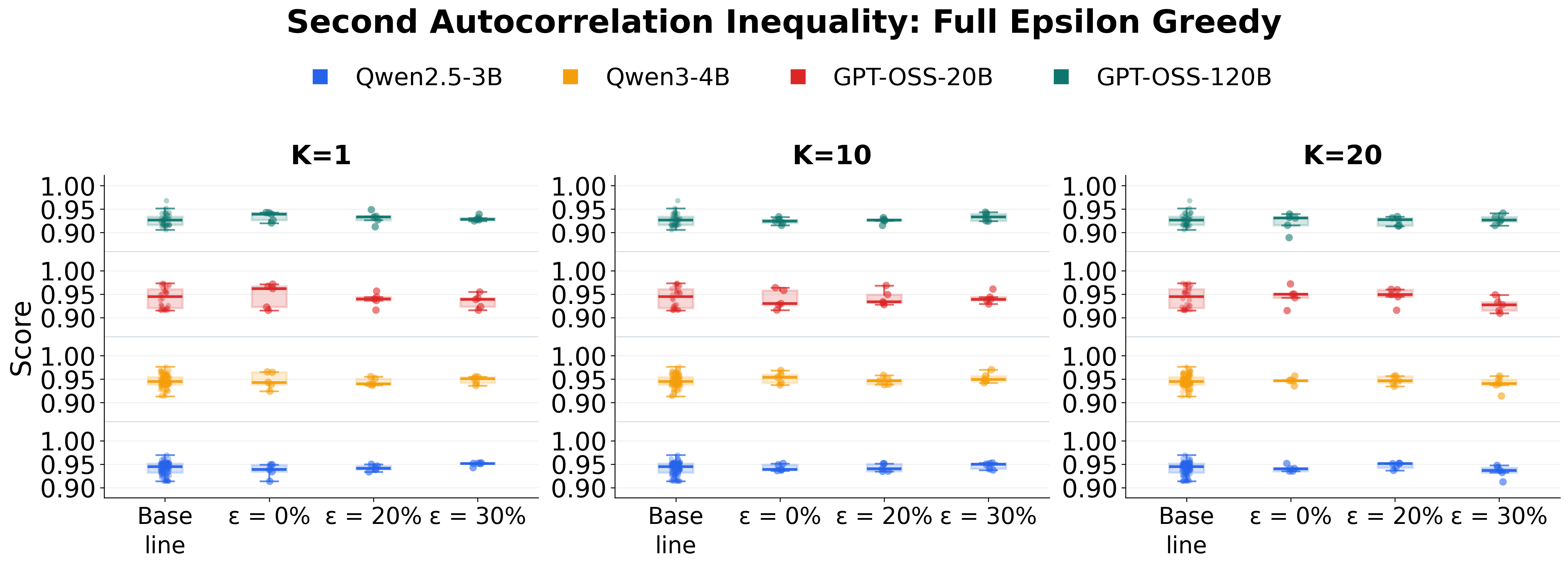}
    \caption{Second autocorrelation inequality.}
    \label{fig:second_autocorrelation_inequality_full_epsilon_greedy}
  \end{subfigure}

  \caption{\textbf{Full epsilon-greedy grid across all three tasks.} Each panel varies elite archive size and $\epsilon$ probability, with model-specific score distributions overlaid.}
  \label{fig:main_full_epsilon_greedy_summary}
\end{figure}

\FloatBarrier

\subsection{Sequential BoN Baseline}
\begin{figure}[t]
  \centering
  \begin{subfigure}[t]{0.32\textwidth}
    \centering
    \includegraphics[width=\linewidth]{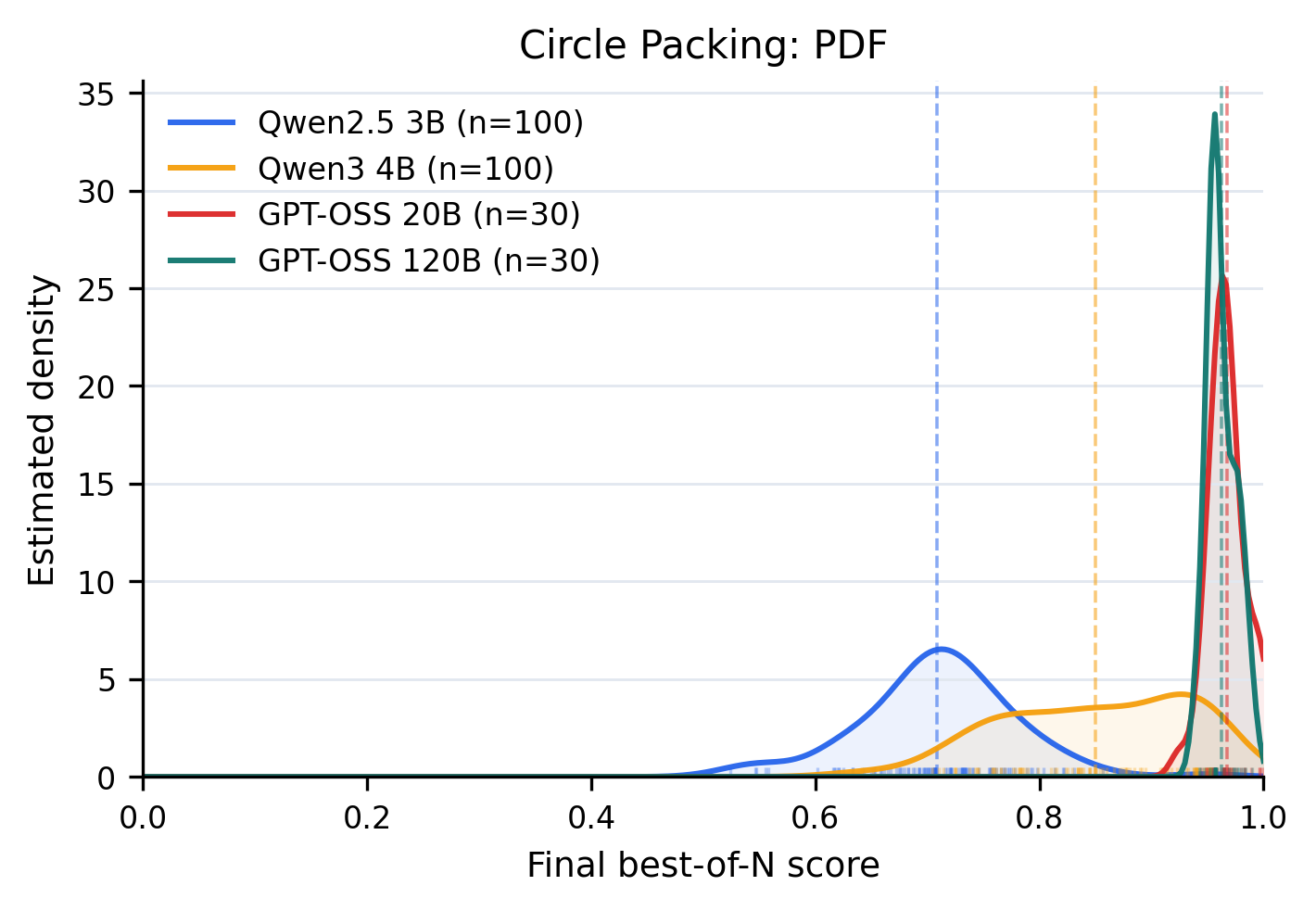}
    \caption{Circle packing.}
    \label{fig:circle_packing_sequential_baseline}
  \end{subfigure}
  \hfill
  \begin{subfigure}[t]{0.32\textwidth}
    \centering
    \includegraphics[width=\linewidth]{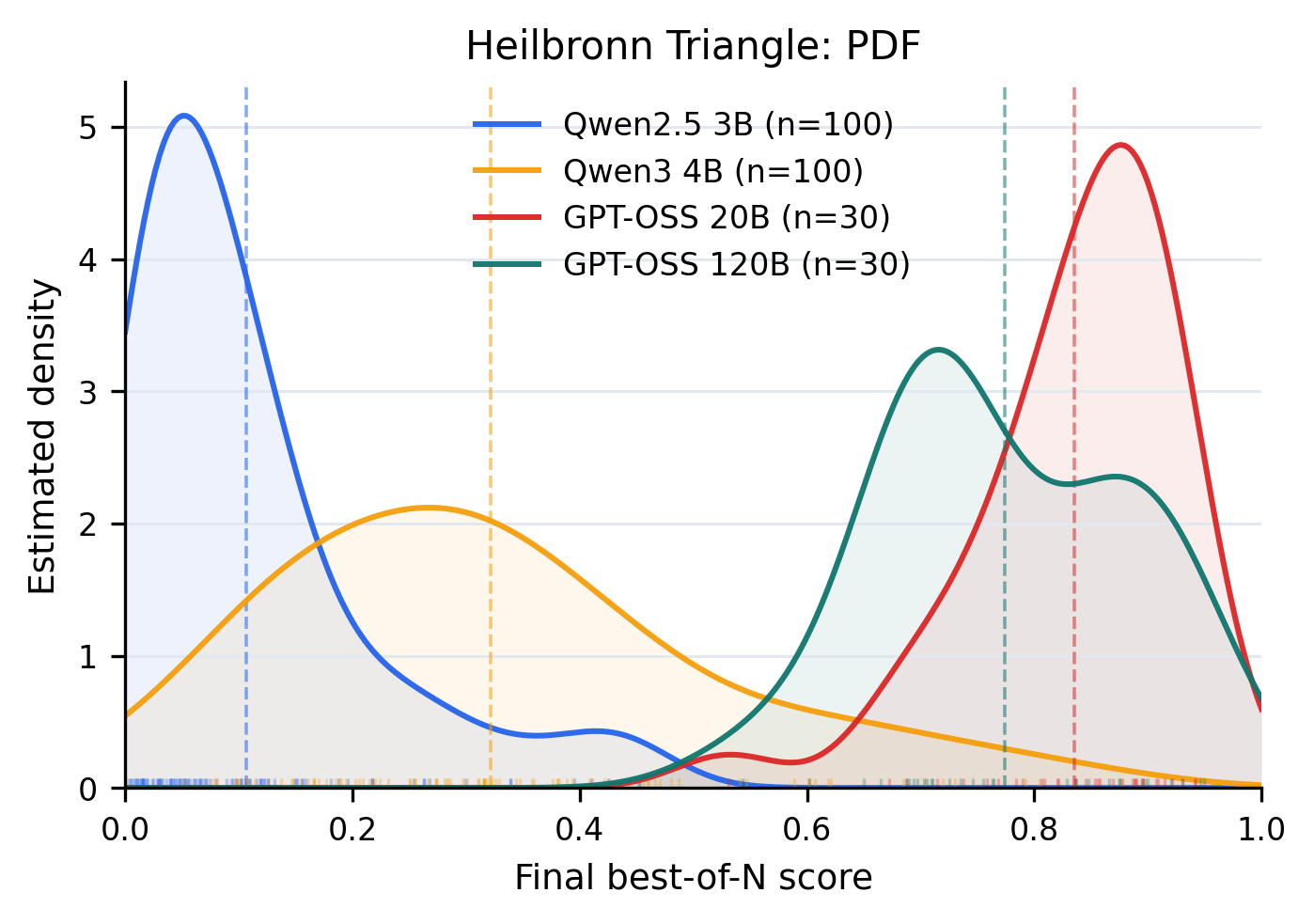}
    \caption{Heilbronn triangle.}
    \label{fig:heilbronn_triangle_sequential_baseline}
  \end{subfigure}
  \hfill
  \begin{subfigure}[t]{0.32\textwidth}
    \centering
    \includegraphics[width=\linewidth]{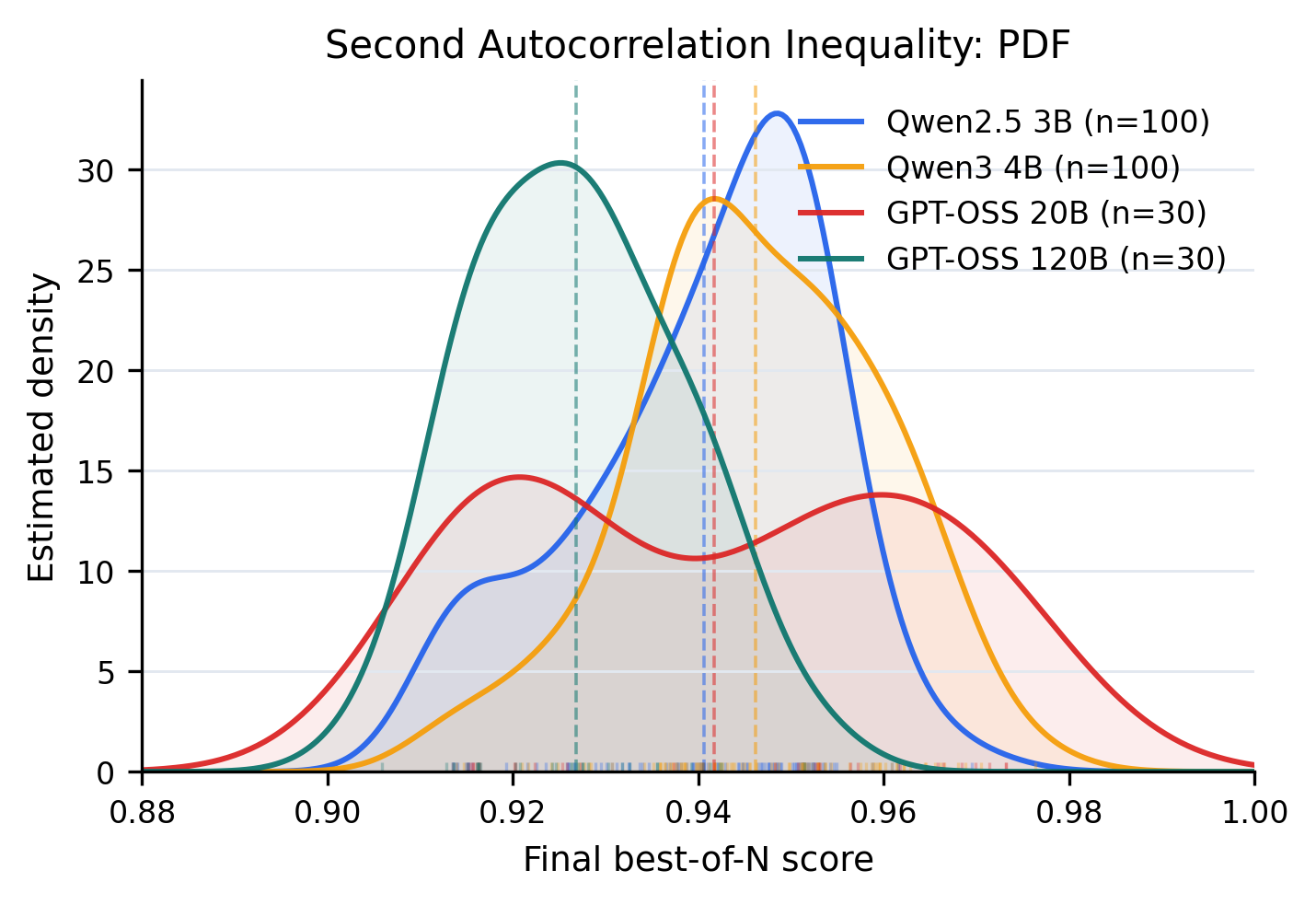}
    \caption{Second autocorrelation inequality.}
    \label{fig:second_autocorrelation_sequential_baseline}
  \end{subfigure}
  \caption{\textbf{Sequential BoN baseline distributions.} Kernel-density estimates show final scores for each task and model. Dashed vertical lines mark model-specific means, and rug marks show individual runs. These 100-run distributions define the empirical reference used in the bootstrap comparisons.}
  \label{fig:sequential_baseline_kdes}
\end{figure}

% Plotting and task analysis for Sequential BoN baselines:
% 1. Include one panel for each task.
% 2. Show score distributions for the three models in different colors.
% 3. Use the score distributions to compare model behavior and baseline score ranges.
We first evaluate a Sequential Best-of-$N$ (BoN) baseline for each model--problem pair. Sequential BoN is the $K=1$, $\epsilon=0$ special case of the archive formulation in Equation~\ref{eq:epsilon-greedy-parent-selection}: it always expands the best program found so far and never samples from the non-elite history. We use the baseline score distribution (30 runs for GPT-OSS models, and 100 runs for Qwen models) as the reference for the archive, exploration, UCT/PUCT, and OpenEvolve-style comparisons. Figure~\ref{fig:sequential_baseline_kdes} shows kernel-density estimates of the final baseline scores.

The baseline distributions already reveal a large model effect before any search-harness modification is introduced. On circle packing, GPT-OSS-20B is tightly concentrated near the top of the score range, indicating both strong performance and low variance. Qwen3-4B reaches high-scoring regions but has a broader distribution, whereas Qwen2.5-3B is centered substantially lower. Consequently, GPT-OSS-20B has less headroom for visible improvement, while the Qwen models leave more room for a search method to stabilize or redirect weak runs.

For Heilbronn triangle, the separation between models is even more pronounced. GPT-OSS-20B again forms a high-score mode, while Qwen2.5-3B is concentrated near very low scores and Qwen3-4B occupies a broad middle range. The wider Qwen3-4B distribution suggests that it occasionally finds useful solutions but does so inconsistently. On the second autocorrelation inequality, the three models are much closer in score, although GPT-OSS-20B has a broader distribution that reflects its distinct low- and high-score optimizer regimes. Overall, these baseline KDEs provide an important reference point: later search algorithms should be interpreted relative to both the model's average baseline performance and the spread of its stochastic runs, since improvements for a high-performing, low-variance model mean something different from improvements for a lower-performing, high-variance model.

\subsection{Adding an Elite Archive}
% Experimenting with side by side view, note to increase the size of the axis labels
\begin{figure*}[t]
  \centering
  \begin{subfigure}[t]{0.49\textwidth}
    \centering
    \appendixsweepgraphics{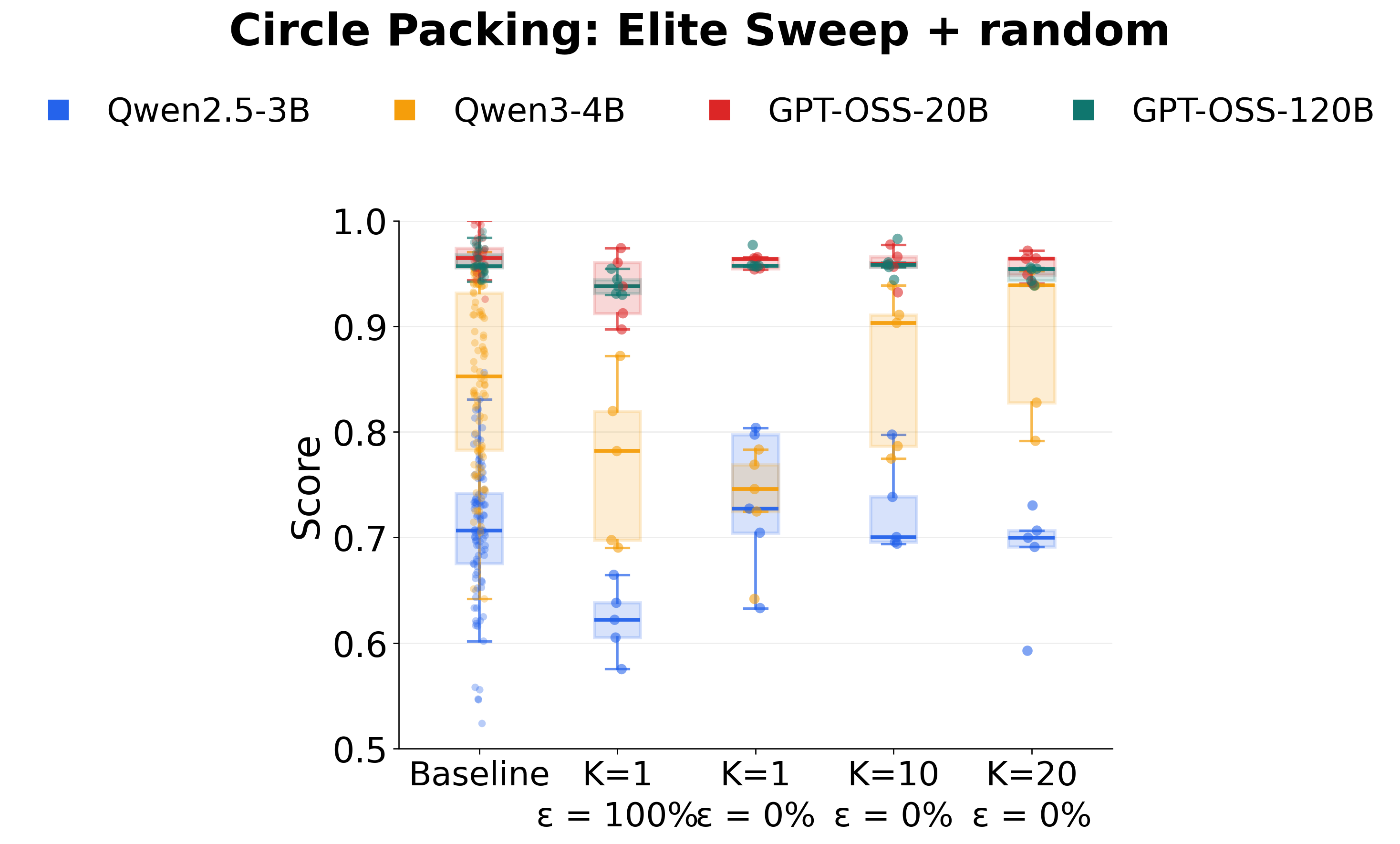}
    \caption{Circle packing.}
    \label{fig:circle_packing_elite_archive}
  \end{subfigure}
  \hfill
  \begin{subfigure}[t]{0.49\textwidth}
    \centering
    \appendixsweepgraphics{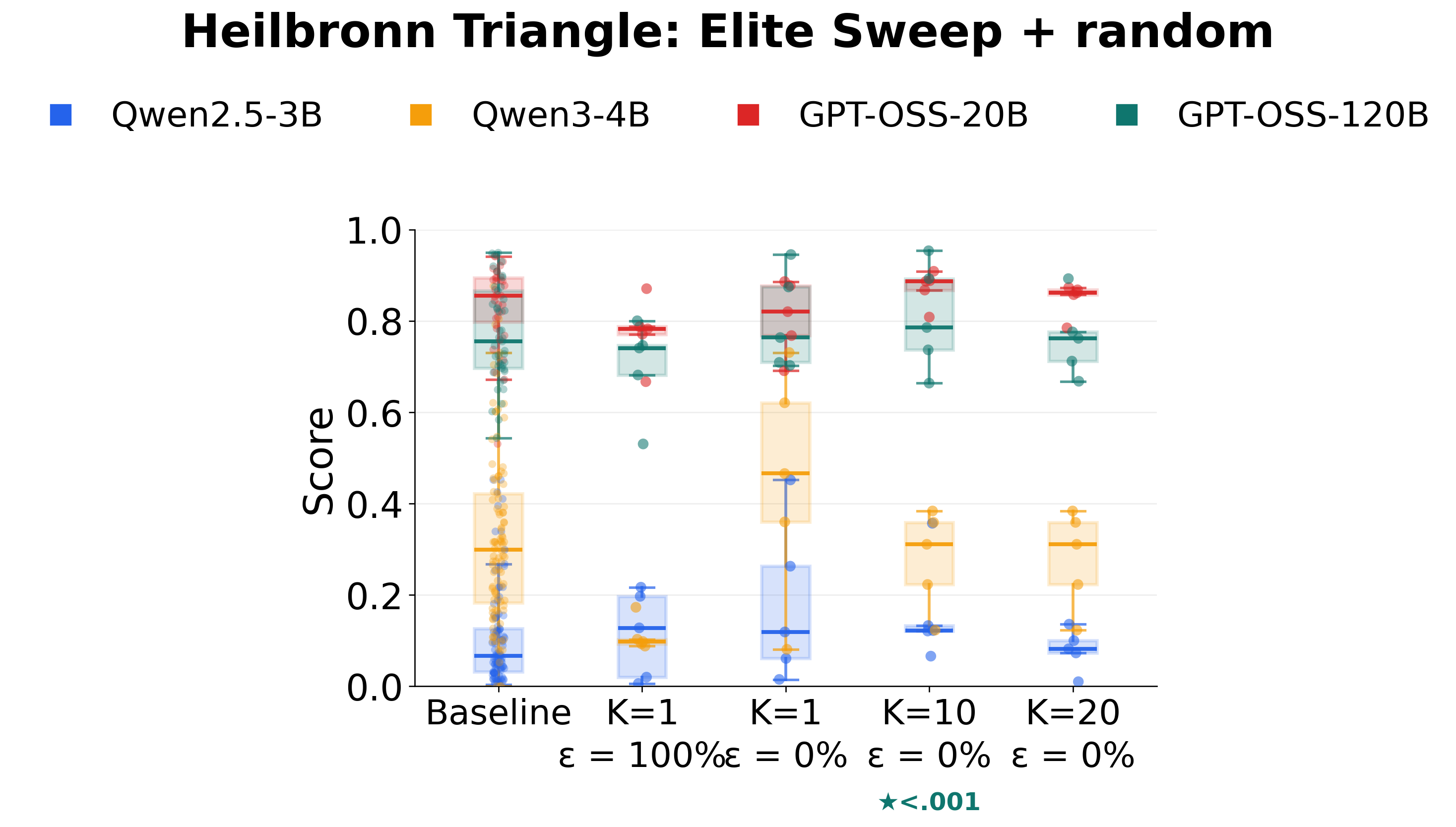}
    \caption{Heilbronn triangle.}
    \label{fig:heilbronn_triangle_elite_archive}
  \end{subfigure}
  \vspace{0.5em}

  \begin{subfigure}[t]{0.7\textwidth}
    \centering
    \appendixsweepgraphics{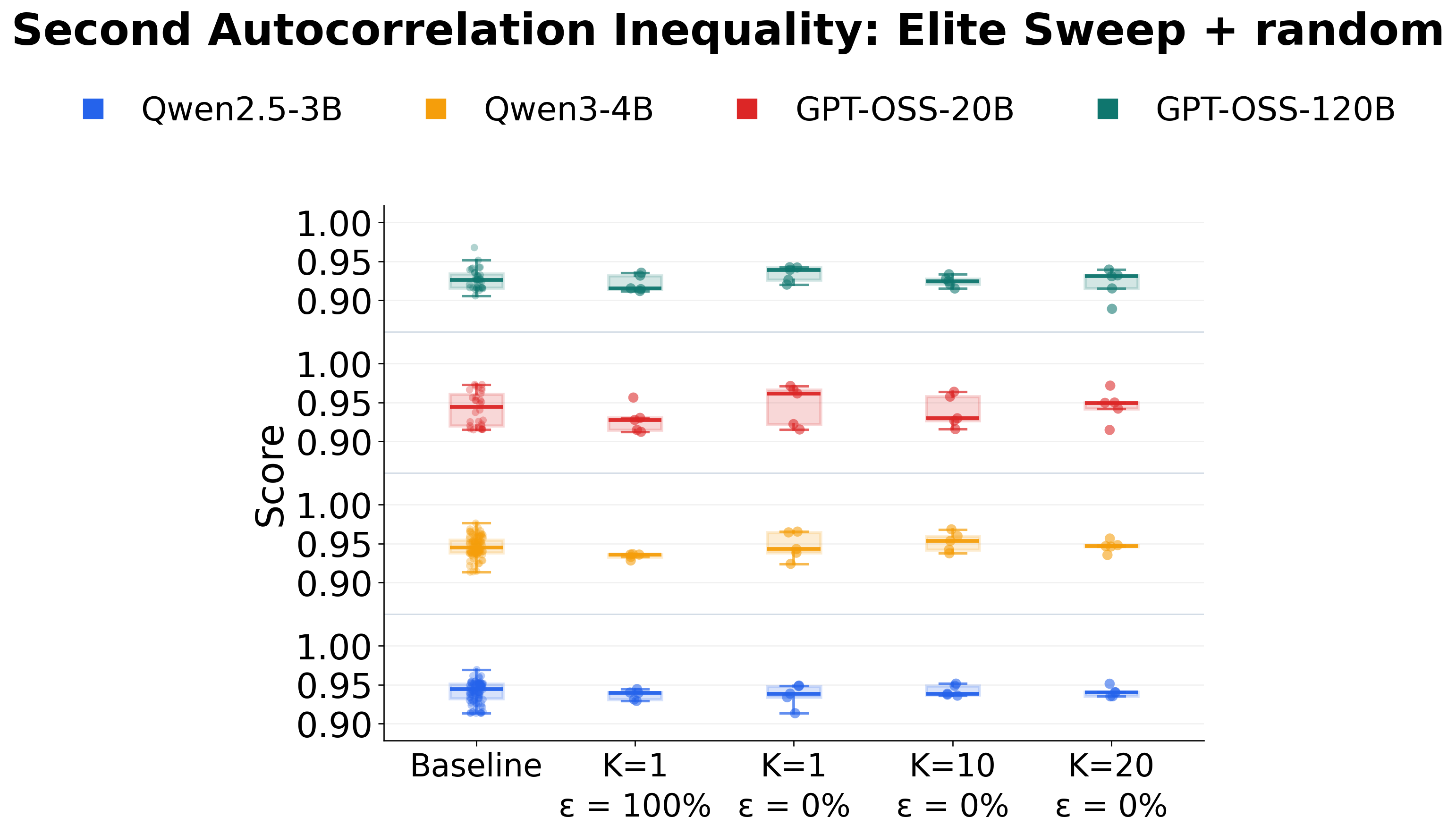}
    \caption{Second Autocorrelation Inequality.}
    \label{fig:second_autocorrelation_inequality_elite_archive}
  \end{subfigure}
  \caption{\textbf{Elite-archive sweep across all three tasks.} Parent programs are sampled from a top-$K$ archive under $\epsilon = 0\%$. Colored stars denote configurations whose five-run maximum significantly exceeds the bootstrapped Sequential BoN reference. The absence of a star indicates insufficient evidence under this test, not evidence of no effect.}
  \label{fig:elite_archive_sweep}
\end{figure*}

\paragraph{Experimental setup.}
We next sample parent programs from the top-$K$ archive. We evaluate $K \in \{1,10,20,30\}$ with zero general-history exploration, and include the $K=1$ setting with full general-history sampling as a random-selection comparison. Sequential BoN is shown in the first column of each panel.

Figures~\ref{fig:circle_packing_elite_archive}, \ref{fig:heilbronn_triangle_elite_archive}, and~\ref{fig:second_autocorrelation_inequality_elite_archive} compare Sequential BoN with increasingly large elite archives. The central pattern is model--problem dependence rather than monotonic improvement with $K$. GPT-OSS-20B starts from a high baseline and remains near the top of the score range across archive settings, leaving limited headroom. Qwen3-4B moves more visibly, particularly on circle packing, where some archive settings shift the distribution upward or reduce its variance. Qwen2.5-3B remains more variable and shows weaker evidence of consistent archive-driven gains.

On Heilbronn triangle, the elite archive does not substantially change the overall ordering between models. GPT-OSS-20B remains dominant, while the Qwen models continue to exhibit lower and more variable score distributions. The second autocorrelation inequality is similarly stable across archive sizes. The distributions shift only modestly, with no consistent improvement as $K$ increases. This suggests that the elite archive is most useful when the model can already generate multiple plausible high-quality lineages. When candidate quality is poor or highly unstable, increasing $K$ alone does not reliably compensate.

\subsection{Adding Exploration}

\begin{figure*}[t]
  \centering
  \begin{subfigure}[t]{0.49\textwidth}
    \centering
    \appendixsweepgraphics{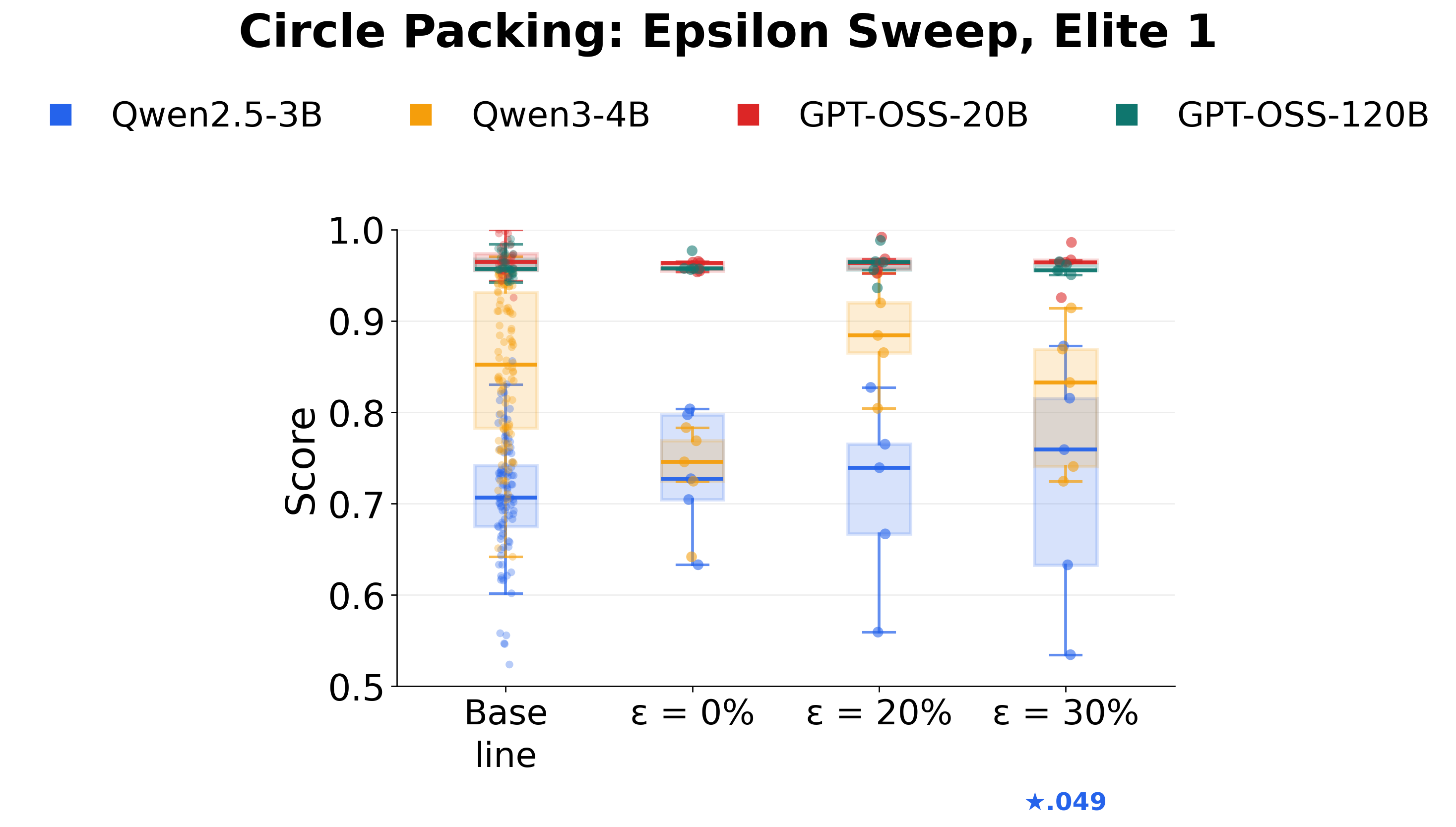}
    \caption{Circle packing.}
    \label{fig:circle_packing_exploration}
  \end{subfigure}
  \hfill
  \begin{subfigure}[t]{0.49\textwidth}
    \centering
    \appendixsweepgraphics{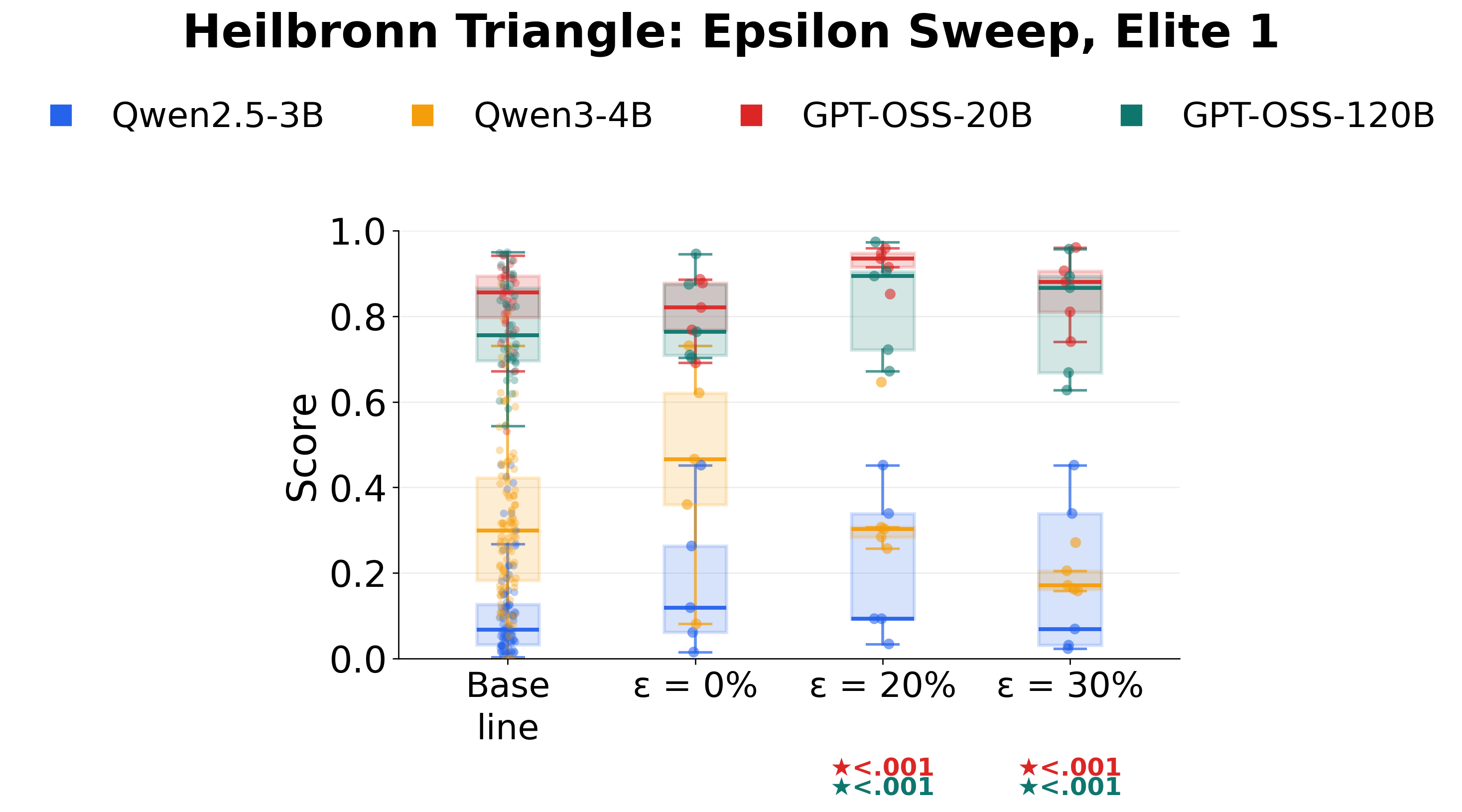}
    \caption{Heilbronn triangle.}
    \label{fig:heilbronn_triangle_exploration}
  \end{subfigure}
  \vspace{0.5em}

  \begin{subfigure}[t]{0.49\textwidth}
    \centering
    \appendixsweepgraphics{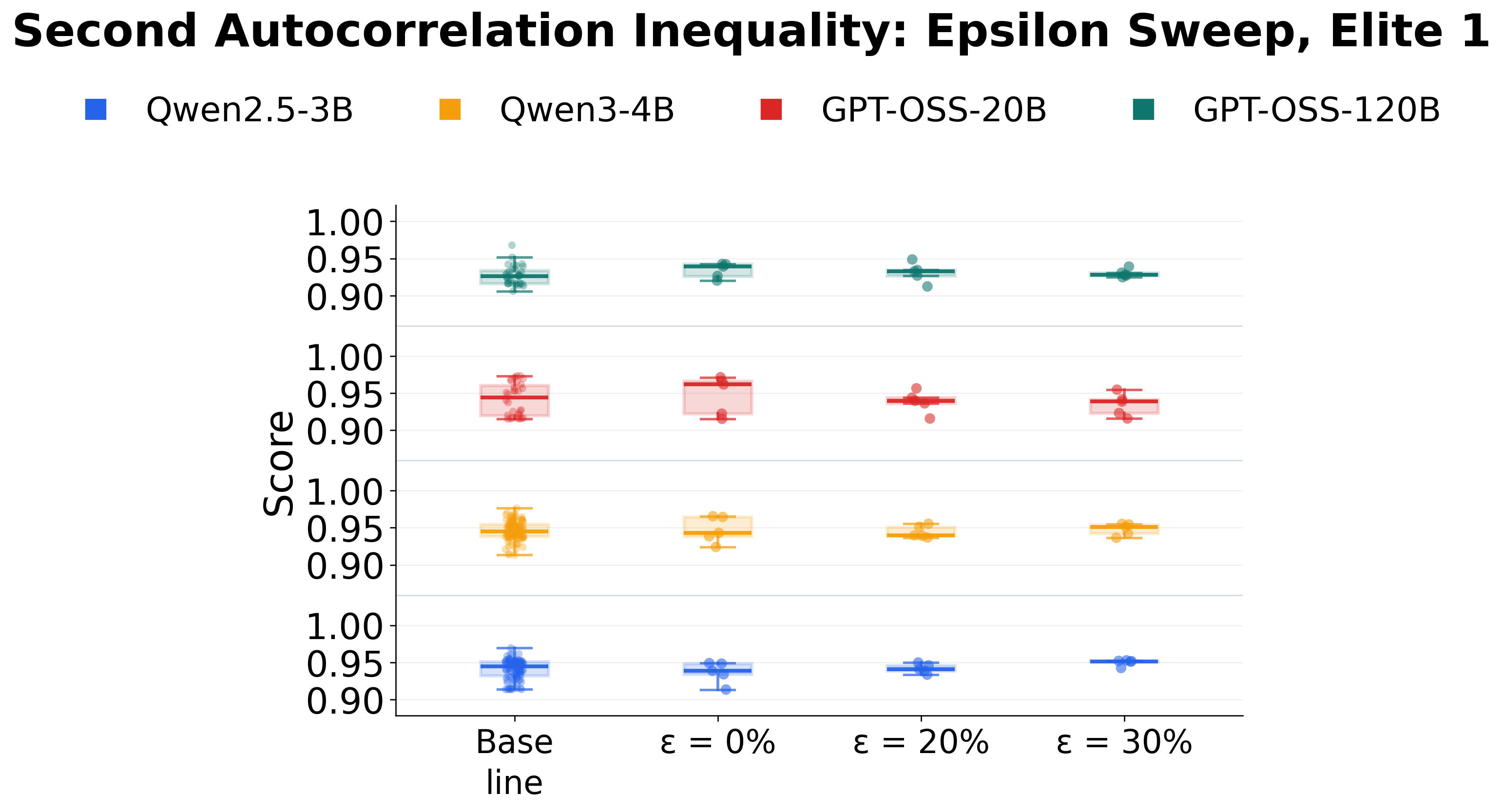}
    \caption{Second Autocorrelation Inequality.}
    \label{fig:second_autocorrelation_inequality_exploration}
  \end{subfigure}
  \caption{\textbf{General-history exploration sweep across all three tasks.} The elite archive is fixed at $K=1$, while the epsilon probability is increased from $0\%$ to $30\%$. Higher epsilon probability corresponds to more frequent sampling from the full program history.}
  \label{fig:exploration_sweep}
\end{figure*}

\paragraph{Experimental setup.}
We isolate exploration by fixing the elite archive size to $K=1$ and varying the $\epsilon$ probability over $\{0\%,20\%,30\%\}$. In the notation of Equation~\ref{eq:epsilon-greedy-parent-selection}, these settings correspond to exploration probabilities $\epsilon\in\{0,0.2,0.3\}$. The sweep tests whether occasional access to non-elite historical programs improves over always expanding the best program found so far.

Figures~\ref{fig:circle_packing_exploration}, ~\ref{fig:heilbronn_triangle_exploration}, and ~\ref{fig:second_autocorrelation_inequality_exploration} evaluate whether sampling from the general archive of all solutions improves over always exploiting the elite archive, and suggest that adding random access to the general archive is not automatically beneficial. The results do not show a single exploration rate that dominates across models or tasks. On circle packing, GPT-OSS-20B remains strong across the available exploration settings, while Qwen3-4B is competitive but more variable. Qwen2.5-3B exhibits the largest spread, suggesting that exploration can introduce instability when the model's proposal distribution is weaker, as we spend more compute budget experimenting with less performant solutions.

The Heilbronn triangle results are even more model-dependent. GPT-OSS-20B maintains high scores, but the Qwen models show mixed behavior and no clear monotonic trend as exploration increases. On the second autocorrelation inequality, the score distributions remain tightly clustered across $\epsilon$ probabilities. Qwen2.5-3B improves slightly at $\epsilon = 30\%$ , but this small shift does not transfer consistently to the other models. These results are consistent with viewing general-archive exploration as an anti-premature-convergence mechanism rather than a uniformly better parent-selection rule. Exploration may help recover diverse lineages, but it also spends budget on programs that may be less relevant than the current elite candidates.

\subsection{Full Epsilon-Greedy Grid}

\paragraph{Experimental setup.}
The full $\epsilon$-greedy sweep combines the previous interventions by varying both elite archive size and $\epsilon$. We evaluate $K \in \{1,10,20\}$ and $\epsilon$ probabilities in $\{0\%,20\%,30\%\}$ where runs are available. GPT-OSS-20B has incomplete coverage: the $10\%$ $\epsilon$ and $K=30$ settings are missing because of compute constraints. Empty positions therefore denote unavailable runs rather than zero performance.

Figures~\ref{fig:circle_packing_full_epsilon_greedy}, ~\ref{fig:heilbronn_triangle_full_epsilon_greedy}, and ~\ref{fig:second_autocorrelation_inequality_full_epsilon_greedy} show that these two hyperparameters interact rather than contributing independently. There is no globally dominant $(K,\epsilon)$ setting. On circle packing, Qwen3-4B often benefits from archive and exploration settings, while GPT-OSS-20B remains near ceiling for the configurations that were run. Qwen2.5-3B remains lower and more sensitive to hyperparameter choice. On Heilbronn triangle, the same sweep is less forgiving: GPT-OSS-20B stays strong, but the Qwen models show substantial variance and many configurations remain far below the GPT-OSS score range. On the second autocorrelation inequality, most configurations remain close to their model-specific baselines, and changing $K$ or the $\epsilon$ probability produces small, non-monotonic shifts rather than a clear winner. Thus, epsilon-greedy search should be treated as a family of policies whose effectiveness depends on the model's ability to produce useful candidate programs, rather than as a single robust improvement over Sequential BoN.

\subsection{Progression toward OpenEvolve}

\paragraph{Experimental setup.}
Finally, we compare archive and epsilon-greedy variants with a progression of increasingly OpenEvolve-like configurations. This sweep reallocates a fixed rollout budget from breadth to depth by decreasing the number of children per expansion, $N$, and increasing the number of update steps, $T$. The final settings add OpenEvolve-style island mechanisms. Because the available progression is model specific, we interpret these plots as within-model comparisons rather than a perfectly matched grid across models.

For Qwen2.5-3B and Qwen3-4B, the $(N,T)$ ladder is
$(N,T) \in \{(16,100),(8,200),(4,400),(2,800),(1,1600)\}$,
followed by OpenEvolve 1-island and multi-island variants. For GPT-OSS-20B, the available ladder is
$(N,T) \in \{(4,80),(2,160),(1,320)\}$,
followed by OpenEvolve 1-island and multi-island variants. Figure~\ref{fig:openevolve_progression_all_models} presents the complete model-specific progression.

The main trend is that deeper OpenEvolve-style refinement is competitive but does not uniformly dominate simpler search variants. On circle packing, GPT-OSS-20B remains strong across the progression, while Qwen3-4B achieves competitive scores in several intermediate settings. On Heilbronn triangle, GPT-OSS-20B again remains the most reliable, but the OpenEvolve-style settings do not consistently improve the Qwen models. The second autocorrelation inequality provides the clearest negative example. Intermediate breadth--depth settings remain near the baseline, while the one-island and four-island endpoints reduce performance for both Qwen models. This suggests that longer refinement chains are valuable when local improvements compound, but they can also overcommit the search to a narrow trajectory. In this sense, OpenEvolve-style search is best understood as one point on a breadth--depth tradeoff rather than as a uniformly superior endpoint.

\subsection{From Sequential BoN to UCT and PUCT}

The previous ablations treat parent selection as a relatively simple archive-sampling problem: we either sample from the elite archive, sample from the broader program database, or interpolate between the two. We next evaluate whether a more structured tree-search-inspired selection rule improves over these simpler Sequential BoN-style baselines. In particular, we compare PUCT and UCT variants that allocate the generation budget across multiple parent-selection branches.

For each task and model, we use the same Sequential BoN score distribution as the baseline. We evaluate UCT and PUCT under matched rollout budgets: 1600 generations for the Qwen models and 320 for GPT-OSS-20B. The sweeps vary three components: the exploration constant $C$, the number of children generated per selected parent, and the number of parent programs selected in parallel, $P$. The exact child batch size differs by model in the $P$ sweep because GPT-OSS-20B uses a smaller total budget. Those runs are therefore aligned by $P$ rather than by an identical batch size.

Colored boxes denote model-specific score distributions, and colored stars mark configurations whose five-run maximum significantly exceeds the Sequential BoN reference under the bootstrap test. Because each non-baseline configuration contains only five runs, we emphasize distributional trends and consistency across model--problem pairs rather than treating an individual setting as a definitive optimum.

\subsubsection{Exploration-Constant Ablations}
\begin{figure*}[t]
  \centering
  \begin{subfigure}[t]{0.49\textwidth}
    \centering
    \treesweepsubfiguregraphics{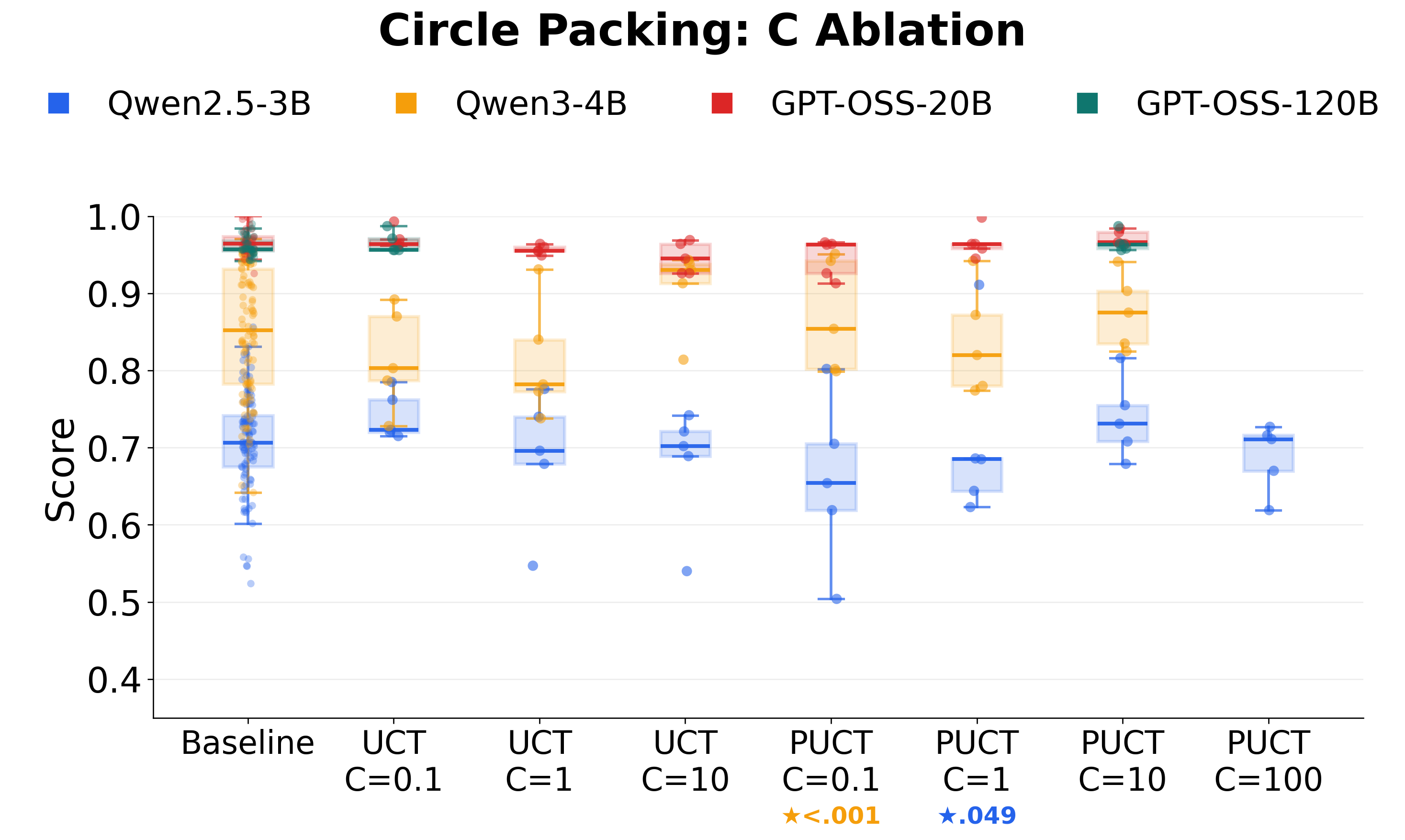}
    \caption{Circle packing.}
  \end{subfigure}
  \hfill
  \begin{subfigure}[t]{0.49\textwidth}
    \centering
    \treesweepsubfiguregraphics{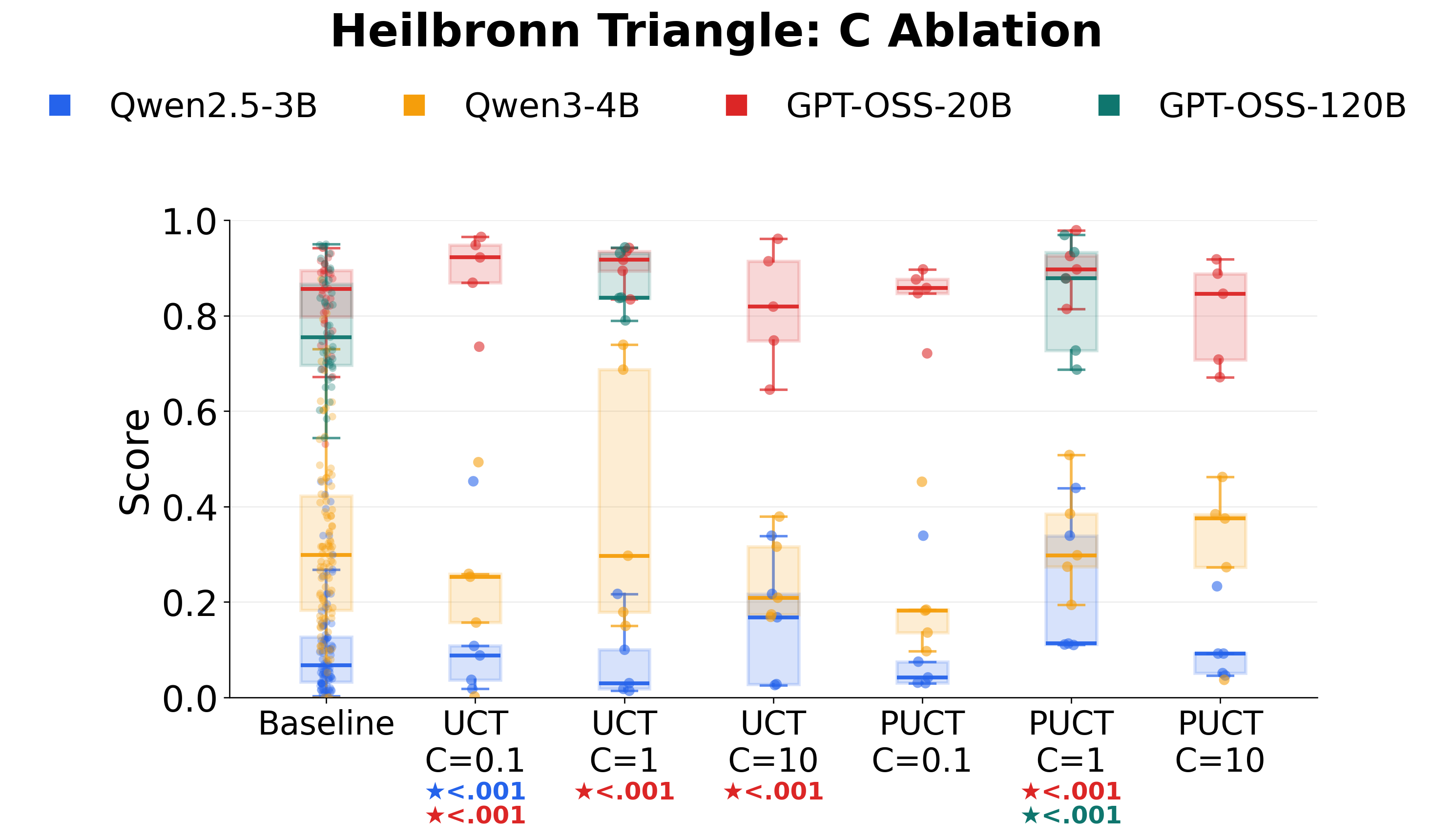}
    \caption{Heilbronn triangle.}
  \end{subfigure}
  \vspace{0.5em}

  \begin{subfigure}[t]{0.49\textwidth}
    \centering
    \treesweepsubfiguregraphics{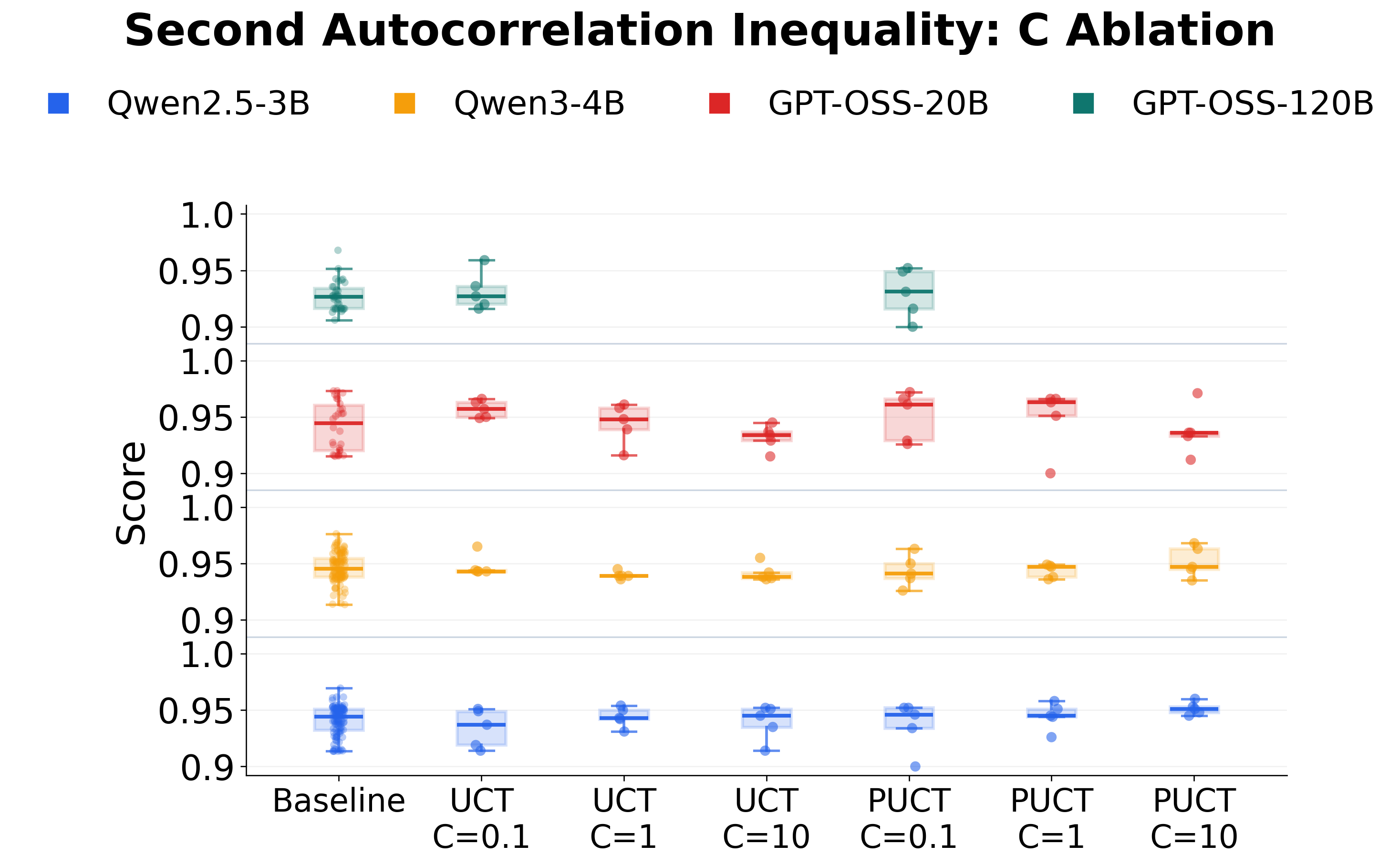}
    \caption{Second Autocorrelation Inequality.}
  \end{subfigure}
  \caption{\textbf{UCT/PUCT exploration-constant ablations across all three tasks.} PUCT and UCT are shown side by side for each value of $C$, with model-specific score distributions overlaid. The preferred exploration strength varies across model--problem pairs.}
  \label{fig:appendix_uct_puct_c_ablation}
\end{figure*}

\subsubsection{Breadth--Depth Ablations}

\begin{figure*}[t]
  \centering
  \begin{subfigure}[t]{0.49\textwidth}
    \centering
    \treesweepsubfiguregraphics{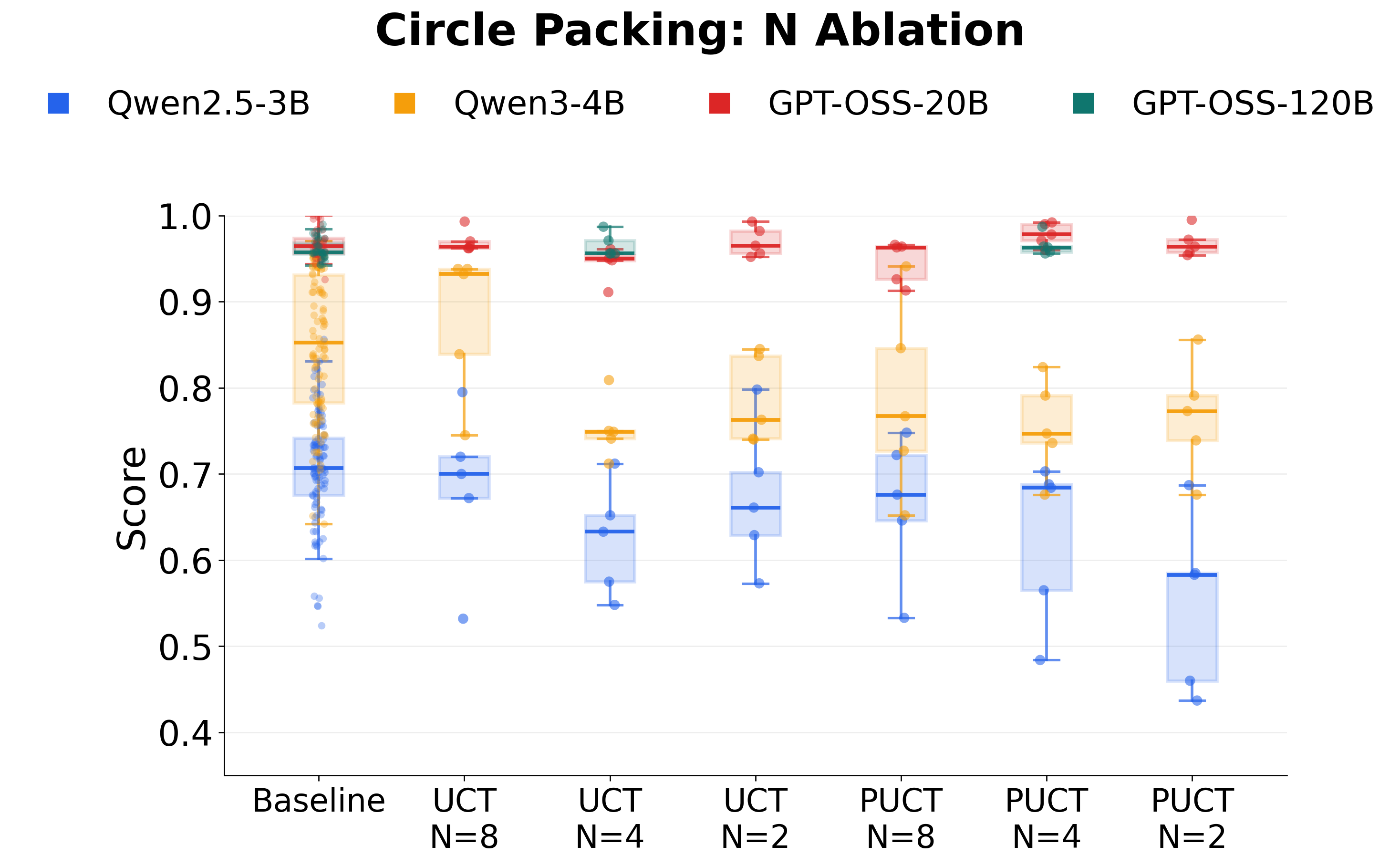}
    \caption{Circle Packing.}
    \label{fig:circle_packing_puct_b_ablation}
  \end{subfigure}
  \hfill
  \begin{subfigure}[t]{0.49\textwidth}
    \centering
    \treesweepsubfiguregraphics{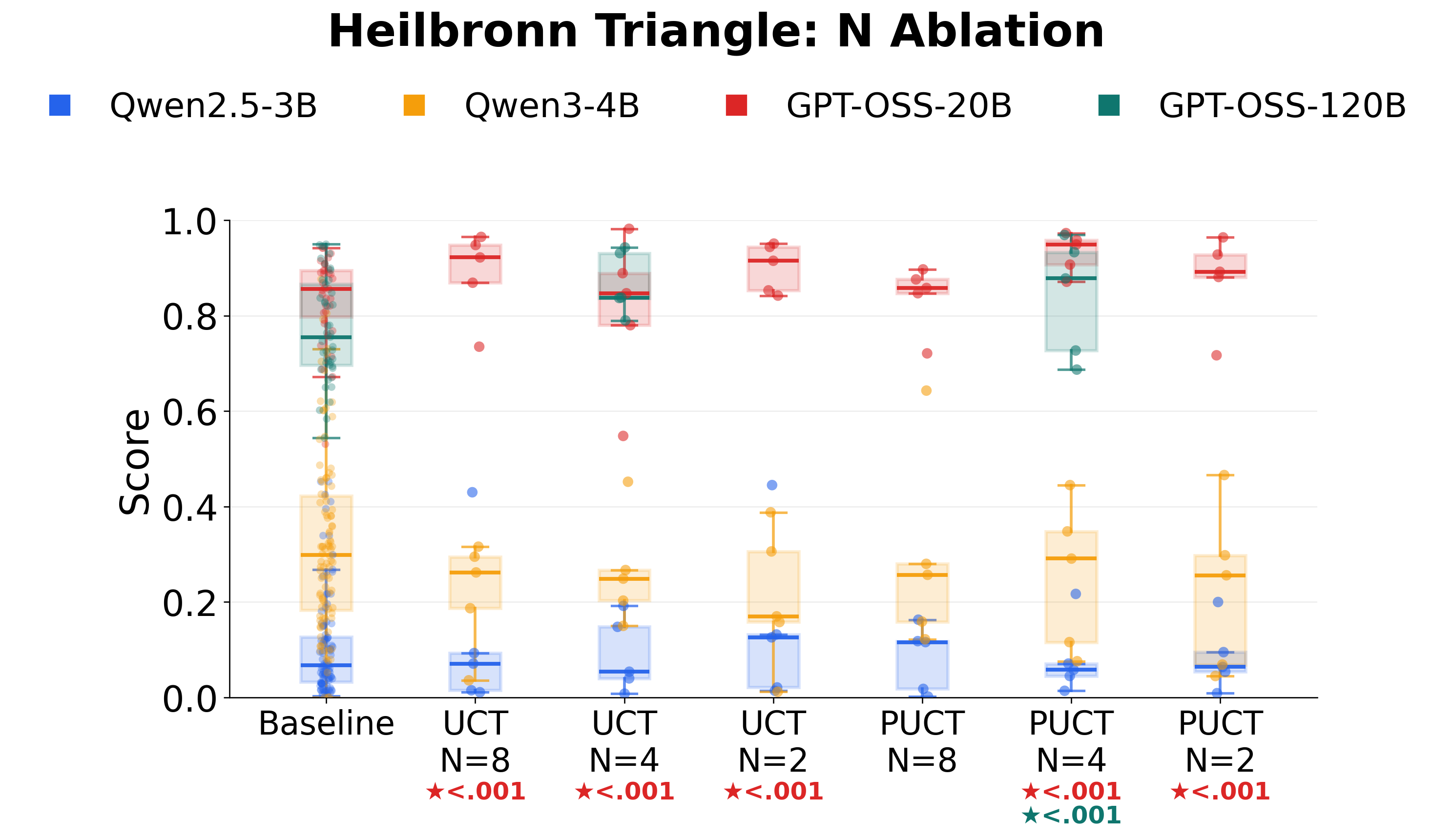}
    \caption{Heilbronn Triangle.}
    \label{fig:heilbronn_triangle_puct_b_ablation}
  \end{subfigure}
  \vspace{0.5em}

  \begin{subfigure}[t]{0.49\textwidth}
    \centering
    \treesweepsubfiguregraphics{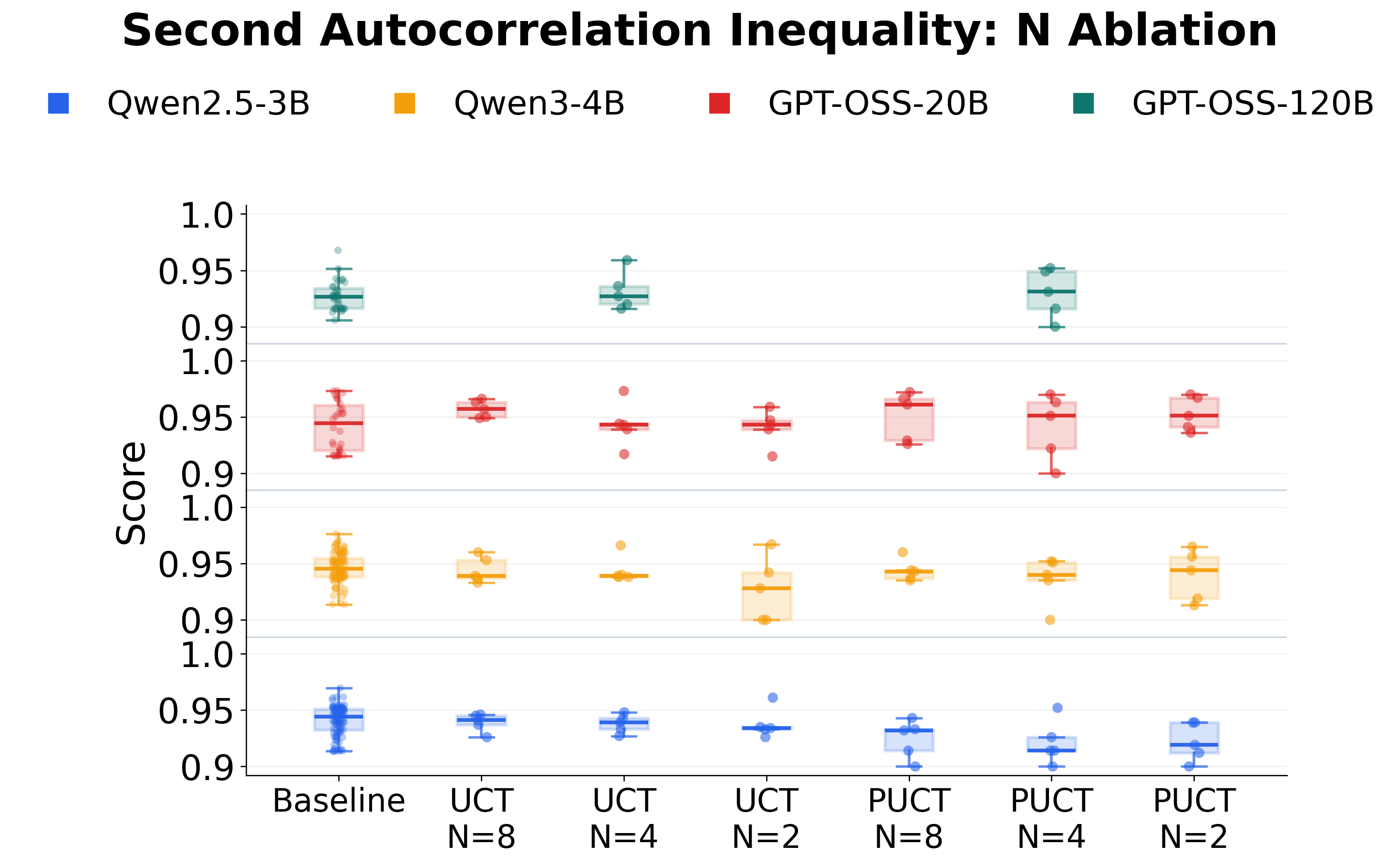}
    \caption{Second Autocorrelation Inequality.}
    \label{fig:second_autocorr_ineq_puct_b_ablation}
  \end{subfigure}
  \caption{\textbf{UCT/PUCT breadth--depth ablations.} The number of children generated per selected parent varies while the total rollout budget remains fixed. No single allocation is best across all model--problem pairs.}
  \label{fig:puct_b_ablation}
\end{figure*}

Figure~\ref{fig:puct_b_ablation} varies the child batch size $N \in \{8,4,2\}$ and adjusts the number of update steps $T$ so that the total rollout budget remains fixed. This tests whether the budget is better spent on broader branching or longer refinement chains. As in the $C$ sweep, no allocation is universally best. GPT-OSS-20B remains strong across the tested values on circle packing, while the Qwen models are more sensitive. In several settings, decreasing the child batch size does not improve the weaker models, suggesting that deeper refinement helps primarily when the selected lineage is already promising.

For Heilbronn triangle, changing the breadth--depth allocation has a smaller and less consistent effect. GPT-OSS-20B remains strong across the sweep, but the Qwen models do not improve reliably as $N$ changes. The second autocorrelation inequality also lacks a consistent breadth--depth preference across models, with most settings remaining close to their corresponding baselines. Search-budget allocation is therefore conditional on proposal quality: deeper search can amplify a promising lineage, but it can also overcommit to a weak trajectory.

\subsubsection{PUCT: P ablations}

\begin{figure*}[t]
  \centering
  \begin{subfigure}[t]{0.49\textwidth}
    \centering
    \treesweepsubfiguregraphics{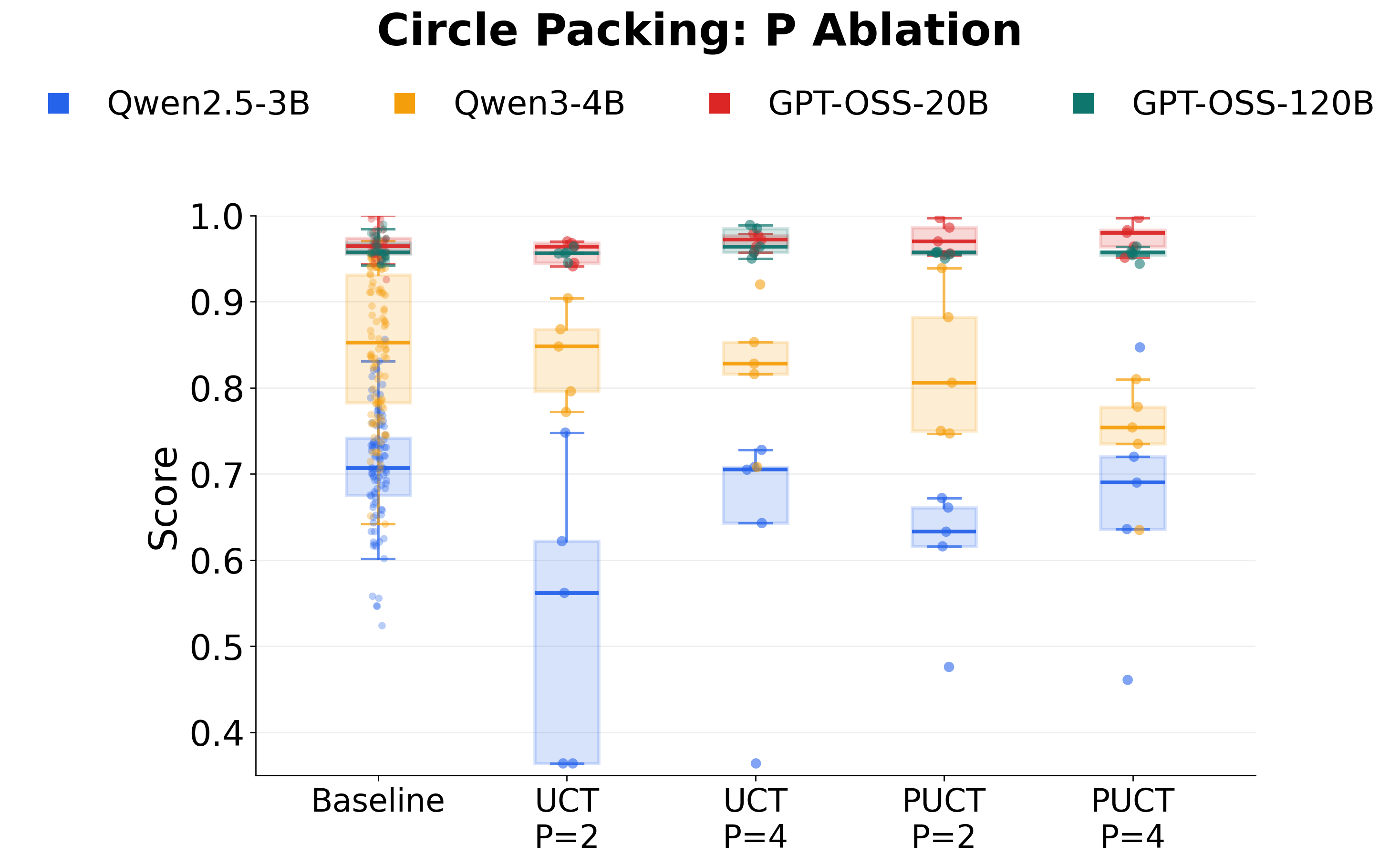}
    \caption{Circle Packing.}
    \label{fig:circle_packing_puct_p_ablation}
  \end{subfigure}
  \hfill
  \begin{subfigure}[t]{0.49\textwidth}
    \centering
    \treesweepsubfiguregraphics{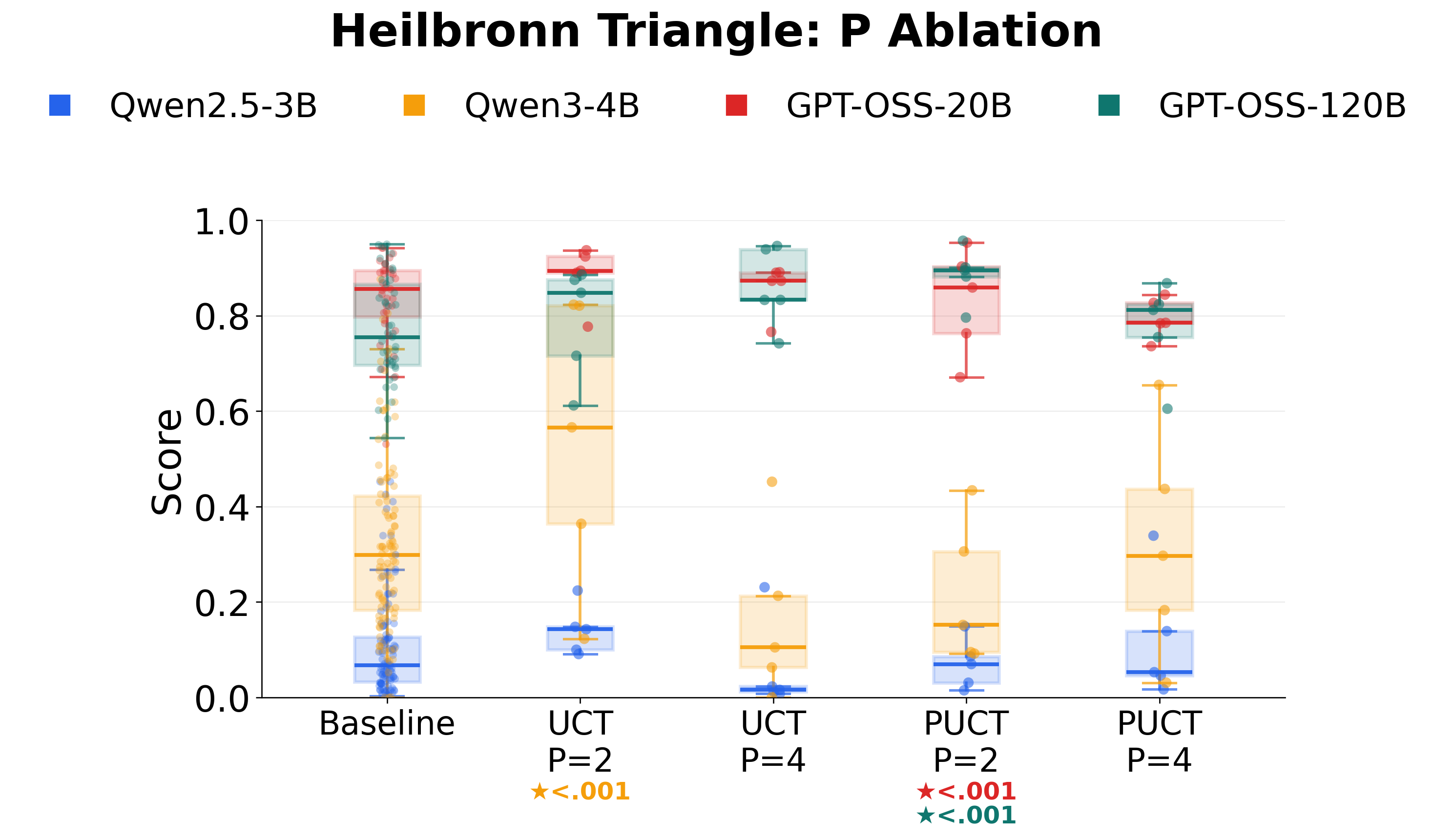}
    \caption{Heilbronn Triangle.}
    \label{fig:heilbronn_triangle_puct_p_ablation}
  \end{subfigure}
  \vspace{0.5em}

  \begin{subfigure}[t]{0.49\textwidth}
    \centering
    \treesweepsubfiguregraphics{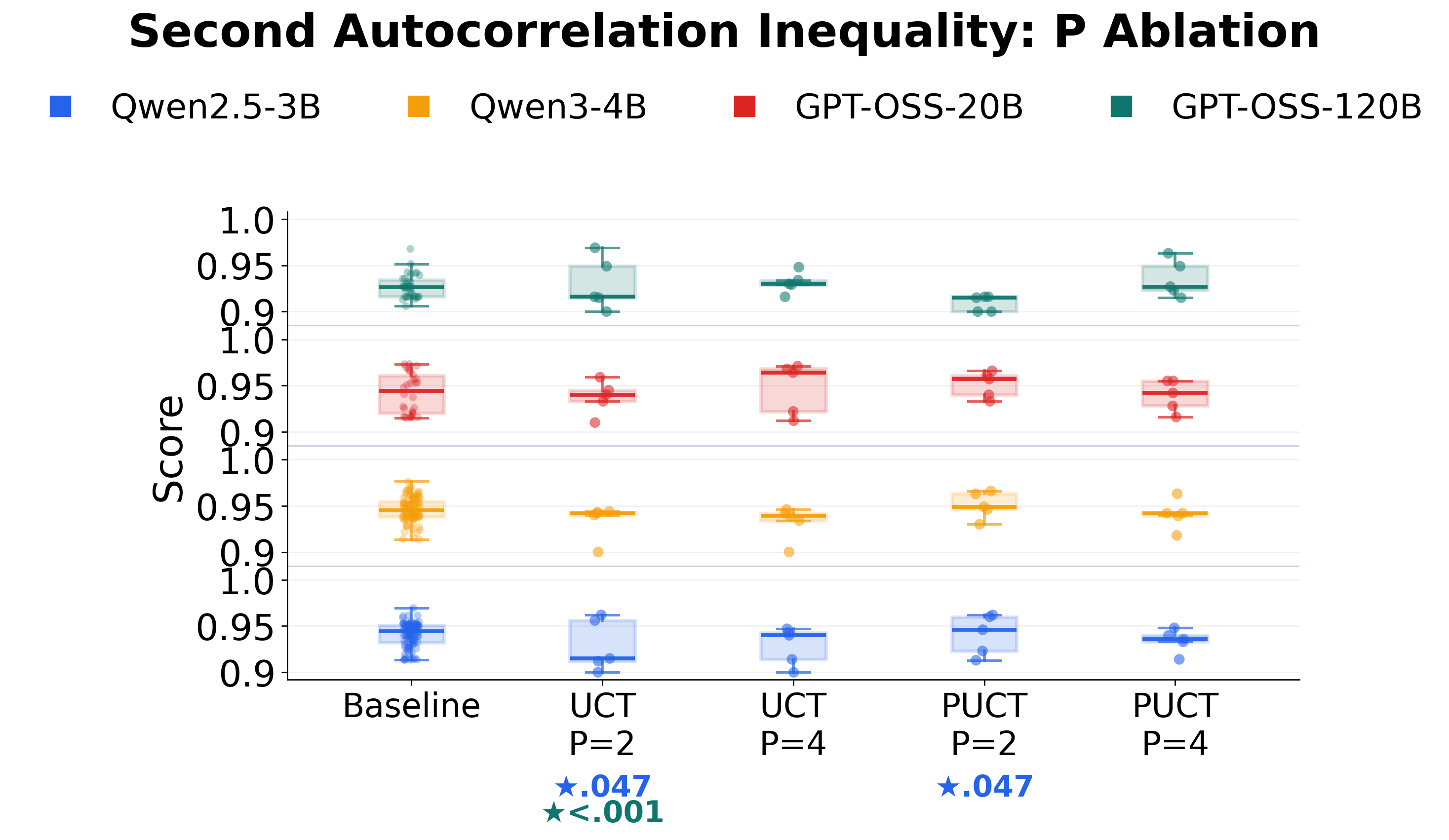}
    \caption{Second Autocorrelation Inequality.}
    \label{fig:second_autocorr_ineq_puct_p_ablation}
  \end{subfigure}
  \caption{\textbf{UCT/PUCT parallel-parent ablations.} Runs are grouped by the number of selected parent programs, $P$. Because total budgets differ by model, configurations are aligned by $P$ rather than by an identical child batch size.}
  \label{fig:puct_p_ablation}
\end{figure*}

Figure~\ref{fig:puct_p_ablation} varies the number of parent programs selected per step, $P \in \{2,4\}$. We keep the child batch size fixed at four for the Qwen models and use the corresponding budget-matched GPT-OSS-20B runs, adjusting the number of update steps to preserve the total rollout budget. Because the child batch sizes differ across model families, configurations are grouped by $P$. The figure should therefore be read as a comparison of parallel-parent search structure rather than a strict comparison of identical batch sizes.

Across the three tasks, increasing $P$ does not uniformly improve performance. On Circle Packing, GPT-OSS-20B remains near ceiling for both $P=2$ and $P=4$, while Qwen3-4B remains competitive but variable. Qwen2.5-3B again shows weaker and less stable results. On Heilbronn Triangle, the separation between models is especially large: GPT-OSS-20B remains high, whereas Qwen3-4B and Qwen2.5-3B show much lower distributions. On the second autocorrelation inequality, changing the number of root programs again produces only modest and model-dependent shifts. These results suggest that increasing the number of root programs is not by itself sufficient to improve search quality. The benefit depends on whether each root can seed a useful optimization trajectory.

Overall, the PUCT and UCT ablations reinforce the main pattern from the simpler archive-based experiments. Search structure matters, but its effect is conditional on the model's proposal quality. GPT-OSS-20B benefits from or at least remains robust under more structured search, while weaker models show greater sensitivity and less reliable improvement. In this setting, PUCT/UCT should be viewed not as a replacement for model capability, but as a mechanism for allocating search effort when the model can already generate promising candidates.

\section{Additional Config Vector Battle Leaderboards}\label{app:full_battle}
\paragraph{Construction of the progression plot.}
The progression plot in Figure~\ref{fig:progression_delta} is derived from the cross-pair majority-win probability leaderboard for the max statistic. For each configuration, we take the estimated probability that its best-of-five run beats the bootstrapped Sequential BoN baseline on a majority of the 12 model--problem pairs. Because the baseline has probability $0.5$ of beating itself, we first center each configuration by subtracting $0.5$. We then rescale the centered value by a factor of two:
\[
  s(c) = 2\left(\hat{p}_{\mathrm{maj}}(c) - 0.5\right).
\]
This maps the baseline to $0$, configurations that always beat the baseline to $+1$, and configurations that never beat the baseline to $-1$. The plotted points are ordered to match the experimental progressions in Figure~\ref{fig:search_progression}: the OpenEvolve panel moves from the Sequential BoN baseline through elite-archive and depth-budget ablations to OpenEvolve variants, while the TTT-Discover panel moves from the same baseline through tree-policy choices, depth-budget reductions, and the final $P=4$ TTT-Discover setting. Positive values therefore indicate improvement over the baseline majority-win rate, while negative values indicate degradation. The main trend is that simpler search variants, especially UCT/PUCT and small archive settings, remain above baseline, whereas adding the full OpenEvolve machinery drives performance substantially below baseline.

Figures~\ref{fig:config_vector_battle_max_full}, \ref{fig:config_vector_battle_median_full}, and~\ref{fig:config_vector_battle_mean_full} show the complete cross-pair leaderboard used to construct the compressed progression plot. Each bar reports how often a configuration beats the Sequential BoN baseline across a majority of model--problem pairs under the corresponding summary statistic. The ranking is consistent with the main-text message: lightweight tree-search and archive-based variants occupy most of the top positions, while the full OpenEvolve-style configurations appear near the bottom. 
\FloatBarrier
\input{figures/battle_figures_full}
\FloatBarrier

\clearpage
\onecolumn
\section{Complete Per-Configuration Results}
\label{app:complete_per_configuration_results}

Tables~\ref{tab:results_sequential_bon}--%
\ref{tab:results_puct_multi_parent} organize the results by search-harness category. Table~\ref{tab:results_sequential_bon} summarizes the full Sequential BoN baseline pool for each model--task pair, reporting the number of runs, mean, median, and maximum. The remaining tables report each five-run configuration's mean, median, and maximum beside their matched-baseline one-sided bootstrap p-values.

Every null distribution uses 100{,}000 bootstrap samples of five runs. Bold denotes $p<0.05$, while $<10^{-5}$ denotes that the configuration outperformed all bootstrap samples for the respective statistic. Unless otherwise noted, Qwen configurations use $N=16$ and $T=100$, while GPT-OSS-20B configurations use $N=8$ and $T=40$.

Pair codes are CP (circle packing), HT (Heilbronn triangle), and SAI (second autocorrelation inequality). Model codes are Q2.5 (Qwen2.5-3B-Instruct), Q3 (Qwen3-4B-Instruct-2507), and GPT (GPT-OSS-20B).

\begingroup
\scriptsize
\setlength{\tabcolsep}{6pt}
\renewcommand{\arraystretch}{1.08}
\begin{longtable}{@{}lrrrr@{}}
\caption{\textbf{Sequential BoN baselines.} Summary statistics for the complete matched baseline pools used by the bootstrap tests.}\label{tab:results_sequential_bon}\\
\toprule
Pair & Runs & Mean & Median & Max \\
\midrule
\endfirsthead
\multicolumn{5}{c}{\tablename\ \thetable\ (continued)}\\
\toprule
Pair & Runs & Mean & Median & Max \\
\midrule
\endhead
\midrule
\multicolumn{5}{r}{Continued on next page}\\
\endfoot
\bottomrule
\endlastfoot
CP/Q2.5 & 100 & 0.7087 & 0.7066 & 0.9488 \\
CP/Q3 & 100 & 0.8502 & 0.8524 & 0.9705 \\
CP/GPT20 & 30 & 0.9679 & 0.9658 & 0.9984 \\
CP/GPT-120B & 30 & 0.9625 & 0.9573 & 0.9898 \\
HT/Q2.5 & 100 & 0.1066 & 0.0672 & 0.4522 \\
HT/Q3 & 100 & 0.3213 & 0.2990 & 0.8765 \\
HT/GPT20 & 30 & 0.8300 & 0.8490 & 0.9417 \\
HT/GPT-120B & 30 & 0.7738 & 0.7507 & 0.9496 \\
SAI/Q2.5 & 100 & 0.9406 & 0.9444 & 0.9695 \\
SAI/Q3 & 100 & 0.9462 & 0.9452 & 0.9764 \\
SAI/GPT20 & 30 & 0.9417 & 0.9444 & 0.9732 \\
SAI/GPT-120B & 30 & 0.9268 & 0.9267 & 0.9515 \\
\end{longtable}
\endgroup

\begingroup
\scriptsize
\setlength{\tabcolsep}{2pt}
\renewcommand{\arraystretch}{1.08}
\begin{longtable}{@{}lp{0.50\textwidth}rrrrrr@{}}
\caption{\textbf{Sequential BoN + top-$K$.} Adds a top-$K$ elite archive while sampling exclusively from that archive.}\label{tab:results_top_k}\\
\toprule
\multicolumn{2}{c}{} & \multicolumn{3}{c}{Observed statistic} & \multicolumn{3}{c}{One-sided p-value}\\
\cmidrule(lr){3-5}\cmidrule(l){6-8}
Pair & Configuration & Mean & Median & Max & $p_{\mathrm{mean}}$ & $p_{\mathrm{median}}$ & $p_{\mathrm{max}}$ \\
\midrule
\endfirsthead
\multicolumn{8}{c}{\tablename\ \thetable\ (continued)}\\
\toprule
\multicolumn{2}{c}{} & \multicolumn{3}{c}{Observed statistic} & \multicolumn{3}{c}{One-sided p-value}\\
\cmidrule(lr){3-5}\cmidrule(l){6-8}
Pair & Configuration & Mean & Median & Max & $p_{\mathrm{mean}}$ & $p_{\mathrm{median}}$ & $p_{\mathrm{max}}$ \\
\midrule
\endhead
\midrule
\multicolumn{8}{r}{Continued on next page}\\
\endfoot
\bottomrule
\endlastfoot
CP/Q2.5 & K=10 & 0.7252 & 0.7004 & 0.7974 & 0.28825 & 0.68088 & 0.34064 \\
CP/Q2.5 & K=20 & 0.6840 & 0.6997 & 0.7303 & 0.79711 & 0.69830 & 0.90788 \\
CP/Q2.5 & K=30 & 0.6975 & 0.6896 & 0.7707 & 0.65234 & 0.80736 & 0.55686 \\
CP/Q3 & K=10 & 0.8628 & 0.9032 & 0.9387 & 0.37690 & 0.25001 & 0.70924 \\
CP/Q3 & K=20 & 0.8898 & 0.9387 & 0.9525 & 0.14044 & 0.07406 & 0.33831 \\
CP/Q3 & K=30 & 0.9101 & 0.9124 & 0.9428 & \textbf{0.04408} & 0.16242 & 0.52763 \\
CP/GPT20 & K=10 & 0.9581 & 0.9589 & 0.9774 & 0.92150 & 0.87938 & 0.73659 \\
CP/GPT20 & K=20 & 0.9579 & 0.9640 & 0.9716 & 0.92569 & 0.73934 & 0.86832 \\
CP/GPT-120B & K=10 & 0.9603 & 0.9581 & 0.9828 & 0.63900 & 0.25300 & 0.28200 \\
CP/GPT-120B & K=20 & 0.9492 & 0.9541 & 0.9556 & 0.99700 & 0.96500 & 1.00000 \\
HT/Q2.5 & K=10 & 0.1598 & 0.1220 & 0.3570 & 0.13651 & 0.14980 & 0.22633 \\
HT/Q2.5 & K=20 & 0.0802 & 0.0820 & 0.1360 & 0.66310 & 0.40512 & 0.74757 \\
HT/Q2.5 & K=30 & 0.1622 & 0.0990 & 0.4000 & 0.12800 & 0.31593 & 0.18496 \\
HT/Q3 & K=10 & 0.2800 & 0.3110 & 0.3840 & 0.66657 & 0.44289 & 0.83360 \\
HT/Q3 & K=20 & 0.2800 & 0.3110 & 0.3840 & 0.66657 & 0.44289 & 0.83360 \\
HT/Q3 & K=30 & 0.2734 & 0.2840 & 0.4080 & 0.69598 & 0.55453 & 0.80732 \\
HT/GPT20 & K=10 & 0.8722 & 0.8871 & 0.9090 & 0.14883 & 0.26081 & 0.51170 \\
HT/GPT20 & K=20 & 0.8493 & 0.8618 & 0.8734 & 0.35021 & 0.43606 & 0.94127 \\
HT/GPT-120B & K=10 & 0.8067 & 0.7858 & 0.9540 & 0.24600 & 0.30700 & \textbf{$<10^{-5}$} \\
HT/GPT-120B & K=20 & 0.7623 & 0.7621 & 0.8929 & 0.59900 & 0.43700 & 0.66800 \\
SAI/Q2.5 & K=10 & 0.9425 & 0.9385 & 0.9515 & 0.39766 & 0.74919 & 0.65380 \\
SAI/Q2.5 & K=20 & 0.9405 & 0.9402 & 0.9514 & 0.52628 & 0.64846 & 0.65380 \\
SAI/Q2.5 & K=30 & 0.9476 & 0.9510 & 0.9521 & 0.11450 & 0.10397 & 0.58486 \\
SAI/Q3 & K=10 & 0.9523 & 0.9536 & 0.9684 & 0.14249 & 0.13692 & 0.14021 \\
SAI/Q3 & K=20 & 0.9467 & 0.9468 & 0.9566 & 0.47437 & 0.44256 & 0.74515 \\
SAI/Q3 & K=30 & 0.9407 & 0.9385 & 0.9548 & 0.82718 & 0.86311 & 0.76110 \\
SAI/GPT20 & K=10 & 0.9388 & 0.9298 & 0.9638 & 0.61273 & 0.62422 & 0.73484 \\
SAI/GPT20 & K=20 & 0.9457 & 0.9496 & 0.9717 & 0.34475 & 0.43836 & 0.29137 \\
SAI/GPT20 & K=30 & 0.9292 & 0.9374 & 0.9588 & 0.89866 & 0.56195 & 0.78812 \\
SAI/GPT-120B & K=10 & 0.9243 & 0.9246 & 0.9335 & 0.68400 & 0.67800 & 0.78500 \\
SAI/GPT-120B & K=20 & 0.9214 & 0.9310 & 0.9395 & 0.85600 & 0.23600 & 0.60400 \\
\end{longtable}
\endgroup

\begingroup
\scriptsize
\setlength{\tabcolsep}{2pt}
\renewcommand{\arraystretch}{1.08}
\begin{longtable}{@{}lp{0.50\textwidth}rrrrrr@{}}
\caption{\textbf{Sequential BoN + full history.} Adds full-history exploration while retaining a single elite program.}\label{tab:results_full_history}\\
\toprule
\multicolumn{2}{c}{} & \multicolumn{3}{c}{Observed statistic} & \multicolumn{3}{c}{One-sided p-value}\\
\cmidrule(lr){3-5}\cmidrule(l){6-8}
Pair & Configuration & Mean & Median & Max & $p_{\mathrm{mean}}$ & $p_{\mathrm{median}}$ & $p_{\mathrm{max}}$ \\
\midrule
\endfirsthead
\multicolumn{8}{c}{\tablename\ \thetable\ (continued)}\\
\toprule
\multicolumn{2}{c}{} & \multicolumn{3}{c}{Observed statistic} & \multicolumn{3}{c}{One-sided p-value}\\
\cmidrule(lr){3-5}\cmidrule(l){6-8}
Pair & Configuration & Mean & Median & Max & $p_{\mathrm{mean}}$ & $p_{\mathrm{median}}$ & $p_{\mathrm{max}}$ \\
\midrule
\endhead
\midrule
\multicolumn{8}{r}{Continued on next page}\\
\endfoot
\bottomrule
\endlastfoot
CP/Q2.5 & epsilon=10\% & 0.7130 & 0.6842 & 0.7685 & 0.44745 & 0.83474 & 0.58263 \\
CP/Q2.5 & epsilon=100\% & 0.6210 & 0.6220 & 0.6646 & 0.99709 & 0.98858 & 0.99961 \\
CP/Q2.5 & epsilon=20\% & 0.7116 & 0.7394 & 0.8273 & 0.46639 & 0.12700 & 0.14229 \\
CP/Q2.5 & epsilon=30\% & 0.7230 & 0.7592 & 0.8727 & 0.31476 & 0.05864 & \textbf{0.04919} \\
CP/Q2.5 & epsilon=30\%, N=2 & 0.7099 & 0.7261 & 0.7398 & 0.48928 & 0.31851 & 0.77817 \\
CP/Q2.5 & epsilon=30\%, N=4 & 0.7473 & 0.7304 & 0.9108 & 0.09636 & 0.28469 & \textbf{0.04919} \\
CP/Q2.5 & epsilon=30\%, N=8 & 0.6865 & 0.6893 & 0.7207 & 0.77441 & 0.80736 & 0.93858 \\
CP/Q3 & epsilon=10\% & 0.8555 & 0.8708 & 0.9376 & 0.45281 & 0.42621 & 0.74454 \\
CP/Q3 & epsilon=100\% & 0.7723 & 0.7818 & 0.8719 & 0.98124 & 0.90733 & 0.94882 \\
CP/Q3 & epsilon=20\% & 0.8853 & 0.8843 & 0.9526 & 0.17137 & 0.30004 & 0.13977 \\
CP/Q3 & epsilon=30\% & 0.8163 & 0.8327 & 0.9144 & 0.82201 & 0.71718 & 0.81911 \\
CP/GPT20 & epsilon=100\% & 0.9363 & 0.9377 & 0.9740 & 0.99999 & 0.99969 & 0.78871 \\
CP/GPT20 & epsilon=20\% & 0.9666 & 0.9638 & 0.9919 & 0.57688 & 0.79216 & 0.41107 \\
CP/GPT20 & epsilon=30\% & 0.9615 & 0.9644 & 0.9861 & 0.82669 & 0.68277 & 0.51342 \\
CP/GPT20 & epsilon=30\%, N=2 & 0.9700 & 0.9710 & 0.9803 & 0.38067 & 0.20907 & 0.67449 \\
CP/GPT20 & epsilon=30\%, N=4 & 0.9747 & 0.9722 & 0.9944 & 0.16410 & 0.16236 & 0.41107 \\
CP/GPT-120B & epsilon=100\% & 0.9395 & 0.9377 & 0.9545 & 1.00000 & 1.00000 & 1.00000 \\
CP/GPT-120B & epsilon=20\% & 0.9621 & 0.9645 & 0.9883 & 0.51700 & 0.20200 & 0.15200 \\
CP/GPT-120B & epsilon=30\% & 0.9577 & 0.9555 & 0.9650 & 0.78700 & 0.91100 & 0.85800 \\
HT/Q2.5 & epsilon=10\% & 0.0684 & 0.0480 & 0.1460 & 0.76706 & 0.74734 & 0.74757 \\
HT/Q2.5 & epsilon=100\% & 0.1136 & 0.1280 & 0.2170 & 0.39538 & 0.10327 & 0.55556 \\
HT/Q2.5 & epsilon=20\% & 0.2022 & 0.0930 & 0.4520 & \textbf{0.03948} & 0.38619 & 0.09626 \\
HT/Q2.5 & epsilon=30\% & 0.1828 & 0.0690 & 0.4520 & 0.07190 & 0.47918 & 0.09626 \\
HT/Q3 & epsilon=10\% & 0.2806 & 0.2280 & 0.6390 & 0.66406 & 0.74674 & 0.30433 \\
HT/Q3 & epsilon=100\% & 0.1110 & 0.0980 & 0.1730 & 0.99868 & 0.99561 & 0.99923 \\
HT/Q3 & epsilon=20\% & 0.3594 & 0.3030 & 0.6460 & 0.31180 & 0.48019 & 0.30433 \\
HT/Q3 & epsilon=30\% & 0.1934 & 0.1710 & 0.2710 & 0.94759 & 0.90518 & 0.98706 \\
HT/GPT20 & epsilon=100\% & 0.7760 & 0.7829 & 0.8709 & 0.89569 & 0.91397 & 0.94127 \\
HT/GPT20 & epsilon=20\% & 0.9215 & 0.9347 & 0.9590 & \textbf{0.00088} & \textbf{0.00258} & \textbf{$<10^{-5}$} \\
HT/GPT20 & epsilon=30\% & 0.8597 & 0.8804 & 0.9608 & 0.25169 & 0.31523 & \textbf{$<10^{-5}$} \\
HT/GPT-120B & epsilon=100\% & 0.7001 & 0.7409 & 0.8006 & 0.93700 & 0.56100 & 0.92100 \\
HT/GPT-120B & epsilon=20\% & 0.8333 & 0.8940 & 0.9738 & 0.09800 & \textbf{0.04800} & \textbf{$<10^{-5}$} \\
HT/GPT-120B & epsilon=30\% & 0.8025 & 0.8667 & 0.9570 & 0.27100 & 0.10500 & \textbf{$<10^{-5}$} \\
SAI/Q2.5 & epsilon=10\% & 0.9385 & 0.9361 & 0.9513 & 0.65448 & 0.83689 & 0.71473 \\
SAI/Q2.5 & epsilon=100\% & 0.9369 & 0.9394 & 0.9445 & 0.74589 & 0.71651 & 0.96594 \\
SAI/Q2.5 & epsilon=20\% & 0.9417 & 0.9408 & 0.9498 & 0.44758 & 0.59412 & 0.84587 \\
SAI/Q2.5 & epsilon=30\% & 0.9500 & 0.9515 & 0.9527 & \textbf{0.04536} & 0.05082 & 0.44333 \\
SAI/Q3 & epsilon=10\% & 0.9460 & 0.9484 & 0.9696 & 0.52224 & 0.40567 & 0.09589 \\
SAI/Q3 & epsilon=100\% & 0.9340 & 0.9360 & 0.9365 & 0.97919 & 0.96258 & 0.99978 \\
SAI/Q3 & epsilon=20\% & 0.9444 & 0.9397 & 0.9553 & 0.62436 & 0.83745 & 0.76110 \\
SAI/Q3 & epsilon=30\% & 0.9479 & 0.9510 & 0.9551 & 0.39146 & 0.33360 & 0.76110 \\
SAI/GPT20 & epsilon=10\% & 0.9341 & 0.9377 & 0.9528 & 0.78303 & 0.56195 & 0.92358 \\
SAI/GPT20 & epsilon=100\% & 0.9283 & 0.9277 & 0.9564 & 0.91452 & 0.62422 & 0.83268 \\
SAI/GPT20 & epsilon=20\% & 0.9384 & 0.9396 & 0.9564 & 0.62869 & 0.56195 & 0.83268 \\
SAI/GPT20 & epsilon=30\% & 0.9347 & 0.9386 & 0.9546 & 0.76217 & 0.56195 & 0.86877 \\
SAI/GPT-120B & epsilon=100\% & 0.9218 & 0.9154 & 0.9355 & 0.83700 & 0.97800 & 0.78500 \\
SAI/GPT-120B & epsilon=20\% & 0.9310 & 0.9328 & 0.9486 & 0.20800 & 0.13400 & 0.17000 \\
SAI/GPT-120B & epsilon=30\% & 0.9299 & 0.9281 & 0.9390 & 0.28000 & 0.35200 & 0.67500 \\
\end{longtable}
\endgroup

\begingroup
\scriptsize
\setlength{\tabcolsep}{2pt}
\renewcommand{\arraystretch}{1.08}
\begin{longtable}{@{}lp{0.50\textwidth}rrrrrr@{}}
\caption{\textbf{Epsilon-greedy (top-$K$ + full history).} Combines a top-$K$ elite archive with epsilon-greedy sampling from the full history.}\label{tab:results_epsilon_greedy}\\
\toprule
\multicolumn{2}{c}{} & \multicolumn{3}{c}{Observed statistic} & \multicolumn{3}{c}{One-sided p-value}\\
\cmidrule(lr){3-5}\cmidrule(l){6-8}
Pair & Configuration & Mean & Median & Max & $p_{\mathrm{mean}}$ & $p_{\mathrm{median}}$ & $p_{\mathrm{max}}$ \\
\midrule
\endfirsthead
\multicolumn{8}{c}{\tablename\ \thetable\ (continued)}\\
\toprule
\multicolumn{2}{c}{} & \multicolumn{3}{c}{Observed statistic} & \multicolumn{3}{c}{One-sided p-value}\\
\cmidrule(lr){3-5}\cmidrule(l){6-8}
Pair & Configuration & Mean & Median & Max & $p_{\mathrm{mean}}$ & $p_{\mathrm{median}}$ & $p_{\mathrm{max}}$ \\
\midrule
\endhead
\midrule
\multicolumn{8}{r}{Continued on next page}\\
\endfoot
\bottomrule
\endlastfoot
CP/Q2.5 & K=10, epsilon=10\% & 0.7538 & 0.7339 & 0.8664 & 0.06534 & 0.17797 & \textbf{0.04919} \\
CP/Q2.5 & K=10, epsilon=10\%, N=2 & 0.6922 & 0.6830 & 0.7308 & 0.71403 & 0.84749 & 0.90013 \\
CP/Q2.5 & K=10, epsilon=10\%, N=4 & 0.6249 & 0.6448 & 0.6892 & 0.99576 & 0.97289 & 0.99655 \\
CP/Q2.5 & K=10, epsilon=10\%, N=8 & 0.7269 & 0.7223 & 0.8057 & 0.26981 & 0.31851 & 0.26652 \\
CP/Q2.5 & K=10, epsilon=20\% & 0.6669 & 0.6822 & 0.7129 & 0.91558 & 0.85919 & 0.95786 \\
CP/Q2.5 & K=10, epsilon=30\% & 0.7343 & 0.7318 & 0.7812 & 0.19188 & 0.23640 & 0.44275 \\
CP/Q2.5 & K=20, epsilon=10\% & 0.6823 & 0.6842 & 0.7016 & 0.81295 & 0.83474 & 0.98853 \\
CP/Q2.5 & K=20, epsilon=20\% & 0.6517 & 0.6858 & 0.7163 & 0.96690 & 0.83474 & 0.95382 \\
CP/Q2.5 & K=20, epsilon=30\% & 0.6805 & 0.6712 & 0.7463 & 0.82760 & 0.91465 & 0.74639 \\
CP/Q2.5 & K=30, epsilon=10\% & 0.7624 & 0.7313 & 0.9398 & \textbf{0.03862} & 0.23640 & \textbf{0.04919} \\
CP/Q2.5 & K=30, epsilon=10\%, N=2 & 0.6068 & 0.6167 & 0.6395 & 0.99938 & 0.99561 & 0.99992 \\
CP/Q2.5 & K=30, epsilon=10\%, N=4 & 0.7137 & 0.7185 & 0.8073 & 0.43800 & 0.40798 & 0.26652 \\
CP/Q2.5 & K=30, epsilon=10\%, N=8 & 0.6891 & 0.6842 & 0.7065 & 0.74799 & 0.83474 & 0.96879 \\
CP/Q2.5 & K=30, epsilon=20\% & 0.6764 & 0.6853 & 0.7810 & 0.85846 & 0.83474 & 0.44275 \\
CP/Q2.5 & K=30, epsilon=30\% & 0.6915 & 0.6723 & 0.7362 & 0.72300 & 0.91465 & 0.80599 \\
CP/Q3 & K=10, epsilon=10\% & 0.8937 & 0.8849 & 0.9488 & 0.11659 & 0.30004 & 0.47015 \\
CP/Q3 & K=10, epsilon=10\%, N=2 & 0.8178 & 0.8127 & 0.8665 & 0.81215 & 0.79433 & 0.95327 \\
CP/Q3 & K=10, epsilon=10\%, N=4 & 0.8343 & 0.8256 & 0.9227 & 0.67360 & 0.73264 & 0.77671 \\
CP/Q3 & K=10, epsilon=10\%, N=8 & 0.8277 & 0.8452 & 0.8709 & 0.73327 & 0.53908 & 0.95327 \\
CP/Q3 & K=10, epsilon=20\% & 0.8435 & 0.8644 & 0.8684 & 0.58162 & 0.44469 & 0.95327 \\
CP/Q3 & K=10, epsilon=30\% & 0.8349 & 0.8417 & 0.9023 & 0.66782 & 0.59354 & 0.89254 \\
CP/Q3 & K=20, epsilon=10\% & 0.9047 & 0.9108 & 0.9324 & 0.06266 & 0.21906 & 0.74454 \\
CP/Q3 & K=20, epsilon=20\% & 0.8554 & 0.8306 & 0.9387 & 0.45456 & 0.71718 & 0.70924 \\
CP/Q3 & K=20, epsilon=30\% & 0.8946 & 0.8728 & 0.9376 & 0.11157 & 0.40724 & 0.74454 \\
CP/Q3 & K=30, epsilon=10\% & 0.8914 & 0.8880 & 0.9376 & 0.12996 & 0.30004 & 0.74454 \\
CP/Q3 & K=30, epsilon=20\% & 0.8285 & 0.8531 & 0.9191 & 0.72698 & 0.50124 & 0.79139 \\
CP/Q3 & K=30, epsilon=30\% & 0.8897 & 0.8956 & 0.9525 & 0.14132 & 0.25001 & 0.33831 \\
CP/GPT20 & K=10, epsilon=20\% & 0.9569 & 0.9547 & 0.9645 & 0.94240 & 0.98057 & 0.96825 \\
CP/GPT20 & K=10, epsilon=30\% & 0.9472 & 0.9543 & 0.9563 & 0.99767 & 0.99127 & 0.99924 \\
CP/GPT20 & K=20, epsilon=20\% & 0.9517 & 0.9526 & 0.9603 & 0.98826 & 0.99127 & 0.99868 \\
CP/GPT20 & K=20, epsilon=30\% & 0.9575 & 0.9566 & 0.9701 & 0.93306 & 0.91425 & 0.89798 \\
CP/GPT-120B & K=10, epsilon=20\% & 0.9581 & 0.9537 & 0.9765 & 0.77100 & 0.96500 & 0.59900 \\
CP/GPT-120B & K=10, epsilon=30\% & 0.9596 & 0.9602 & 0.9650 & 0.68800 & 0.25300 & 0.85800 \\
CP/GPT-120B & K=20, epsilon=20\% & 0.9482 & 0.9525 & 0.9558 & 0.99900 & 0.98000 & 1.00000 \\
CP/GPT-120B & K=20, epsilon=30\% & 0.9539 & 0.9517 & 0.9725 & 0.94900 & 0.99200 & 0.77900 \\
HT/Q2.5 & K=10, epsilon=10\% & 0.0646 & 0.0800 & 0.1010 & 0.79999 & 0.40512 & 0.92260 \\
HT/Q2.5 & K=10, epsilon=20\% & 0.1114 & 0.0870 & 0.2170 & 0.41120 & 0.40512 & 0.55556 \\
HT/Q2.5 & K=10, epsilon=30\% & 0.1268 & 0.0910 & 0.2550 & 0.30544 & 0.38619 & 0.44074 \\
HT/Q2.5 & K=20, epsilon=10\% & 0.1452 & 0.0760 & 0.4200 & 0.19958 & 0.42335 & 0.14172 \\
HT/Q2.5 & K=20, epsilon=20\% & 0.2366 & 0.2170 & 0.4250 & \textbf{0.01109} & \textbf{0.02685} & 0.14172 \\
HT/Q2.5 & K=20, epsilon=20\%, N=2 & 0.1948 & 0.1500 & 0.3390 & \textbf{0.04961} & 0.08385 & 0.30511 \\
HT/Q2.5 & K=20, epsilon=20\%, N=4 & 0.2174 & 0.1670 & 0.4240 & \textbf{0.02292} & \textbf{0.04405} & 0.14172 \\
HT/Q2.5 & K=20, epsilon=20\%, N=8 & 0.1562 & 0.1060 & 0.4070 & 0.15006 & 0.26568 & 0.18496 \\
HT/Q2.5 & K=20, epsilon=30\% & 0.1078 & 0.0870 & 0.2170 & 0.43784 & 0.40512 & 0.55556 \\
HT/Q2.5 & K=30, epsilon=10\% & 0.1222 & 0.0990 & 0.3390 & 0.33535 & 0.31593 & 0.30511 \\
HT/Q2.5 & K=30, epsilon=20\% & 0.1248 & 0.0740 & 0.3390 & 0.31798 & 0.44176 & 0.30511 \\
HT/Q2.5 & K=30, epsilon=30\% & 0.1126 & 0.0940 & 0.2170 & 0.40253 & 0.36810 & 0.55556 \\
HT/Q3 & K=10, epsilon=10\% & 0.3002 & 0.2220 & 0.8160 & 0.57368 & 0.76230 & \textbf{0.04812} \\
HT/Q3 & K=10, epsilon=10\%, N=2 & 0.4094 & 0.3200 & 0.7900 & 0.15134 & 0.33432 & 0.14090 \\
HT/Q3 & K=10, epsilon=10\%, N=4 & 0.2684 & 0.2510 & 0.3940 & 0.71737 & 0.69831 & 0.80732 \\
HT/Q3 & K=10, epsilon=10\%, N=8 & 0.4366 & 0.3520 & 0.6850 & 0.09469 & 0.25189 & 0.30433 \\
HT/Q3 & K=10, epsilon=20\% & 0.4376 & 0.4200 & 0.5810 & 0.09286 & 0.11447 & 0.50308 \\
HT/Q3 & K=10, epsilon=30\% & 0.4384 & 0.4520 & 0.5850 & 0.09169 & 0.07421 & 0.50308 \\
HT/Q3 & K=20, epsilon=10\% & 0.3442 & 0.3420 & 0.5760 & 0.37269 & 0.28405 & 0.50308 \\
HT/Q3 & K=20, epsilon=20\% & 0.2968 & 0.2740 & 0.5000 & 0.58955 & 0.59139 & 0.55733 \\
HT/Q3 & K=20, epsilon=30\% & 0.2426 & 0.2010 & 0.4380 & 0.82032 & 0.84840 & 0.74780 \\
HT/Q3 & K=30, epsilon=10\% & 0.3452 & 0.2710 & 0.6180 & 0.36846 & 0.64660 & 0.37579 \\
HT/Q3 & K=30, epsilon=20\% & 0.2946 & 0.2740 & 0.6490 & 0.60011 & 0.59139 & 0.30433 \\
HT/Q3 & K=30, epsilon=30\% & 0.2166 & 0.2170 & 0.3490 & 0.89929 & 0.79377 & 0.89351 \\
HT/GPT20 & K=10, epsilon=20\% & 0.8276 & 0.8147 & 0.9080 & 0.56379 & 0.78986 & 0.59982 \\
HT/GPT20 & K=10, epsilon=20\%, N=2 & 0.8500 & 0.8334 & 0.9223 & 0.34271 & 0.68064 & 0.41158 \\
HT/GPT20 & K=10, epsilon=20\%, N=4 & 0.8381 & 0.8648 & 0.9028 & 0.46355 & 0.43606 & 0.67431 \\
HT/GPT20 & K=10, epsilon=30\% & 0.8824 & 0.8861 & 0.9086 & 0.08463 & 0.31523 & 0.59982 \\
HT/GPT20 & K=20, epsilon=20\% & 0.7643 & 0.7957 & 0.8911 & 0.93140 & 0.83698 & 0.78826 \\
HT/GPT20 & K=20, epsilon=30\% & 0.8431 & 0.8483 & 0.8722 & 0.41274 & 0.49768 & 0.94127 \\
HT/GPT-120B & K=10, epsilon=20\% & 0.7440 & 0.7201 & 0.8199 & 0.73000 & 0.68400 & 0.92100 \\
HT/GPT-120B & K=10, epsilon=30\% & 0.7594 & 0.7185 & 0.9333 & 0.62100 & 0.68400 & 0.28600 \\
HT/GPT-120B & K=20, epsilon=20\% & 0.8084 & 0.8240 & 0.9350 & 0.22700 & 0.30700 & 0.28600 \\
HT/GPT-120B & K=20, epsilon=30\% & 0.7667 & 0.7617 & 0.8092 & 0.55900 & 0.43700 & 0.92100 \\
SAI/Q2.5 & K=10, epsilon=10\% & 0.9430 & 0.9457 & 0.9470 & 0.36388 & 0.40912 & 0.93558 \\
SAI/Q2.5 & K=10, epsilon=20\% & 0.9425 & 0.9405 & 0.9512 & 0.39766 & 0.61204 & 0.74869 \\
SAI/Q2.5 & K=10, epsilon=30\% & 0.9462 & 0.9499 & 0.9525 & 0.17781 & 0.17893 & 0.47375 \\
SAI/Q2.5 & K=20, epsilon=10\% & 0.9458 & 0.9499 & 0.9617 & 0.19775 & 0.17893 & \textbf{0.04840} \\
SAI/Q2.5 & K=20, epsilon=10\%, N=2 & 0.9407 & 0.9497 & 0.9576 & 0.51850 & 0.17893 & 0.22680 \\
SAI/Q2.5 & K=20, epsilon=10\%, N=4 & 0.9354 & 0.9361 & 0.9463 & 0.81608 & 0.83689 & 0.95114 \\
SAI/Q2.5 & K=20, epsilon=10\%, N=8 & 0.9442 & 0.9473 & 0.9639 & 0.28502 & 0.33573 & \textbf{0.04840} \\
SAI/Q2.5 & K=20, epsilon=20\% & 0.9466 & 0.9510 & 0.9518 & 0.15848 & 0.10397 & 0.60877 \\
SAI/Q2.5 & K=20, epsilon=30\% & 0.9342 & 0.9365 & 0.9475 & 0.85971 & 0.83689 & 0.93004 \\
SAI/Q2.5 & K=30, epsilon=10\% & 0.9433 & 0.9448 & 0.9511 & 0.34454 & 0.44545 & 0.74869 \\
SAI/Q2.5 & K=30, epsilon=20\% & 0.9405 & 0.9367 & 0.9522 & 0.52897 & 0.82340 & 0.58486 \\
SAI/Q2.5 & K=30, epsilon=30\% & 0.9384 & 0.9360 & 0.9567 & 0.66251 & 0.83689 & 0.22680 \\
SAI/Q3 & K=10, epsilon=10\% & 0.9475 & 0.9484 & 0.9533 & 0.41961 & 0.40567 & 0.80509 \\
SAI/Q3 & K=10, epsilon=20\% & 0.9462 & 0.9463 & 0.9580 & 0.50432 & 0.44256 & 0.72794 \\
SAI/Q3 & K=10, epsilon=30\% & 0.9530 & 0.9497 & 0.9702 & 0.11903 & 0.38751 & 0.09589 \\
SAI/Q3 & K=10, epsilon=30\%, N=2 & 0.9480 & 0.9463 & 0.9524 & 0.38582 & 0.44256 & 0.85371 \\
SAI/Q3 & K=10, epsilon=30\%, N=4 & 0.9465 & 0.9447 & 0.9583 & 0.48515 & 0.51771 & 0.71010 \\
SAI/Q3 & K=10, epsilon=30\%, N=8 & 0.9408 & 0.9397 & 0.9481 & 0.82553 & 0.83745 & 0.95491 \\
SAI/Q3 & K=20, epsilon=10\% & 0.9538 & 0.9523 & 0.9648 & 0.09218 & 0.18901 & 0.30253 \\
SAI/Q3 & K=20, epsilon=20\% & 0.9473 & 0.9462 & 0.9571 & 0.42745 & 0.46077 & 0.74515 \\
SAI/Q3 & K=20, epsilon=30\% & 0.9396 & 0.9406 & 0.9565 & 0.87131 & 0.74802 & 0.74515 \\
SAI/Q3 & K=30, epsilon=10\% & 0.9417 & 0.9390 & 0.9559 & 0.77926 & 0.83745 & 0.76110 \\
SAI/Q3 & K=30, epsilon=20\% & 0.9414 & 0.9394 & 0.9555 & 0.79521 & 0.83745 & 0.76110 \\
SAI/Q3 & K=30, epsilon=30\% & 0.9476 & 0.9466 & 0.9517 & 0.41308 & 0.44256 & 0.87386 \\
SAI/GPT20 & K=10, epsilon=10\% & 0.9276 & 0.9274 & 0.9385 & 0.92757 & 0.62422 & 0.97795 \\
SAI/GPT20 & K=10, epsilon=20\% & 0.9419 & 0.9336 & 0.9684 & 0.49269 & 0.62422 & 0.59723 \\
SAI/GPT20 & K=10, epsilon=30\% & 0.9417 & 0.9387 & 0.9606 & 0.49954 & 0.56195 & 0.78812 \\
SAI/GPT20 & K=20, epsilon=10\% & 0.9486 & 0.9427 & 0.9696 & 0.24124 & 0.50014 & 0.51003 \\
SAI/GPT20 & K=20, epsilon=20\% & 0.9457 & 0.9488 & 0.9602 & 0.34475 & 0.43836 & 0.78812 \\
SAI/GPT20 & K=20, epsilon=30\% & 0.9266 & 0.9272 & 0.9485 & 0.94348 & 0.68273 & 0.95708 \\
SAI/GPT20 & K=30, epsilon=10\% & 0.9393 & 0.9326 & 0.9540 & 0.59249 & 0.62422 & 0.86877 \\
SAI/GPT20 & K=30, epsilon=20\% & 0.9255 & 0.9290 & 0.9348 & 0.95767 & 0.62422 & 0.98511 \\
SAI/GPT20 & K=30, epsilon=30\% & 0.9375 & 0.9361 & 0.9619 & 0.66309 & 0.62422 & 0.73484 \\
SAI/GPT-120B & K=10, epsilon=20\% & 0.9253 & 0.9268 & 0.9318 & 0.62300 & 0.50200 & 0.82400 \\
SAI/GPT-120B & K=10, epsilon=30\% & 0.9330 & 0.9328 & 0.9433 & 0.10500 & 0.13400 & 0.17000 \\
SAI/GPT-120B & K=20, epsilon=20\% & 0.9242 & 0.9273 & 0.9340 & 0.69000 & 0.44900 & 0.78500 \\
SAI/GPT-120B & K=20, epsilon=30\% & 0.9279 & 0.9268 & 0.9413 & 0.42900 & 0.50200 & 0.42300 \\
\end{longtable}
\endgroup

\begingroup
\scriptsize
\setlength{\tabcolsep}{2pt}
\renewcommand{\arraystretch}{1.08}
\begin{longtable}{@{}lp{0.50\textwidth}rrrrrr@{}}
\caption{\textbf{Epsilon-greedy + single child ($N=1$).} Uses the epsilon-greedy harness with one generated child per update step.}\label{tab:results_single_child}\\
\toprule
\multicolumn{2}{c}{} & \multicolumn{3}{c}{Observed statistic} & \multicolumn{3}{c}{One-sided p-value}\\
\cmidrule(lr){3-5}\cmidrule(l){6-8}
Pair & Configuration & Mean & Median & Max & $p_{\mathrm{mean}}$ & $p_{\mathrm{median}}$ & $p_{\mathrm{max}}$ \\
\midrule
\endfirsthead
\multicolumn{8}{c}{\tablename\ \thetable\ (continued)}\\
\toprule
\multicolumn{2}{c}{} & \multicolumn{3}{c}{Observed statistic} & \multicolumn{3}{c}{One-sided p-value}\\
\cmidrule(lr){3-5}\cmidrule(l){6-8}
Pair & Configuration & Mean & Median & Max & $p_{\mathrm{mean}}$ & $p_{\mathrm{median}}$ & $p_{\mathrm{max}}$ \\
\midrule
\endhead
\midrule
\multicolumn{8}{r}{Continued on next page}\\
\endfoot
\bottomrule
\endlastfoot
CP/Q2.5 & K=30, epsilon=10\% & 0.6667 & 0.6936 & 0.7417 & 0.91621 & 0.76343 & 0.76270 \\
CP/Q3 & K=10, epsilon=10\% & 0.8240 & 0.7835 & 0.9463 & 0.76467 & 0.87456 & 0.52763 \\
CP/GPT20 & K=1, epsilon=30\% & 0.9612 & 0.9644 & 0.9665 & 0.83787 & 0.68277 & 0.96825 \\
HT/Q2.5 & K=20, epsilon=20\% & 0.2090 & 0.2550 & 0.3330 & \textbf{0.03107} & \textbf{0.01162} & 0.30511 \\
HT/Q3 & K=10, epsilon=10\% & 0.2328 & 0.2070 & 0.4280 & 0.85351 & 0.82232 & 0.74780 \\
HT/GPT20 & K=10, epsilon=20\% & 0.9175 & 0.9530 & 0.9673 & \textbf{0.00213} & \textbf{$<10^{-5}$} & \textbf{$<10^{-5}$} \\
SAI/Q2.5 & K=20, epsilon=10\% & 0.9297 & 0.9337 & 0.9495 & 0.96352 & 0.86292 & 0.84587 \\
SAI/Q3 & K=10, epsilon=30\% & 0.9452 & 0.9473 & 0.9567 & 0.57139 & 0.44256 & 0.74515 \\
SAI/GPT20 & K=20, epsilon=0\% & 0.9446 & 0.9485 & 0.9646 & 0.38498 & 0.43836 & 0.73484 \\
\end{longtable}
\endgroup

\begingroup
\scriptsize
\setlength{\tabcolsep}{2pt}
\renewcommand{\arraystretch}{1.08}
\begin{longtable}{@{}lp{0.50\textwidth}rrrrrr@{}}
\caption{\textbf{Basic OpenEvolve (one island).} Canonical one-island OpenEvolve runs.}\label{tab:results_openevolve_one_island}\\
\toprule
\multicolumn{2}{c}{} & \multicolumn{3}{c}{Observed statistic} & \multicolumn{3}{c}{One-sided p-value}\\
\cmidrule(lr){3-5}\cmidrule(l){6-8}
Pair & Configuration & Mean & Median & Max & $p_{\mathrm{mean}}$ & $p_{\mathrm{median}}$ & $p_{\mathrm{max}}$ \\
\midrule
\endfirsthead
\multicolumn{8}{c}{\tablename\ \thetable\ (continued)}\\
\toprule
\multicolumn{2}{c}{} & \multicolumn{3}{c}{Observed statistic} & \multicolumn{3}{c}{One-sided p-value}\\
\cmidrule(lr){3-5}\cmidrule(l){6-8}
Pair & Configuration & Mean & Median & Max & $p_{\mathrm{mean}}$ & $p_{\mathrm{median}}$ & $p_{\mathrm{max}}$ \\
\midrule
\endhead
\midrule
\multicolumn{8}{r}{Continued on next page}\\
\endfoot
\bottomrule
\endlastfoot
CP/Q2.5 & K=30, epsilon=10\% & 0.4423 & 0.4246 & 0.5632 & 1.00000 & 1.00000 & 1.00000 \\
CP/Q3 & K=10, epsilon=10\% & 0.7030 & 0.7028 & 0.8029 & 0.99995 & 0.99991 & 0.99692 \\
CP/GPT20 & K=1, epsilon=30\% & 0.9438 & 0.9551 & 0.9689 & 0.99943 & 0.91425 & 0.94119 \\
CP/GPT-120B & K=1, epsilon=0\% & 0.9532 & 0.9581 & 0.9648 & 0.96800 & 0.25300 & 0.85800 \\
HT/Q2.5 & K=20, epsilon=20\% & 0.0036 & 0.0000 & 0.0168 & 1.00000 & 1.00000 & 0.99993 \\
HT/Q3 & K=10, epsilon=10\% & 0.5890 & 0.6770 & 0.7810 & \textbf{0.00214} & \textbf{0.00352} & 0.14090 \\
HT/GPT20 & K=10, epsilon=20\% & 0.6522 & 0.6572 & 0.8304 & 0.99976 & 0.99964 & 0.98996 \\
HT/GPT-120B & K=1, epsilon=0\% & 0.7535 & 0.6967 & 0.9134 & 0.66200 & 0.90600 & 0.50800 \\
SAI/Q2.5 & K=20, epsilon=10\% & 0.9135 & 0.9265 & 0.9318 & 1.00000 & 0.96381 & 0.99898 \\
SAI/Q3 & K=10, epsilon=30\% & 0.9121 & 0.9156 & 0.9364 & 1.00000 & 0.99976 & 0.99982 \\
SAI/GPT20 & K=20, epsilon=0\% & 0.9257 & 0.9256 & 0.9411 & 0.95488 & 0.73931 & 0.96924 \\
SAI/GPT-120B & K=1, epsilon=0\% & 0.9183 & 0.9139 & 0.9330 & 0.96300 & 0.99100 & 0.78500 \\
\end{longtable}
\endgroup

\begingroup
\scriptsize
\setlength{\tabcolsep}{2pt}
\renewcommand{\arraystretch}{1.08}
\begin{longtable}{@{}lp{0.50\textwidth}rrrrrr@{}}
\caption{\textbf{OpenEvolve + multiple islands.} Extends OpenEvolve with multiple interacting islands.}\label{tab:results_openevolve_multi_island}\\
\toprule
\multicolumn{2}{c}{} & \multicolumn{3}{c}{Observed statistic} & \multicolumn{3}{c}{One-sided p-value}\\
\cmidrule(lr){3-5}\cmidrule(l){6-8}
Pair & Configuration & Mean & Median & Max & $p_{\mathrm{mean}}$ & $p_{\mathrm{median}}$ & $p_{\mathrm{max}}$ \\
\midrule
\endfirsthead
\multicolumn{8}{c}{\tablename\ \thetable\ (continued)}\\
\toprule
\multicolumn{2}{c}{} & \multicolumn{3}{c}{Observed statistic} & \multicolumn{3}{c}{One-sided p-value}\\
\cmidrule(lr){3-5}\cmidrule(l){6-8}
Pair & Configuration & Mean & Median & Max & $p_{\mathrm{mean}}$ & $p_{\mathrm{median}}$ & $p_{\mathrm{max}}$ \\
\midrule
\endhead
\midrule
\multicolumn{8}{r}{Continued on next page}\\
\endfoot
\bottomrule
\endlastfoot
CP/Q2.5 & 4 islands, K=30, epsilon=10\% & 0.6114 & 0.6167 & 0.6392 & 0.99888 & 0.99561 & 0.99992 \\
CP/Q3 & 4 islands, K=10, epsilon=10\% & 0.7398 & 0.7440 & 0.8146 & 0.99866 & 0.99141 & 0.99579 \\
CP/GPT20 & 3 islands, K=1, epsilon=30\% & 0.9534 & 0.9509 & 0.9709 & 0.97995 & 0.99127 & 0.86832 \\
CP/GPT-120B & 4 islands, K=1, epsilon=0\% & 0.9587 & 0.9578 & 0.9693 & 0.73600 & 0.31200 & 0.81900 \\
HT/Q2.5 & 4 islands, K=20, epsilon=20\% & 0.0000 & 0.0000 & 0.0002 & 1.00000 & 1.00000 & 1.00000 \\
HT/Q3 & 4 islands, K=10, epsilon=10\% & 0.1164 & 0.0360 & 0.3840 & 0.99823 & 0.99977 & 0.83360 \\
HT/GPT20 & 3 islands, K=10, epsilon=20\% & 0.8916 & 0.9239 & 0.9567 & \textbf{0.04449} & \textbf{0.00855} & \textbf{$<10^{-5}$} \\
HT/GPT-120B & 4 islands, K=1, epsilon=0\% & 0.6943 & 0.6669 & 0.8829 & 0.95100 & 0.98200 & 0.66800 \\
SAI/Q2.5 & 4 islands, K=20, epsilon=10\% & 0.8943 & 0.8897 & 0.9135 & 1.00000 & 1.00000 & 1.00000 \\
SAI/Q3 & 4 islands, K=10, epsilon=30\% & 0.9127 & 0.9129 & 0.9192 & 1.00000 & 1.00000 & 1.00000 \\
SAI/GPT20 & 4 islands, K=20, epsilon=0\% & 0.9378 & 0.9377 & 0.9424 & 0.65028 & 0.56195 & 0.96924 \\
SAI/GPT-120B & 4 islands, K=1, epsilon=0\% & 0.9183 & 0.9137 & 0.9270 & 0.96300 & 0.99100 & 0.95600 \\
\end{longtable}
\endgroup

\begingroup
\scriptsize
\setlength{\tabcolsep}{2pt}
\renewcommand{\arraystretch}{1.08}
\begin{longtable}{@{}lp{0.50\textwidth}rrrrrr@{}}
\caption{\textbf{UCT.} Uses UCT parent selection, including breadth--depth and parent-count ablations.}\label{tab:results_uct}\\
\toprule
\multicolumn{2}{c}{} & \multicolumn{3}{c}{Observed statistic} & \multicolumn{3}{c}{One-sided p-value}\\
\cmidrule(lr){3-5}\cmidrule(l){6-8}
Pair & Configuration & Mean & Median & Max & $p_{\mathrm{mean}}$ & $p_{\mathrm{median}}$ & $p_{\mathrm{max}}$ \\
\midrule
\endfirsthead
\multicolumn{8}{c}{\tablename\ \thetable\ (continued)}\\
\toprule
\multicolumn{2}{c}{} & \multicolumn{3}{c}{Observed statistic} & \multicolumn{3}{c}{One-sided p-value}\\
\cmidrule(lr){3-5}\cmidrule(l){6-8}
Pair & Configuration & Mean & Median & Max & $p_{\mathrm{mean}}$ & $p_{\mathrm{median}}$ & $p_{\mathrm{max}}$ \\
\midrule
\endhead
\midrule
\multicolumn{8}{r}{Continued on next page}\\
\endfoot
\bottomrule
\endlastfoot
CP/Q2.5 & P=1, C=0.1 & 0.7410 & 0.7230 & 0.7850 & 0.13680 & 0.31851 & 0.44275 \\
CP/Q2.5 & P=1, C=0.1, N=2 & 0.6726 & 0.6610 & 0.7980 & 0.88342 & 0.94107 & 0.30482 \\
CP/Q2.5 & P=1, C=0.1, N=4 & 0.6240 & 0.6330 & 0.7120 & 0.99619 & 0.98521 & 0.95786 \\
CP/Q2.5 & P=1, C=0.1, N=8 & 0.6838 & 0.7000 & 0.7950 & 0.79952 & 0.69830 & 0.34064 \\
CP/Q2.5 & P=1, C=1 & 0.6876 & 0.6960 & 0.7760 & 0.76323 & 0.76343 & 0.47341 \\
CP/Q2.5 & P=1, C=1, N=2 & 0.6266 & 0.6780 & 0.7160 & 0.99516 & 0.87168 & 0.95382 \\
CP/Q2.5 & P=1, C=1, N=4 & 0.6478 & 0.6390 & 0.7250 & 0.97487 & 0.97756 & 0.92146 \\
CP/Q2.5 & P=1, C=1, N=8 & 0.6032 & 0.6170 & 0.6910 & 0.99966 & 0.99561 & 0.99655 \\
CP/Q2.5 & P=1, C=10 & 0.6788 & 0.7020 & 0.7420 & 0.84125 & 0.62712 & 0.76270 \\
CP/Q2.5 & P=2, C=0.1, N=4 & 0.5320 & 0.5620 & 0.7480 & 1.00000 & 0.99884 & 0.74639 \\
CP/Q2.5 & P=4, C=0.1, N=4 & 0.6296 & 0.7050 & 0.7280 & 0.99391 & 0.59158 & 0.91471 \\
CP/Q2.5 & P=4, C=0.1, N=8 & 0.7160 & 0.7150 & 0.7480 & 0.40574 & 0.44443 & 0.74639 \\
CP/Q3 & P=1, C=0.1 & 0.8160 & 0.8030 & 0.8920 & 0.82404 & 0.80883 & 0.90064 \\
CP/Q3 & P=1, C=1 & 0.8128 & 0.7820 & 0.9310 & 0.84487 & 0.90733 & 0.77671 \\
CP/Q3 & P=1, C=10 & 0.9076 & 0.9300 & 0.9430 & 0.05196 & 0.11381 & 0.52763 \\
CP/Q3 & P=1, C=10, N=2 & 0.7852 & 0.7630 & 0.8450 & 0.95840 & 0.95608 & 0.97678 \\
CP/Q3 & P=1, C=10, N=4 & 0.7522 & 0.7490 & 0.8090 & 0.99589 & 0.98200 & 0.99692 \\
CP/Q3 & P=1, C=10, N=8 & 0.8784 & 0.9320 & 0.9380 & 0.22664 & 0.10266 & 0.70924 \\
CP/Q3 & P=2, C=10, N=4 & 0.8376 & 0.8480 & 0.9040 & 0.64174 & 0.53908 & 0.89254 \\
CP/Q3 & P=4, C=10, N=4 & 0.8250 & 0.8280 & 0.9200 & 0.75643 & 0.73264 & 0.79139 \\
CP/Q3 & P=4, C=10, N=8 & 0.9058 & 0.9390 & 0.9400 & 0.05846 & 0.06536 & 0.57909 \\
CP/GPT20 & P=1, C=0.1 & 0.9702 & 0.9640 & 0.9930 & 0.36936 & 0.73934 & 0.41107 \\
CP/GPT20 & P=1, C=0.1, N=2 & 0.9696 & 0.9650 & 0.9930 & 0.40283 & 0.49901 & 0.41107 \\
CP/GPT20 & P=1, C=0.1, N=4 & 0.9448 & 0.9500 & 0.9610 & 0.99904 & 0.99728 & 0.99868 \\
CP/GPT20 & P=1, C=1 & 0.9566 & 0.9550 & 0.9640 & 0.94627 & 0.94302 & 0.99303 \\
CP/GPT20 & P=1, C=10 & 0.9460 & 0.9450 & 0.9690 & 0.99850 & 0.99728 & 0.92242 \\
CP/GPT20 & P=2, C=0.1, N=2 & 0.9576 & 0.9640 & 0.9700 & 0.93082 & 0.73934 & 0.89798 \\
CP/GPT20 & P=4, C=0.1 & 0.9740 & 0.9730 & 0.9860 & 0.18798 & 0.16236 & 0.51342 \\
CP/GPT20 & P=4, C=0.1, N=2 & 0.9698 & 0.9720 & 0.9790 & 0.39106 & 0.16236 & 0.73659 \\
CP/GPT-120B & P=1, C=0.1, N=4 & 0.9652 & 0.9560 & 0.9870 & 0.19000 & 0.86400 & 0.11700 \\
CP/GPT-120B & P=2, C=0.1, N=2 & 0.9556 & 0.9560 & 0.9640 & 0.84300 & 0.86400 & 0.88000 \\
CP/GPT-120B & P=4, C=0.1, N=2 & 0.9690 & 0.9640 & 0.9890 & 0.06900 & 0.20900 & 0.11700 \\
HT/Q2.5 & P=1, C=0.1 & 0.1408 & 0.0880 & 0.4530 & 0.22304 & 0.40512 & \textbf{$<10^{-5}$} \\
HT/Q2.5 & P=1, C=1 & 0.0758 & 0.0300 & 0.2170 & 0.70197 & 0.91657 & 0.55556 \\
HT/Q2.5 & P=1, C=10 & 0.1556 & 0.1680 & 0.3390 & 0.15217 & \textbf{0.04405} & 0.30511 \\
HT/Q2.5 & P=1, C=10, N=2 & 0.1476 & 0.1260 & 0.4450 & 0.18767 & 0.10327 & 0.09626 \\
HT/Q2.5 & P=1, C=10, N=4 & 0.0884 & 0.0540 & 0.1920 & 0.59141 & 0.66224 & 0.58066 \\
HT/Q2.5 & P=1, C=10, N=8 & 0.1240 & 0.0710 & 0.4300 & 0.32310 & 0.44176 & 0.09626 \\
HT/Q2.5 & P=2, C=10, N=4 & 0.1412 & 0.1430 & 0.2240 & 0.22071 & 0.09362 & 0.47177 \\
HT/Q2.5 & P=4, C=10, N=4 & 0.0580 & 0.0160 & 0.2310 & 0.85343 & 0.98186 & 0.47177 \\
HT/Q2.5 & P=4, C=10, N=8 & 0.1578 & 0.0820 & 0.4520 & 0.14372 & 0.40512 & 0.09626 \\
HT/Q3 & P=1, C=0.1 & 0.2328 & 0.2530 & 0.4930 & 0.85351 & 0.69831 & 0.55733 \\
HT/Q3 & P=1, C=1 & 0.4104 & 0.2970 & 0.7390 & 0.14890 & 0.49807 & 0.14090 \\
HT/Q3 & P=1, C=1, N=2 & 0.2068 & 0.1700 & 0.3880 & 0.92248 & 0.91492 & 0.83360 \\
HT/Q3 & P=1, C=1, N=4 & 0.2642 & 0.2490 & 0.4520 & 0.73473 & 0.73081 & 0.71236 \\
HT/Q3 & P=1, C=1, N=8 & 0.2192 & 0.2620 & 0.3160 & 0.89298 & 0.66464 & 0.95845 \\
HT/Q3 & P=1, C=10 & 0.2494 & 0.2090 & 0.3790 & 0.79521 & 0.80802 & 0.86636 \\
HT/Q3 & P=2, C=1, N=4 & 0.5394 & 0.5660 & 0.8230 & \textbf{0.00913} & \textbf{0.01798} & \textbf{0.04812} \\
HT/Q3 & P=4, C=1, N=4 & 0.1666 & 0.1050 & 0.4520 & 0.97927 & 0.98868 & 0.71236 \\
HT/Q3 & P=4, C=1, N=8 & 0.3014 & 0.2590 & 0.6810 & 0.56770 & 0.66464 & 0.30433 \\
HT/GPT20 & P=1, C=0.1 & 0.8878 & 0.9220 & 0.9650 & 0.05911 & \textbf{0.00855} & \textbf{$<10^{-5}$} \\
HT/GPT20 & P=1, C=1 & 0.9044 & 0.9170 & 0.9420 & \textbf{0.01281} & \textbf{0.01980} & \textbf{$<10^{-5}$} \\
HT/GPT20 & P=1, C=1, N=2 & 0.9010 & 0.9150 & 0.9510 & \textbf{0.01876} & \textbf{0.01980} & \textbf{$<10^{-5}$} \\
HT/GPT20 & P=1, C=1, N=4 & 0.8092 & 0.8470 & 0.9820 & 0.71614 & 0.49768 & \textbf{$<10^{-5}$} \\
HT/GPT20 & P=1, C=10 & 0.8174 & 0.8190 & 0.9610 & 0.65252 & 0.78986 & \textbf{$<10^{-5}$} \\
HT/GPT20 & P=2, C=1, N=2 & 0.8844 & 0.8940 & 0.9370 & 0.07407 & 0.12120 & 0.29467 \\
HT/GPT20 & P=4, C=1 & 0.9238 & 0.9280 & 0.9640 & \textbf{0.00051} & \textbf{0.00855} & \textbf{$<10^{-5}$} \\
HT/GPT20 & P=4, C=1, N=2 & 0.8586 & 0.8730 & 0.8910 & 0.26189 & 0.37360 & 0.78826 \\
HT/GPT-120B & P=1, C=1, N=4 & 0.8678 & 0.8380 & 0.9430 & \textbf{0.01700} & 0.15300 & 0.36300 \\
HT/GPT-120B & P=2, C=1, N=2 & 0.7874 & 0.8480 & 0.8860 & 0.34900 & 0.12600 & 0.66500 \\
HT/GPT-120B & P=4, C=1, N=2 & 0.8586 & 0.8330 & 0.9460 & \textbf{0.03500} & 0.18100 & 0.19700 \\
SAI/Q2.5 & P=1, C=0.1 & 0.9340 & 0.9370 & 0.9510 & 0.86687 & 0.80903 & 0.76539 \\
SAI/Q2.5 & P=1, C=1 & 0.9440 & 0.9430 & 0.9540 & 0.29986 & 0.52074 & 0.34117 \\
SAI/Q2.5 & P=1, C=1, N=2 & 0.9378 & 0.9340 & 0.9610 & 0.69619 & 0.85026 & 0.14083 \\
SAI/Q2.5 & P=1, C=1, N=4 & 0.9380 & 0.9390 & 0.9480 & 0.68483 & 0.73268 & 0.90266 \\
SAI/Q2.5 & P=1, C=1, N=8 & 0.9390 & 0.9410 & 0.9460 & 0.62576 & 0.57561 & 0.95114 \\
SAI/Q2.5 & P=1, C=10 & 0.9394 & 0.9450 & 0.9520 & 0.60007 & 0.44545 & 0.58486 \\
SAI/Q2.5 & P=2, C=1, N=4 & 0.9270 & 0.9150 & 0.9620 & 0.98706 & 0.99726 & \textbf{0.04840} \\
SAI/Q2.5 & P=4, C=1, N=4 & 0.9266 & 0.9400 & 0.9470 & 0.98909 & 0.66541 & 0.93558 \\
SAI/Q3 & P=1, C=0.1 & 0.9476 & 0.9430 & 0.9650 & 0.41034 & 0.61084 & 0.30253 \\
SAI/Q3 & P=1, C=0.1, N=2 & 0.9230 & 0.9280 & 0.9670 & 0.99992 & 0.99338 & 0.18164 \\
SAI/Q3 & P=1, C=0.1, N=4 & 0.9442 & 0.9390 & 0.9660 & 0.63883 & 0.83745 & 0.26412 \\
SAI/Q3 & P=1, C=0.1, N=8 & 0.9444 & 0.9390 & 0.9600 & 0.62564 & 0.83745 & 0.58001 \\
SAI/Q3 & P=1, C=1 & 0.9396 & 0.9390 & 0.9450 & 0.86991 & 0.83745 & 0.96881 \\
SAI/Q3 & P=1, C=10 & 0.9416 & 0.9380 & 0.9550 & 0.78632 & 0.90631 & 0.76110 \\
SAI/Q3 & P=2, C=0.1, N=4 & 0.8278 & 0.9420 & 0.9440 & 1.00000 & 0.66493 & 0.97445 \\
SAI/Q3 & P=4, C=0.1, N=4 & 0.9300 & 0.9390 & 0.9460 & 0.99624 & 0.83745 & 0.96254 \\
SAI/GPT20 & P=1, C=0.1 & 0.9570 & 0.9570 & 0.9660 & 0.05656 & 0.16345 & 0.67160 \\
SAI/GPT20 & P=1, C=0.1, N=2 & 0.9406 & 0.9430 & 0.9590 & 0.54298 & 0.50014 & 0.78812 \\
SAI/GPT20 & P=1, C=0.1, N=4 & 0.9432 & 0.9430 & 0.9730 & 0.44120 & 0.50014 & 0.29137 \\
SAI/GPT20 & P=1, C=1 & 0.9444 & 0.9480 & 0.9610 & 0.39458 & 0.50014 & 0.78812 \\
SAI/GPT20 & P=1, C=10 & 0.9320 & 0.9340 & 0.9450 & 0.83972 & 0.62422 & 0.96924 \\
SAI/GPT20 & P=2, C=0.1, N=2 & 0.9374 & 0.9400 & 0.9590 & 0.66544 & 0.56195 & 0.78812 \\
SAI/GPT20 & P=4, C=0.1, N=2 & 0.9474 & 0.9640 & 0.9710 & 0.27980 & 0.08733 & 0.40929 \\
SAI/GPT-120B & P=1, C=0.1, N=4 & 0.9316 & 0.9270 & 0.9590 & 0.19800 & 0.38500 & 0.13700 \\
SAI/GPT-120B & P=2, C=0.1, N=2 & 0.9276 & 0.9160 & 0.9690 & 0.44000 & 0.97700 & \textbf{$<10^{-5}$} \\
SAI/GPT-120B & P=4, C=0.1, N=2 & 0.9314 & 0.9300 & 0.9480 & 0.20400 & 0.22300 & 0.25700 \\
\end{longtable}
\endgroup

\begingroup
\scriptsize
\setlength{\tabcolsep}{2pt}
\renewcommand{\arraystretch}{1.08}
\begin{longtable}{@{}lp{0.50\textwidth}rrrrrr@{}}
\caption{\textbf{PUCT.} Uses PUCT parent selection with one parent program ($P=1$).}\label{tab:results_puct}\\
\toprule
\multicolumn{2}{c}{} & \multicolumn{3}{c}{Observed statistic} & \multicolumn{3}{c}{One-sided p-value}\\
\cmidrule(lr){3-5}\cmidrule(l){6-8}
Pair & Configuration & Mean & Median & Max & $p_{\mathrm{mean}}$ & $p_{\mathrm{median}}$ & $p_{\mathrm{max}}$ \\
\midrule
\endfirsthead
\multicolumn{8}{c}{\tablename\ \thetable\ (continued)}\\
\toprule
\multicolumn{2}{c}{} & \multicolumn{3}{c}{Observed statistic} & \multicolumn{3}{c}{One-sided p-value}\\
\cmidrule(lr){3-5}\cmidrule(l){6-8}
Pair & Configuration & Mean & Median & Max & $p_{\mathrm{mean}}$ & $p_{\mathrm{median}}$ & $p_{\mathrm{max}}$ \\
\midrule
\endhead
\midrule
\multicolumn{8}{r}{Continued on next page}\\
\endfoot
\bottomrule
\endlastfoot
CP/Q2.5 & P=1, C=0.1 & 0.6568 & 0.6540 & 0.8020 & 0.95431 & 0.95594 & 0.30482 \\
CP/Q2.5 & P=1, C=1 & 0.7098 & 0.6850 & 0.9110 & 0.49005 & 0.83474 & \textbf{0.04919} \\
CP/Q2.5 & P=1, C=1, N=2 & 0.5992 & 0.6050 & 0.7160 & 0.99979 & 0.99803 & 0.95382 \\
CP/Q2.5 & P=1, C=1, N=4 & 0.6266 & 0.6440 & 0.7340 & 0.99516 & 0.97289 & 0.84270 \\
CP/Q2.5 & P=1, C=1, N=8 & 0.7626 & 0.7380 & 0.8740 & \textbf{0.03793} & 0.12700 & \textbf{0.04919} \\
CP/Q2.5 & P=1, C=10 & 0.7378 & 0.7310 & 0.8160 & 0.16189 & 0.23640 & 0.22610 \\
CP/Q2.5 & P=1, C=10, N=2 & 0.5504 & 0.5830 & 0.6870 & 1.00000 & 0.99884 & 0.99751 \\
CP/Q2.5 & P=1, C=10, N=4 & 0.6248 & 0.6840 & 0.7030 & 0.99586 & 0.83474 & 0.98546 \\
CP/Q2.5 & P=1, C=10, N=8 & 0.6650 & 0.6760 & 0.7480 & 0.92427 & 0.88334 & 0.74639 \\
CP/Q2.5 & P=1, C=100 & 0.6886 & 0.7110 & 0.7270 & 0.75310 & 0.44443 & 0.92146 \\
CP/Q3 & P=1, C=0.1 & 0.8696 & 0.8540 & 0.9510 & 0.30819 & 0.48266 & 0.37270 \\
CP/Q3 & P=1, C=1 & 0.8376 & 0.8200 & 0.9420 & 0.64174 & 0.76412 & 0.55379 \\
CP/Q3 & P=1, C=10 & 0.8758 & 0.8750 & 0.9410 & 0.24919 & 0.38896 & 0.57909 \\
CP/Q3 & P=1, C=10, N=2 & 0.7670 & 0.7730 & 0.8560 & 0.98711 & 0.93390 & 0.96492 \\
CP/Q3 & P=1, C=10, N=4 & 0.7548 & 0.7470 & 0.8240 & 0.99499 & 0.98200 & 0.99417 \\
CP/Q3 & P=1, C=10, N=8 & 0.7866 & 0.7670 & 0.9410 & 0.95519 & 0.94164 & 0.57909 \\
CP/GPT20 & P=1, C=0.1 & 0.9464 & 0.9630 & 0.9660 & 0.99832 & 0.83853 & 0.96825 \\
CP/GPT20 & P=1, C=1 & 0.9658 & 0.9640 & 0.9980 & 0.62339 & 0.73934 & 0.15662 \\
CP/GPT20 & P=1, C=10 & 0.9714 & 0.9660 & 0.9840 & 0.30619 & 0.49901 & 0.51342 \\
CP/GPT20 & P=1, C=10, N=2 & 0.9684 & 0.9640 & 0.9950 & 0.47275 & 0.73934 & 0.41107 \\
CP/GPT20 & P=1, C=10, N=4 & 0.9782 & 0.9780 & 0.9920 & 0.06822 & 0.08764 & 0.41107 \\
CP/GPT-120B & P=1, C=10, N=4 & 0.9656 & 0.9630 & 0.9870 & 0.17700 & 0.20900 & 0.11700 \\
HT/Q2.5 & P=1, C=0.1 & 0.1034 & 0.0420 & 0.3390 & 0.47112 & 0.82297 & 0.30511 \\
HT/Q2.5 & P=1, C=1 & 0.2224 & 0.1130 & 0.4390 & \textbf{0.01923} & 0.20401 & 0.09626 \\
HT/Q2.5 & P=1, C=1, N=2 & 0.0844 & 0.0640 & 0.2000 & 0.62629 & 0.55420 & 0.55556 \\
HT/Q2.5 & P=1, C=1, N=4 & 0.0810 & 0.0580 & 0.2170 & 0.65577 & 0.60874 & 0.55556 \\
HT/Q2.5 & P=1, C=1, N=8 & 0.0834 & 0.1160 & 0.1630 & 0.63452 & 0.18953 & 0.62742 \\
HT/Q2.5 & P=1, C=10 & 0.1028 & 0.0920 & 0.2330 & 0.47551 & 0.38619 & 0.47177 \\
HT/Q3 & P=1, C=0.1 & 0.2102 & 0.1820 & 0.4520 & 0.91501 & 0.89516 & 0.71236 \\
HT/Q3 & P=1, C=1 & 0.3318 & 0.2980 & 0.5080 & 0.42711 & 0.49807 & 0.55733 \\
HT/Q3 & P=1, C=1, N=2 & 0.2268 & 0.2560 & 0.4660 & 0.87116 & 0.66464 & 0.65256 \\
HT/Q3 & P=1, C=1, N=4 & 0.2552 & 0.2910 & 0.4450 & 0.77199 & 0.51661 & 0.73052 \\
HT/Q3 & P=1, C=1, N=8 & 0.2922 & 0.2570 & 0.6430 & 0.61081 & 0.66464 & 0.30433 \\
HT/Q3 & P=1, C=10 & 0.3062 & 0.3750 & 0.4620 & 0.54582 & 0.22132 & 0.65256 \\
HT/GPT20 & P=1, C=0.1 & 0.8398 & 0.8580 & 0.8970 & 0.44654 & 0.43606 & 0.67431 \\
HT/GPT20 & P=1, C=1 & 0.8986 & 0.8970 & 0.9790 & \textbf{0.02391} & 0.05778 & \textbf{$<10^{-5}$} \\
HT/GPT20 & P=1, C=1, N=2 & 0.8764 & 0.8920 & 0.9640 & 0.12006 & 0.12120 & \textbf{$<10^{-5}$} \\
HT/GPT20 & P=1, C=1, N=4 & 0.9316 & 0.9490 & 0.9730 & \textbf{0.00004} & \textbf{$<10^{-5}$} & \textbf{$<10^{-5}$} \\
HT/GPT20 & P=1, C=10 & 0.8062 & 0.8460 & 0.9180 & 0.73715 & 0.49768 & 0.51170 \\
HT/GPT-120B & P=1, C=1, N=4 & 0.8388 & 0.8780 & 0.9690 & 0.08700 & 0.05300 & \textbf{$<10^{-5}$} \\
SAI/Q2.5 & P=1, C=0.1 & 0.9346 & 0.9460 & 0.9520 & 0.84563 & 0.40912 & 0.58486 \\
SAI/Q2.5 & P=1, C=1 & 0.9448 & 0.9450 & 0.9580 & 0.25160 & 0.44545 & 0.22680 \\
SAI/Q2.5 & P=1, C=10 & 0.9514 & 0.9510 & 0.9600 & \textbf{0.02104} & 0.10397 & 0.18462 \\
SAI/Q2.5 & P=1, C=10, N=2 & 0.9196 & 0.9190 & 0.9390 & 0.99965 & 0.99189 & 0.99342 \\
SAI/Q2.5 & P=1, C=10, N=4 & 0.9190 & 0.9140 & 0.9520 & 0.99972 & 0.99895 & 0.58486 \\
SAI/Q2.5 & P=1, C=10, N=8 & 0.9222 & 0.9320 & 0.9430 & 0.99844 & 0.89762 & 0.97267 \\
SAI/Q3 & P=1, C=0.1 & 0.9434 & 0.9410 & 0.9630 & 0.68864 & 0.74802 & 0.37556 \\
SAI/Q3 & P=1, C=1 & 0.9436 & 0.9470 & 0.9490 & 0.67668 & 0.44256 & 0.94565 \\
SAI/Q3 & P=1, C=10 & 0.9516 & 0.9470 & 0.9680 & 0.17475 & 0.44256 & 0.14021 \\
SAI/Q3 & P=1, C=10, N=2 & 0.9394 & 0.9440 & 0.9650 & 0.87651 & 0.53645 & 0.30253 \\
SAI/Q3 & P=1, C=10, N=4 & 0.9334 & 0.9400 & 0.9520 & 0.98363 & 0.80900 & 0.86387 \\
SAI/Q3 & P=1, C=10, N=8 & 0.9438 & 0.9430 & 0.9600 & 0.66462 & 0.61084 & 0.58001 \\
SAI/GPT20 & P=1, C=0.1 & 0.9508 & 0.9610 & 0.9720 & 0.17744 & 0.12218 & 0.29137 \\
SAI/GPT20 & P=1, C=0.1, N=2 & 0.9530 & 0.9510 & 0.9700 & 0.12568 & 0.37733 & 0.40929 \\
SAI/GPT20 & P=1, C=0.1, N=4 & 0.9390 & 0.9510 & 0.9700 & 0.60347 & 0.37733 & 0.40929 \\
SAI/GPT20 & P=1, C=1 & 0.9488 & 0.9630 & 0.9660 & 0.23407 & 0.08733 & 0.67160 \\
SAI/GPT20 & P=1, C=10 & 0.9376 & 0.9360 & 0.9710 & 0.65755 & 0.62422 & 0.40929 \\
SAI/GPT-120B & P=1, C=0.1, N=4 & 0.9274 & 0.9310 & 0.9520 & 0.46000 & 0.17900 & 0.13700 \\
\end{longtable}
\endgroup

\begingroup
\scriptsize
\setlength{\tabcolsep}{2pt}
\renewcommand{\arraystretch}{1.08}
\begin{longtable}{@{}lp{0.50\textwidth}rrrrrr@{}}
\caption{\textbf{PUCT + multiple parents (TTT-Discover).} Extends PUCT to multiple parent programs ($P>1$), corresponding to the TTT-Discover endpoint.}\label{tab:results_puct_multi_parent}\\
\toprule
\multicolumn{2}{c}{} & \multicolumn{3}{c}{Observed statistic} & \multicolumn{3}{c}{One-sided p-value}\\
\cmidrule(lr){3-5}\cmidrule(l){6-8}
Pair & Configuration & Mean & Median & Max & $p_{\mathrm{mean}}$ & $p_{\mathrm{median}}$ & $p_{\mathrm{max}}$ \\
\midrule
\endfirsthead
\multicolumn{8}{c}{\tablename\ \thetable\ (continued)}\\
\toprule
\multicolumn{2}{c}{} & \multicolumn{3}{c}{Observed statistic} & \multicolumn{3}{c}{One-sided p-value}\\
\cmidrule(lr){3-5}\cmidrule(l){6-8}
Pair & Configuration & Mean & Median & Max & $p_{\mathrm{mean}}$ & $p_{\mathrm{median}}$ & $p_{\mathrm{max}}$ \\
\midrule
\endhead
\midrule
\multicolumn{8}{r}{Continued on next page}\\
\endfoot
\bottomrule
\endlastfoot
CP/Q2.5 & P=2, C=10, N=4 & 0.6116 & 0.6330 & 0.6720 & 0.99885 & 0.98521 & 0.99935 \\
CP/Q2.5 & P=4, C=10, N=4 & 0.6708 & 0.6900 & 0.8470 & 0.89449 & 0.80736 & 0.09572 \\
CP/Q2.5 & P=4, C=10, N=8 & 0.7276 & 0.7320 & 0.8120 & 0.26111 & 0.22147 & 0.26652 \\
CP/Q3 & P=2, C=10, N=4 & 0.8248 & 0.8060 & 0.9390 & 0.75796 & 0.80883 & 0.69032 \\
CP/Q3 & P=4, C=10, N=4 & 0.7424 & 0.7540 & 0.8100 & 0.99827 & 0.98200 & 0.99692 \\
CP/Q3 & P=4, C=10, N=8 & 0.8820 & 0.8790 & 0.9490 & 0.19675 & 0.33376 & 0.47015 \\
CP/GPT20 & P=2, C=10, N=2 & 0.9726 & 0.9700 & 0.9970 & 0.24804 & 0.26098 & 0.15662 \\
CP/GPT20 & P=4, C=10 & 0.9792 & 0.9950 & 0.9980 & 0.05085 & \textbf{0.00870} & 0.15662 \\
CP/GPT20 & P=4, C=10, N=2 & 0.9750 & 0.9800 & 0.9970 & 0.15236 & 0.05836 & 0.15662 \\
CP/GPT-120B & P=2, C=10, N=2 & 0.9554 & 0.9570 & 0.9580 & 0.85500 & 0.51500 & 0.88000 \\
CP/GPT-120B & P=4, C=10, N=2 & 0.9554 & 0.9570 & 0.9640 & 0.85500 & 0.51500 & 0.88000 \\
HT/Q2.5 & P=2, C=1, N=4 & 0.0702 & 0.0700 & 0.1490 & 0.75197 & 0.47918 & 0.72934 \\
HT/Q2.5 & P=4, C=1, N=4 & 0.1188 & 0.0530 & 0.3390 & 0.35886 & 0.67919 & 0.30511 \\
HT/Q2.5 & P=4, C=1, N=8 & 0.0816 & 0.0430 & 0.1880 & 0.65085 & 0.82297 & 0.58066 \\
HT/Q3 & P=2, C=1, N=4 & 0.2158 & 0.1520 & 0.4340 & 0.90142 & 0.95627 & 0.74780 \\
HT/Q3 & P=4, C=1, N=4 & 0.3206 & 0.2970 & 0.6550 & 0.47980 & 0.49807 & 0.30433 \\
HT/Q3 & P=4, C=1, N=8 & 0.4342 & 0.4170 & 0.6590 & 0.09900 & 0.11447 & 0.30433 \\
HT/GPT20 & P=2, C=1, N=2 & 0.8298 & 0.8590 & 0.9530 & 0.54280 & 0.43606 & \textbf{$<10^{-5}$} \\
HT/GPT20 & P=4, C=1 & 0.9268 & 0.9410 & 0.9970 & \textbf{0.00027} & \textbf{0.00258} & \textbf{$<10^{-5}$} \\
HT/GPT20 & P=4, C=1, N=2 & 0.7952 & 0.7850 & 0.8440 & 0.80663 & 0.87796 & 0.97799 \\
HT/GPT-120B & P=2, C=1, N=2 & 0.8862 & 0.8950 & 0.9570 & \textbf{0.00500} & 0.05300 & \textbf{$<10^{-5}$} \\
HT/GPT-120B & P=4, C=1, N=2 & 0.7728 & 0.8120 & 0.8680 & 0.46200 & 0.31800 & 0.75000 \\
SAI/Q2.5 & P=2, C=10, N=4 & 0.9408 & 0.9460 & 0.9620 & 0.50924 & 0.40912 & \textbf{0.04840} \\
SAI/Q2.5 & P=4, C=10, N=4 & 0.9342 & 0.9360 & 0.9480 & 0.85971 & 0.83689 & 0.90266 \\
SAI/Q3 & P=2, C=10, N=4 & 0.9508 & 0.9490 & 0.9660 & 0.21374 & 0.38751 & 0.26412 \\
SAI/Q3 & P=4, C=10, N=4 & 0.9408 & 0.9420 & 0.9630 & 0.82390 & 0.66493 & 0.37556 \\
SAI/GPT20 & P=2, C=0.1, N=2 & 0.9514 & 0.9570 & 0.9660 & 0.16287 & 0.16345 & 0.67160 \\
SAI/GPT20 & P=4, C=0.1, N=2 & 0.9392 & 0.9420 & 0.9550 & 0.59551 & 0.50014 & 0.86877 \\
SAI/GPT-120B & P=2, C=0.1, N=2 & 0.9050 & 0.9150 & 0.9160 & 1.00000 & 0.99100 & 1.00000 \\
SAI/GPT-120B & P=4, C=0.1, N=2 & 0.9354 & 0.9270 & 0.9630 & 0.06900 & 0.38500 & 0.13700 \\
\end{longtable}
\endgroup

\clearpage
\section{Full Budget-Matched Allocation Results}
\label{app:full_budget_matched_allocation_results}

Table~\ref{tab:expanded_strategy_deviation_appendix} reports the full budget-matched allocation sweep corresponding to the shortened main-text Table~\ref{tab:expanded_strategy_deviation}. The main table keeps the baselines and policies that finish with at most two surviving trajectories, since schedules with more final survivors are never column-best in this experiment. The appendix table restores those omitted schedules for completeness while using the same mean-score columns and unweighted average statistic.

The statistical comparison in this table is a simulation-based vector battle against a matched Sequential BoN reference. In each resampling iteration and for each task--model pair, we draw one score from the candidate policy's simulated outcome distribution. For pruning and ensembling policies, this score is obtained by replaying the policy against the available run pool; for the Sequential BoN reference row, it is obtained by sampling five Sequential BoN baseline runs with replacement and taking their maximum. We then independently sample five Sequential BoN baseline runs with replacement for the same task--model pair and again take the maximum. The policy and sampled baseline are compared pairwise: the policy wins the pair if its score is larger, the baseline wins if its score is larger, and exact equality is counted as a tie. Across all available task--model pairs, the iteration is decided by majority vote: the policy wins the vector battle if it wins more pairs than the baseline, the baseline wins if it wins more pairs, and the iteration is tied otherwise.

Repeating this procedure over many resampling iterations gives the win, loss, and tie rates reported for each strategy. The pair-level entries in the table are the mean simulated policy scores for each task--model pair, and the Avg. column is their unweighted average across the twelve pairs. To assess whether a strategy wins decisive vector battles more often than expected by chance, we use a one-sided binomial tail test after discarding vector-level ties. The null hypothesis is that, conditional on a decisive vector battle, the policy and baseline are equally likely to win. Thus, if a strategy wins \(W\) decisive battles and the sampled baseline wins \(L\), the p-value is the upper-tail probability of observing at least \(W\) wins under a \(\mathrm{Binomial}(W+L,0.5)\) distribution, with the analogous tail probability used when summarizing baseline wins. The Sequential BoN reference row therefore serves as a baseline-vs-baseline check: its nearly symmetric win rates and non-significant p-value indicate no evidence that either sampled side systematically wins.

\begin{table*}[!t]
\centering
\caption{
\textbf{Budget-matched evaluation.} Entries report mean final scores over $100,000$ empirical resampling trials under a five-full-run-equivalent budget. All score values are reported as percentages, obtained by multiplying the original values by 100. Avg. is the unweighted average across the twelve model--problem pairs, and SE is its standard error. Higher is better; bold indicates the best value in each score column, and underlined averages have a higher battle win rate than the baseline.}
\label{tab:expanded_strategy_deviation_appendix}
\normalsize
\setlength{\tabcolsep}{2.2pt}
\renewcommand{\arraystretch}{1.08}
\resizebox{\textwidth}{!}{%
\begin{tabular}{@{}l*{14}{r}@{}}
\toprule
Strategy / schedule
& \shortstack{CP\\Q3B} & \shortstack{CP\\Q4B} & \shortstack{CP\\GPT\\20B} & \shortstack{CP\\GPT\\120B} & \shortstack{HT\\Q3B} & \shortstack{HT\\Q4B} & \shortstack{HT\\GPT\\20B} & \shortstack{HT\\GPT\\120B} & \shortstack{SA\\Q3B} & \shortstack{SA\\Q4B} & \shortstack{SA\\GPT\\20B} & \shortstack{SA\\GPT\\120B} & Avg. & SE \\
\midrule
Sequential BoN reference, 5 full & 78.42 & 93.60 & 98.63 & 97.56 & 24.50 & \textbf{56.05} & 91.29 & 89.80 & 95.38 & 96.09 & \textbf{96.59} & 94.27 & 84.35 & 0.02 \\
Single-harness baseline, 5 full & 74.46 & 89.61 & 97.77 & 96.90 & 25.39 & 43.90 & 91.69 & 89.81 & 95.06 & 95.47 & 95.65 & 94.20 & 82.49 & 0.02 \\
Unpruned harness portfolio, 5 full & 77.33 & 93.13 & 98.37 & 97.37 & 29.11 & 51.08 & 94.01 & 91.89 & 95.36 & 95.97 & 96.15 & 94.73 & \underline{84.54} & 0.02 \\

\addlinespace[2pt]
\multicolumn{15}{@{}l}{\textit{Single-stage pruning at 25\%}} \\
25\%: 17$\to$1 & 77.74 & 94.78 & 98.68 & 97.39 & 35.82 & 45.48 & 93.92 & 92.57 & 95.44 & 96.30 & 96.03 & 94.27 & \underline{84.87} & 0.02 \\
25\%: 8$\to$4 & 78.40 & 94.02 & 98.50 & 97.54 & 32.03 & 52.32 & 94.65 & 92.91 & 95.48 & 96.16 & 96.28 & 94.76 & \underline{85.25} & 0.02 \\
25\%: 14$\to$2 & 78.68 & \textbf{94.79} & 98.72 & 97.60 & \textbf{35.93} & 48.59 & 94.60 & 93.55 & 95.49 & \textbf{96.32} & 96.26 & 94.29 & \underline{85.40} & 0.02 \\

\addlinespace[2pt]
\multicolumn{15}{@{}l}{\textit{Single-stage pruning at 50\%}} \\
50\%: 9$\to$1 & 78.13 & 94.07 & 98.50 & 97.58 & 32.97 & 50.58 & 94.25 & 92.73 & 95.45 & 96.17 & 96.18 & 94.53 & \underline{85.09} & 0.02 \\
50\%: 6$\to$4 & 77.75 & 93.56 & 98.49 & 97.48 & 30.55 & 52.85 & 94.46 & 92.61 & 95.44 & 96.08 & 96.21 & 94.71 & \underline{85.01} & 0.02 \\
50\%: 8$\to$2 & 78.55 & 94.06 & 98.59 & 97.59 & 32.69 & 52.29 & 94.65 & 93.07 & 95.48 & 96.18 & 96.25 & 94.50 & \underline{85.32} & 0.02 \\

\addlinespace[2pt]
\multicolumn{15}{@{}l}{\textit{Two-stage pruning at 25\%/50\%}} \\
7$\to$5$\to$4 & 78.22 & 93.85 & 98.49 & 97.54 & 31.31 & 53.10 & 94.55 & 92.56 & 95.47 & 96.14 & 96.27 & \textbf{94.78} & \underline{85.19} & 0.02 \\
8$\to$6$\to$3 & 78.76 & 94.09 & 98.59 & 97.62 & 32.51 & 53.27 & 94.79 & 93.00 & 95.49 & 96.20 & 96.31 & 94.62 & \underline{85.44} & 0.02 \\
9$\to$5$\to$3 & 78.99 & 94.26 & 98.64 & 97.65 & 33.19 & 52.92 & 94.91 & 93.29 & 95.50 & 96.22 & 96.32 & 94.61 & \underline{85.54} & 0.02 \\
10$\to$4$\to$3 & 79.01 & 94.41 & 98.66 & 97.64 & 33.77 & 52.00 & 95.02 & 93.60 & 95.51 & 96.25 & 96.31 & 94.49 & \underline{85.55} & 0.02 \\
10$\to$6$\to$2 & \textbf{79.24} & 94.39 & 98.67 & 97.69 & 34.03 & 52.87 & 94.90 & 93.40 & 95.50 & 96.26 & 96.32 & 94.62 & \underline{85.66} & 0.02 \\
10$\to$8$\to$1 & 78.52 & 94.23 & 98.55 & 97.63 & 33.48 & 51.35 & 94.44 & 92.96 & 95.47 & 96.21 & 96.23 & 94.63 & \underline{85.31} & 0.02 \\
12$\to$6$\to$1 & 78.79 & 94.48 & 98.66 & 97.69 & 34.73 & 51.29 & 94.82 & 93.44 & 95.49 & 96.25 & 96.23 & 94.62 & \underline{85.54} & 0.02 \\
14$\to$4$\to$1 & 78.56 & 94.67 & 98.72 & 97.69 & 35.58 & 50.17 & 95.05 & \textbf{93.78} & \textbf{95.51} & 96.28 & 96.22 & 94.38 & \underline{85.55} & 0.02 \\

\addlinespace[2pt]
\multicolumn{15}{@{}l}{\textit{Three-stage pruning at 25\%/50\%/75\%}} \\
7$\to$6$\to$4$\to$3 & 78.29 & 93.89 & 98.54 & 97.56 & 31.64 & 53.73 & 94.69 & 92.83 & 95.48 & 96.16 & 96.26 & 94.67 & \underline{85.31} & 0.02 \\
8$\to$5$\to$4$\to$3 & 78.66 & 94.08 & 98.57 & 97.61 & 32.37 & 53.51 & 94.79 & 92.96 & 95.49 & 96.19 & 96.29 & 94.71 & \underline{85.44} & 0.02 \\
8$\to$6$\to$5$\to$1 & 78.43 & 94.04 & 98.50 & 97.57 & 32.64 & 54.21 & 94.63 & 92.70 & 95.47 & 96.18 & 96.21 & 94.71 & \underline{85.44} & 0.02 \\
10$\to$5$\to$3$\to$2 & 79.23 & 94.42 & 98.68 & 97.69 & 33.99 & 53.09 & 95.07 & 93.50 & 95.51 & 96.26 & 96.31 & 94.58 & \underline{85.69} & 0.02 \\
10$\to$6$\to$3$\to$1 & 79.08 & 94.39 & 98.63 & 97.69 & 34.02 & 53.62 & 94.96 & 93.25 & 95.49 & 96.25 & 96.28 & 94.66 & \underline{85.69} & 0.02 \\
12$\to$5$\to$2$\to$1 & 79.17 & 94.58 & \textbf{98.73} & \textbf{97.72} & 35.07 & 52.42 & \textbf{95.11} & 93.62 & 95.51 & 96.28 & 96.26 & 94.51 & \textbf{\underline{85.75}} & 0.02 \\
\bottomrule
\end{tabular}%
}
\end{table*}

\begin{table*}[!t]
\centering
\caption{
\textbf{Budget-matched evaluation for PUCT.} Entries report mean final scores over $100,000$ empirical resampling trials under a five-full-run-equivalent budget. All score values are reported as percentages, obtained by multiplying the original values by 100. Avg. is the unweighted average across the twelve model--problem pairs, and SE is its standard error. Higher is better; bold indicates the best value in each score column, and underlined averages have a higher battle win rate than the baseline.}
\label{tab:expanded_strategy_deviation_puct_appendix}
\normalsize
\setlength{\tabcolsep}{2.2pt}
\renewcommand{\arraystretch}{1.08}
\resizebox{\textwidth}{!}{%
\begin{tabular}{@{}l*{14}{r}@{}}
\toprule
Strategy / schedule
& \shortstack{CP\\Q2.5} & \shortstack{CP\\Q3} & \shortstack{CP\\GPT\\20B} & \shortstack{CP\\GPT\\120B} & \shortstack{HT\\Q2.5} & \shortstack{HT\\Q3} & \shortstack{HT\\GPT\\20B} & \shortstack{HT\\GPT\\120B} & \shortstack{SA\\Q2.5} & \shortstack{SA\\Q3} & \shortstack{SA\\GPT\\20B} & \shortstack{SA\\GPT\\120B} & Avg. & SE \\
\midrule
Sequential BoN reference, 5 full & 78.37 & \textbf{93.58} & 98.62 & 97.55 & 24.35 & \textbf{55.90} & 91.31 & 89.75 & 95.39 & 96.09 & 96.60 & 94.27 & 84.32 & 0.06 \\
Single-harness baseline, 5 full & 77.27 & 91.16 & 99.08 & 96.98 & 25.17 & 52.49 & 94.01 & 93.11 & 95.19 & 96.05 & 96.65 & 94.36 & 84.29 & 0.06 \\
Unpruned harness portfolio, 5 full & 79.01 & 91.68 & 99.08 & 96.93 & 24.00 & 49.36 & 95.56 & 93.30 & 95.41 & 95.97 & 96.62 & 94.85 & \underline{84.31} & 0.07 \\

\addlinespace[2pt]
\multicolumn{15}{@{}l}{\textit{Single-stage pruning at 25\%}} \\
25\%: 17$\to$1 & 80.09 & 92.98 & 98.91 & 96.94 & \textbf{29.03} & 49.43 & 94.64 & 93.60 & 95.63 & 96.28 & 96.61 & 94.28 & \underline{84.87} & 0.06 \\
25\%: 8$\to$4 & 80.33 & 92.61 & 99.27 & 97.33 & 25.53 & 53.34 & 96.40 & 94.75 & 95.48 & 96.20 & 96.74 & 94.88 & \underline{85.24} & 0.06 \\
25\%: 14$\to$2 & \textbf{81.47} & 92.92 & 99.10 & 97.51 & 27.96 & 51.99 & 94.86 & 94.61 & 95.57 & 96.23 & 96.79 & 94.22 & \underline{85.27} & 0.06 \\

\addlinespace[2pt]
\multicolumn{15}{@{}l}{\textit{Single-stage pruning at 50\%}} \\
50\%: 9$\to$1 & 80.10 & 93.34 & 98.97 & 97.43 & 25.38 & 52.42 & 95.94 & 94.99 & 95.65 & 96.14 & 96.71 & 93.82 & \underline{85.07} & 0.06 \\
50\%: 6$\to$4 & 80.02 & 92.40 & 99.18 & 97.07 & 24.02 & 50.72 & 96.08 & 93.96 & 95.52 & 96.09 & 96.73 & 94.89 & \underline{84.72} & 0.07 \\
50\%: 8$\to$2 & 80.99 & 93.23 & 99.23 & 97.30 & 25.07 & 52.52 & 96.54 & 94.65 & \textbf{95.68} & 96.21 & 96.76 & 94.05 & \underline{85.19} & 0.07 \\

\addlinespace[2pt]
\multicolumn{15}{@{}l}{\textit{Two-stage pruning at 25\%/50\%}} \\
7$\to$5$\to$4 & 80.37 & 92.53 & 99.21 & 97.20 & 24.99 & 52.27 & 96.33 & 94.40 & 95.56 & 96.18 & 96.79 & \textbf{94.94} & \underline{85.06} & 0.07 \\
8$\to$6$\to$3 & 81.21 & 93.08 & 99.29 & 97.33 & 25.77 & 53.18 & \textbf{96.67} & 94.72 & 95.65 & 96.24 & 96.79 & 94.46 & \underline{85.37} & 0.06 \\
9$\to$5$\to$3 & 81.12 & 93.09 & 99.22 & 97.42 & 26.36 & 54.23 & 96.31 & 95.00 & 95.51 & 96.21 & 96.79 & 94.42 & \underline{85.47} & 0.06 \\
10$\to$4$\to$3 & 80.88 & 92.83 & 99.24 & 97.51 & 25.83 & 54.08 & 96.05 & 95.13 & 95.47 & 96.22 & \textbf{96.80} & 94.41 & \underline{85.37} & 0.06 \\
10$\to$6$\to$2 & 81.45 & 93.07 & 99.21 & 97.52 & 25.64 & 53.97 & 96.50 & 95.24 & 95.58 & 96.23 & \textbf{96.81} & 94.10 & \underline{85.44} & 0.06 \\
10$\to$8$\to$1 & 80.38 & 93.21 & 99.04 & 97.50 & 24.98 & 52.64 & 96.19 & 95.20 & 95.67 & 96.19 & 96.74 & 93.89 & \underline{85.14} & 0.07 \\
12$\to$6$\to$1 & 80.45 & 93.21 & 99.20 & 97.70 & 26.47 & 54.40 & 96.44 & \textbf{95.65} & 95.59 & 96.25 & 96.76 & 94.04 & \underline{85.51} & 0.06 \\
14$\to$4$\to$1 & 80.80 & 93.21 & 99.26 & \textbf{97.80} & 27.73 & 54.93 & 95.74 & 95.60 & 95.57 & 96.28 & 96.78 & 94.15 & \textbf{\underline{85.65}} & 0.06 \\

\addlinespace[2pt]
\multicolumn{15}{@{}l}{\textit{Three-stage pruning at 25\%/50\%/75\%}} \\
7$\to$6$\to$4$\to$3 & 80.52 & 92.77 & 99.19 & 97.21 & 25.26 & 52.23 & 96.40 & 94.39 & 95.60 & 96.19 & \textbf{96.80} & 94.81 & \underline{85.11} & 0.06 \\
8$\to$5$\to$4$\to$3 & 80.84 & 92.81 & \textbf{99.31} & 97.31 & 26.07 & 53.48 & 96.64 & 94.73 & 95.58 & 96.24 & 96.78 & 94.81 & \underline{85.38} & 0.06 \\
8$\to$6$\to$5$\to$1 & 80.63 & 93.04 & 99.10 & 97.33 & 26.31 & 53.73 & \textbf{96.67} & 94.73 & 95.63 & 96.25 & 96.70 & 94.44 & \underline{85.38} & 0.06 \\
10$\to$5$\to$3$\to$2 & 81.27 & 92.99 & 99.24 & 97.52 & 26.10 & 54.64 & 96.51 & 95.26 & 95.52 & 96.24 & \textbf{96.80} & 94.34 & \underline{85.54} & 0.06 \\
10$\to$6$\to$3$\to$1 & 81.18 & 93.06 & 99.15 & 97.51 & 26.12 & 54.58 & 96.54 & 95.25 & 95.58 & 96.26 & 96.76 & 94.16 & \underline{85.51} & 0.06 \\
12$\to$5$\to$2$\to$1 & 80.72 & 93.22 & 99.26 & 97.69 & 26.85 & 55.12 & 96.37 & 95.57 & 95.55 & \textbf{96.28} & 96.76 & 94.12 & \underline{85.62} & 0.06 \\
\bottomrule
\end{tabular}%
}
\end{table*}

\begin{table*}[!t]
\centering
\caption{
\textbf{Budget-matched evaluation for UCT.} Entries report mean final scores over $100,000$ empirical resampling trials under a five-full-run-equivalent budget. All score values are reported as percentages, obtained by multiplying the original values by 100. Avg. is the unweighted average across the twelve model--problem pairs, and SE is its standard error. Higher is better; bold indicates the best value in each score column, and underlined averages have a higher battle win rate than the baseline.}
\label{tab:expanded_strategy_deviation_uct_appendix}
\normalsize
\setlength{\tabcolsep}{2.2pt}
\renewcommand{\arraystretch}{1.08}
\resizebox{\textwidth}{!}{%
\begin{tabular}{@{}l*{14}{r}@{}}
\toprule
Strategy / schedule
& \shortstack{CP\\Q2.5} & \shortstack{CP\\Q3} & \shortstack{CP\\GPT\\20B} & \shortstack{CP\\GPT\\120B} & \shortstack{HT\\Q2.5} & \shortstack{HT\\Q3} & \shortstack{HT\\GPT\\20B} & \shortstack{HT\\GPT\\120B} & \shortstack{SA\\Q2.5} & \shortstack{SA\\Q3} & \shortstack{SA\\GPT\\20B} & \shortstack{SA\\GPT\\120B} & Avg. & SE \\
\midrule
Sequential BoN reference, 5 full & \textbf{78.45} & \textbf{93.56} & \textbf{98.62} & 97.58 & 24.36 & 56.19 & 91.34 & 89.87 & 95.38 & \textbf{96.10} & \textbf{96.60} & 94.28 & 84.36 & 0.06 \\
Single-harness baseline, 5 full & 74.55 & 90.27 & 97.69 & 97.99 & 33.24 & 52.44 & 94.90 & 92.47 & 95.27 & 95.63 & 96.16 & \textbf{95.86} & 84.71 & 0.06 \\
Unpruned harness portfolio, 5 full & 74.45 & 92.56 & 97.91 & 97.98 & 29.08 & 56.97 & 94.78 & 92.53 & 95.23 & 95.42 & 96.18 & 95.32 & 84.87 & 0.06 \\

\addlinespace[2pt]
\multicolumn{15}{@{}l}{\textit{Single-stage pruning at 25\%}} \\
25\%: 17$\to$1 & 76.53 & 91.77 & 98.08 & 96.93 & \textbf{39.76} & 37.85 & 95.04 & 89.49 & 95.20 & 95.75 & 95.81 & 93.44 & 83.80 & 0.06 \\
25\%: 8$\to$4 & 75.62 & 92.74 & 98.00 & 97.96 & 33.93 & 53.57 & 95.31 & 91.96 & \textbf{95.49} & 95.61 & 96.13 & 95.78 & \underline{85.18} & 0.06 \\
25\%: 14$\to$2 & 76.42 & 92.50 & 98.18 & 97.56 & 38.58 & 41.21 & 95.18 & 90.98 & 95.38 & 95.68 & 95.85 & 94.41 & 84.33 & 0.06 \\

\addlinespace[2pt]
\multicolumn{15}{@{}l}{\textit{Single-stage pruning at 50\%}} \\
50\%: 9$\to$1 & 75.14 & 92.05 & 98.08 & 98.18 & 34.01 & 44.01 & 94.72 & 90.38 & 95.19 & 95.49 & 96.11 & 95.43 & 84.07 & 0.06 \\
50\%: 6$\to$4 & 74.87 & 92.81 & 98.08 & 98.00 & 31.18 & 57.48 & 95.02 & \textbf{92.96} & 95.35 & 95.48 & 96.13 & 95.48 & \underline{85.24} & 0.06 \\
50\%: 8$\to$2 & 75.21 & 92.57 & 98.27 & 98.12 & 33.99 & 48.20 & 95.06 & 92.18 & 95.29 & 95.51 & 96.12 & 95.39 & 84.66 & 0.06 \\

\addlinespace[2pt]
\multicolumn{15}{@{}l}{\textit{Two-stage pruning at 25\%/50\%}} \\
7$\to$5$\to$4 & 75.32 & 92.84 & 98.05 & 98.04 & 32.60 & 56.53 & 95.22 & 92.44 & 95.42 & 95.58 & 96.20 & 95.67 & \underline{85.33} & 0.06 \\
8$\to$6$\to$3 & 75.56 & 92.88 & 98.19 & 98.10 & 33.81 & 52.11 & 95.25 & 92.39 & 95.42 & 95.57 & 96.17 & 95.60 & \underline{85.09} & 0.06 \\
9$\to$5$\to$3 & 75.80 & 92.85 & 98.04 & 98.07 & 35.32 & 50.41 & 95.25 & 91.67 & 95.42 & 95.55 & 96.23 & 95.69 & \underline{85.03} & 0.06 \\
10$\to$4$\to$3 & 76.00 & 92.54 & 98.03 & 98.01 & 35.59 & 48.29 & 95.35 & 91.49 & 95.45 & 95.59 & 96.17 & 95.62 & \underline{84.84} & 0.06 \\
10$\to$6$\to$2 & 75.91 & 92.46 & 98.10 & 98.22 & 35.73 & 47.58 & 95.23 & 90.70 & 95.33 & 95.62 & 96.27 & 95.59 & \underline{84.73} & 0.06 \\
10$\to$8$\to$1 & 75.51 & 91.98 & 98.10 & 98.25 & 34.29 & 44.21 & 94.85 & 90.07 & 95.23 & 95.55 & 96.17 & 95.56 & 84.15 & 0.06 \\
12$\to$6$\to$1 & 76.04 & 92.26 & 98.00 & \textbf{98.32} & 36.03 & 44.63 & 95.22 & 88.91 & 95.32 & 95.68 & 96.27 & 95.67 & 84.36 & 0.06 \\
14$\to$4$\to$1 & 76.16 & 92.32 & 98.10 & 98.11 & 37.90 & 41.43 & \textbf{95.42} & 89.09 & 95.38 & 95.70 & 96.09 & 95.11 & 84.24 & 0.06 \\

\addlinespace[2pt]
\multicolumn{15}{@{}l}{\textit{Three-stage pruning at 25\%/50\%/75\%}} \\
7$\to$6$\to$4$\to$3 & 75.33 & 92.82 & 98.15 & 98.06 & 32.67 & 57.63 & 95.13 & 92.87 & 95.41 & 95.54 & 96.23 & 95.64 & \underline{85.46} & 0.06 \\
8$\to$5$\to$4$\to$3 & 75.61 & 92.83 & 98.09 & 98.08 & 33.82 & 55.66 & 95.31 & 92.21 & 95.47 & 95.66 & 96.20 & 95.78 & \underline{85.39} & 0.06 \\
8$\to$6$\to$5$\to$1 & 75.69 & 92.52 & 98.19 & 98.10 & 33.38 & \textbf{58.55} & 94.89 & 91.51 & 95.37 & 95.60 & 96.20 & 95.69 & \textbf{\underline{85.47}} & 0.06 \\
10$\to$5$\to$3$\to$2 & 76.08 & 92.62 & 98.08 & 98.15 & 35.40 & 49.24 & 95.31 & 90.83 & 95.41 & 95.63 & 96.26 & 95.74 & \underline{84.90} & 0.06 \\
10$\to$6$\to$3$\to$1 & 76.00 & 92.41 & 98.13 & 98.23 & 35.04 & 49.93 & 95.10 & 89.79 & 95.36 & 95.61 & 96.29 & 95.70 & \underline{84.80} & 0.06 \\
12$\to$5$\to$2$\to$1 & 76.29 & 92.52 & 98.14 & 98.21 & 36.92 & 46.10 & 95.33 & 89.02 & 95.36 & 95.71 & 96.24 & 95.56 & \underline{84.62} & 0.06 \\
\bottomrule
\end{tabular}%
}
\end{table*}

\begin{table*}[!t]
\centering
\caption{
\textbf{Budget-matched evaluation for TopK archive.} Entries report mean final scores over $100,000$ empirical resampling trials under a five-full-run-equivalent budget. All score values are reported as percentages, obtained by multiplying the original values by 100. Avg. is the unweighted average across the twelve model--problem pairs, and SE is its standard error. Higher is better; bold indicates the best value in each score column, and underlined averages have a higher battle win rate than the baseline.}
\label{tab:expanded_strategy_deviation_topk_archive_appendix}
\normalsize
\setlength{\tabcolsep}{2.2pt}
\renewcommand{\arraystretch}{1.08}
\resizebox{\textwidth}{!}{%
\begin{tabular}{@{}l*{14}{r}@{}}
\toprule
Strategy / schedule
& \shortstack{CP\\Q2.5} & \shortstack{CP\\Q3} & \shortstack{CP\\GPT\\20B} & \shortstack{CP\\GPT\\120B} & \shortstack{HT\\Q2.5} & \shortstack{HT\\Q3} & \shortstack{HT\\GPT\\20B} & \shortstack{HT\\GPT\\120B} & \shortstack{SA\\Q2.5} & \shortstack{SA\\Q3} & \shortstack{SA\\GPT\\20B} & \shortstack{SA\\GPT\\120B} & Avg. & SE \\
\midrule
Sequential BoN reference, 5 full & 78.46 & 93.59 & 98.63 & 97.55 & 24.34 & \textbf{56.18} & 91.33 & 89.85 & 95.38 & 96.08 & \textbf{96.59} & 94.28 & 84.35 & 0.06 \\
Single-harness baseline, 5 full & 77.29 & 91.34 & 97.91 & 97.24 & 31.82 & 47.30 & 91.47 & 91.77 & 95.37 & 95.81 & 95.86 & 94.24 & 83.95 & 0.06 \\
Unpruned harness portfolio, 5 full & 77.83 & 93.77 & 97.91 & 97.27 & 31.26 & 50.05 & 92.60 & 90.84 & 95.40 & 96.06 & 95.96 & 94.44 & 84.45 & 0.06 \\

\addlinespace[2pt]
\multicolumn{15}{@{}l}{\textit{Single-stage pruning at 25\%}} \\
25\%: 17$\to$1 & 76.31 & 95.17 & \textbf{98.67} & 97.40 & 34.47 & 45.93 & 91.09 & 91.64 & 95.43 & 96.37 & 95.72 & 94.28 & \underline{84.37} & 0.06 \\
25\%: 8$\to$4 & 78.31 & 94.67 & 98.13 & 97.38 & 33.34 & 52.08 & 92.44 & 92.16 & 95.50 & 96.24 & 96.04 & 94.38 & \underline{85.06} & 0.06 \\
25\%: 14$\to$2 & 78.36 & \textbf{95.19} & 98.50 & 97.41 & 36.11 & 48.53 & 92.26 & 92.92 & 95.47 & \textbf{96.41} & 96.01 & 94.24 & \underline{85.12} & 0.06 \\

\addlinespace[2pt]
\multicolumn{15}{@{}l}{\textit{Single-stage pruning at 50\%}} \\
50\%: 9$\to$1 & 77.99 & 94.63 & 98.09 & 97.43 & 34.63 & 52.30 & 92.25 & 92.15 & 95.40 & 96.25 & 95.89 & 94.37 & \underline{85.12} & 0.06 \\
50\%: 6$\to$4 & 77.67 & 94.21 & 98.04 & 97.36 & 32.63 & 52.18 & 92.97 & 91.58 & 95.44 & 96.18 & 96.02 & 94.37 & \underline{84.89} & 0.06 \\
50\%: 8$\to$2 & 78.51 & 94.66 & 98.09 & 97.47 & 34.50 & 53.55 & \textbf{93.10} & 92.38 & 95.47 & 96.27 & 95.99 & 94.35 & \underline{85.36} & 0.06 \\

\addlinespace[2pt]
\multicolumn{15}{@{}l}{\textit{Two-stage pruning at 25\%/50\%}} \\
7$\to$5$\to$4 & 78.07 & 94.48 & 98.14 & 97.39 & 32.95 & 53.07 & 92.69 & 91.69 & 95.47 & 96.22 & 96.04 & 94.46 & \underline{85.06} & 0.06 \\
8$\to$6$\to$3 & 78.71 & 94.71 & 98.22 & 97.48 & 34.07 & 54.37 & 92.96 & 92.18 & 95.49 & 96.29 & 96.06 & 94.41 & \underline{85.41} & 0.06 \\
9$\to$5$\to$3 & 78.95 & 94.84 & 98.28 & 97.53 & 34.62 & 53.85 & 92.87 & 92.65 & 95.52 & 96.31 & 96.06 & 94.40 & \underline{85.49} & 0.06 \\
10$\to$4$\to$3 & 79.21 & 94.94 & 98.32 & 97.48 & 34.82 & 52.34 & 92.67 & 92.99 & \textbf{95.54} & 96.34 & 96.05 & 94.33 & \underline{85.42} & 0.06 \\
10$\to$6$\to$2 & \textbf{79.26} & 94.93 & 98.34 & 97.59 & 35.40 & 54.55 & 92.81 & 93.00 & 95.51 & 96.35 & 96.02 & 94.44 & \textbf{\underline{85.68}} & 0.06 \\
10$\to$8$\to$1 & 78.32 & 94.80 & 98.22 & 97.46 & 34.74 & 52.98 & 92.24 & 92.45 & 95.44 & 96.29 & 95.88 & \textbf{94.47} & \underline{85.27} & 0.06 \\
12$\to$6$\to$1 & 78.72 & 94.97 & 98.34 & 97.56 & 35.71 & 52.38 & 92.36 & 93.23 & 95.50 & 96.34 & 95.85 & 94.41 & \underline{85.45} & 0.06 \\
14$\to$4$\to$1 & 78.70 & 95.11 & 98.49 & 97.40 & \textbf{36.43} & 50.16 & 92.42 & \textbf{93.33} & 95.52 & 96.35 & 95.78 & 94.25 & \underline{85.33} & 0.06 \\

\addlinespace[2pt]
\multicolumn{15}{@{}l}{\textit{Three-stage pruning at 25\%/50\%/75\%}} \\
7$\to$6$\to$4$\to$3 & 78.26 & 94.51 & 98.17 & 97.44 & 33.29 & 53.48 & 93.09 & 91.75 & 95.49 & 96.24 & 96.00 & 94.37 & \underline{85.17} & 0.06 \\
8$\to$5$\to$4$\to$3 & 78.53 & 94.68 & 98.22 & 97.47 & 33.82 & 53.57 & 92.76 & 92.17 & 95.51 & 96.27 & 96.01 & 94.39 & \underline{85.28} & 0.06 \\
8$\to$6$\to$5$\to$1 & 78.28 & 94.65 & 98.21 & 97.48 & 34.19 & 53.71 & 92.81 & 91.87 & 95.46 & 96.24 & 95.90 & 94.38 & \underline{85.27} & 0.06 \\
10$\to$5$\to$3$\to$2 & 79.22 & 94.94 & 98.36 & 97.58 & 35.36 & 53.81 & 92.89 & 93.01 & \textbf{95.54} & 96.34 & 96.01 & 94.38 & \underline{85.62} & 0.06 \\
10$\to$6$\to$3$\to$1 & 78.87 & 94.94 & 98.34 & \textbf{97.60} & 35.40 & 54.69 & 92.89 & 92.65 & 95.48 & 96.33 & 95.94 & \textbf{94.47} & \underline{85.64} & 0.06 \\
12$\to$5$\to$2$\to$1 & 79.25 & 95.06 & 98.39 & 97.59 & 35.88 & 53.13 & 92.58 & 93.28 & 95.51 & 96.35 & 95.87 & 94.39 & \underline{85.61} & 0.06 \\
\bottomrule
\end{tabular}%
}
\end{table*}

\end{document}